\documentclass[10pt,twocolumn,letterpaper]{article}

\usepackage{3dv}
\usepackage{times}
\usepackage{epsfig}
\usepackage{graphicx}
\usepackage{amsmath}
\usepackage{amssymb}
\usepackage{mathtools}
\usepackage{array}
\usepackage{multirow}
\usepackage{nicefrac}       %
\usepackage{microtype}  
\usepackage{tabularx}
\usepackage{enumitem}\usepackage{wrapfig,booktabs}
\usepackage{pifont}%
\newcommand{\cmark}{\ding{51}}%
\usepackage{afterpage}
\usepackage{scalerel,xparse}
\usepackage{capt-of,etoolbox}
\usepackage{xcolor}
\usepackage{float}
\newcolumntype{N}{>{\centering\arraybackslash\footnotesize}m{.5in}}
\newcolumntype{G}{>{\centering\arraybackslash}m{2in}}
\newcommand{\vect}[1]{{\bf {#1}}}
\newcommand{\imi}[0]{{I}}
\newcommand{\imis}[0]{{I}_s}
\newcommand{\imit}[0]{{I}_t}

\newcommand{\cp}[3]{{CP}(#1; #2; #3)}
\newcommand{\mask}[0]{{M}}
\newcommand{\sqnorm}[1]{{\mid\!\mid\!\!{#1}\!\!\mid\!\mid^2}}

\definecolor{darkgreen}{rgb}{0.0, 0.5, 0.0}

\usepackage[pagebackref=true,breaklinks=true,colorlinks,bookmarks=false]{hyperref}

\threedvfinalcopy %

\def\threedvPaperID{176} %

\ifthreedvfinal\fi
\begin{document}

\title{Cut-and-Paste Object Insertion by Enabling Deep Image Prior for Reshading}

\author{Anand Bhattad \hspace{3em} D.A. Forsyth\\
University of Illinois Urbana Champaign\\
{\tt\small \{bhattad2, daf\}@illinois.edu}
}
\makeatletter
\g@addto@macro\@maketitle{
	\begin{figure}[H]
		\setlength{\linewidth}{\textwidth}
		\setlength{\hsize}{\textwidth}
		\vspace{-8mm}
		\centering
    \begin{tabular}{@{}c@{\hspace{0.005\linewidth}}c@{\hspace{0.005\linewidth}}c@{\hspace{0.005\linewidth}}c@{}}
  \includegraphics[width=0.24\linewidth]{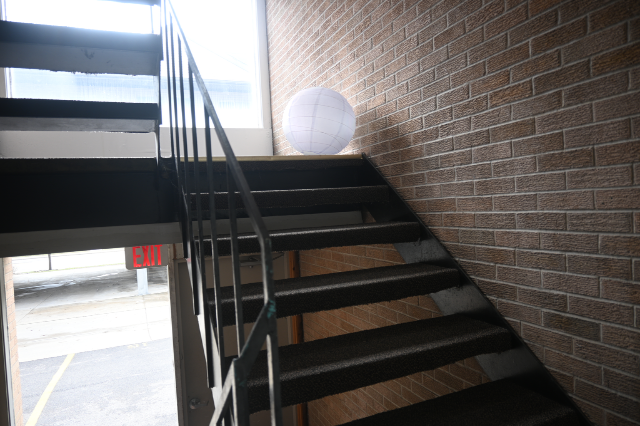} &
  \includegraphics[width=0.24\linewidth]{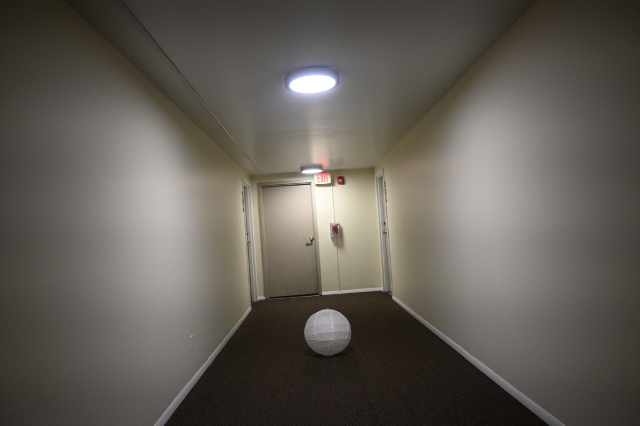} & 
  \includegraphics[width=0.24\linewidth]{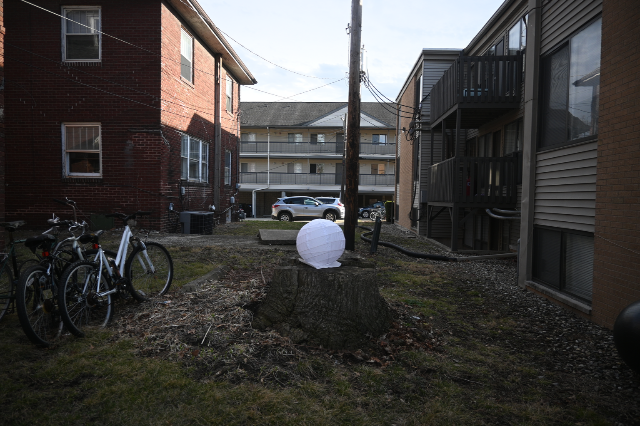} & 
  \includegraphics[width=0.24\linewidth]{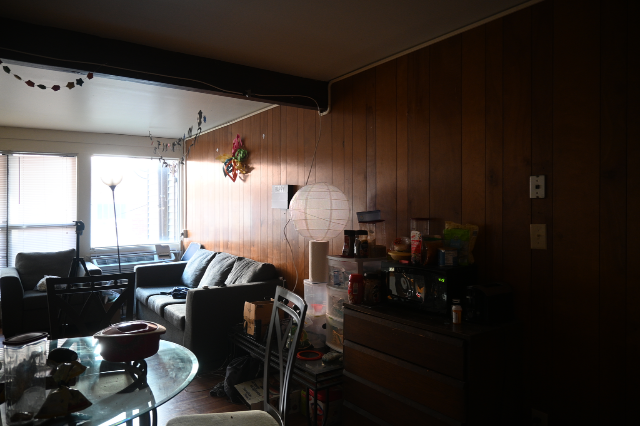}\vspace{-2pt}\\
  \small
  \centering\arraybackslash
(a) & \small \centering\arraybackslash(b)  & \small\centering\arraybackslash (c) & \small\centering\arraybackslash (d)
\end{tabular} 
\caption{In two of these images, the spherical lampshade is produced by our method creates realistically shaded renderings from cut-and-paste information; 
the other two are real photographs.  Can you tell which is which? Answer in Sec \ref{sec:experiments}. 
\label{fig:teaser}
}
\end{figure}
}
\makeatother
\maketitle
\begin{abstract}
   We show how to insert an object from one image to another and get realistic results in the hard case, where the shading of the inserted object clashes with the shading of the scene. Rendering objects using an illumination model of the scene doesn't work, because doing so requires a geometric and material model of the object, which is hard to recover from a single image. In this paper, we introduce a method that corrects shading inconsistencies of the inserted object without requiring a geometric and physical model or an environment map. Our method uses a deep image prior (DIP), trained to produce reshaded renderings of inserted objects via consistent image decomposition inferential losses. The resulting image from DIP aims to have (a) an albedo similar to the cut-and-paste albedo, (b) a similar shading field to that of the target scene, and (c) a shading that is consistent with the cut-and-paste surface normals. The result is a simple procedure that produces convincing shading of the inserted object. We show the efficacy of our method both qualitatively and quantitatively for several objects with complex surface properties and also on a dataset of spherical lampshades for quantitative evaluation. Our method significantly outperforms an Image Harmonization (IH) baseline for all these objects. They also outperform the cut-and-paste and IH baselines in a user study with over 100 users.
 \end{abstract}
\vspace{-10pt}
\section{Introduction}
\begin{figure*}[t]
  \centering
  \resizebox{0.9\linewidth}{!}
{
\setkeys{Gin}{width=0.98\linewidth}
    \renewcommand\arraystretch{0}
    \begin{tabularx}{\linewidth}{@{}X@{}X@{}X@{}X@{}}
      \includegraphics{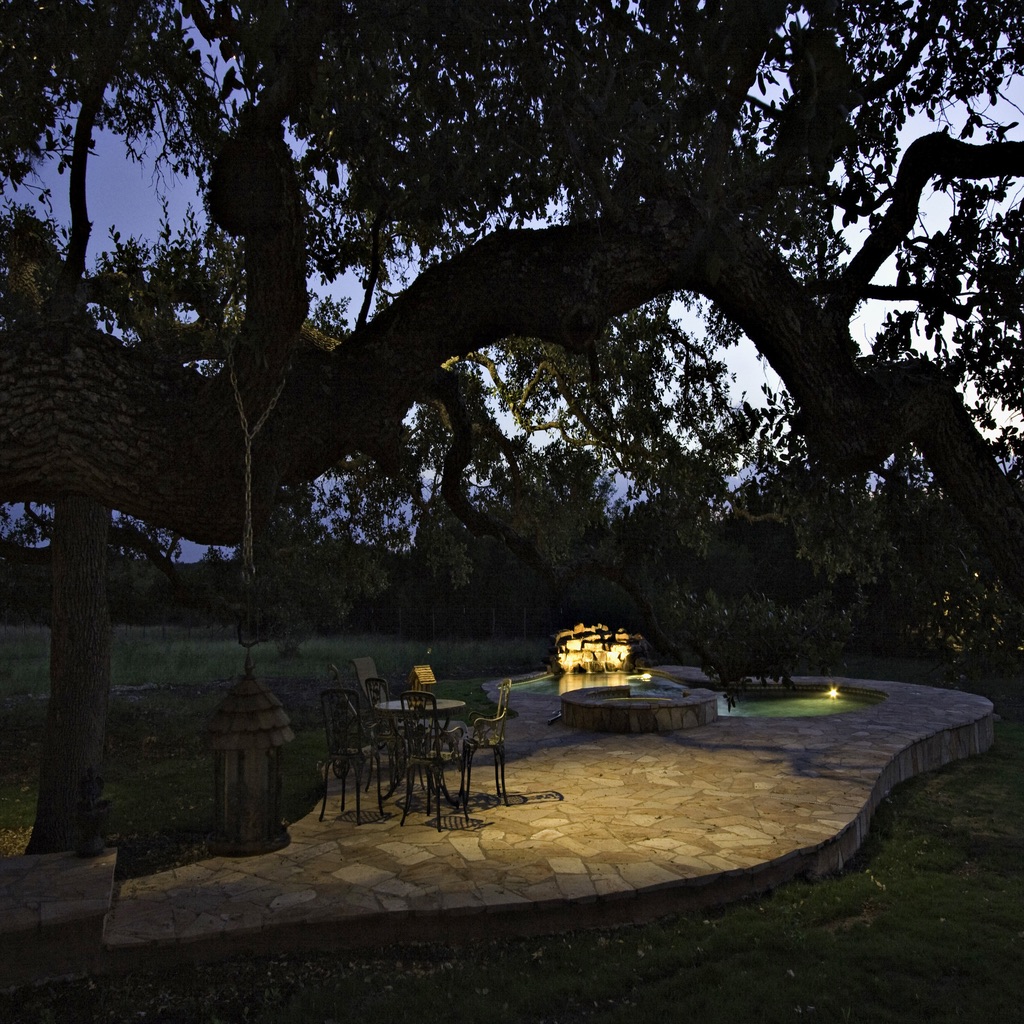} &
      \includegraphics{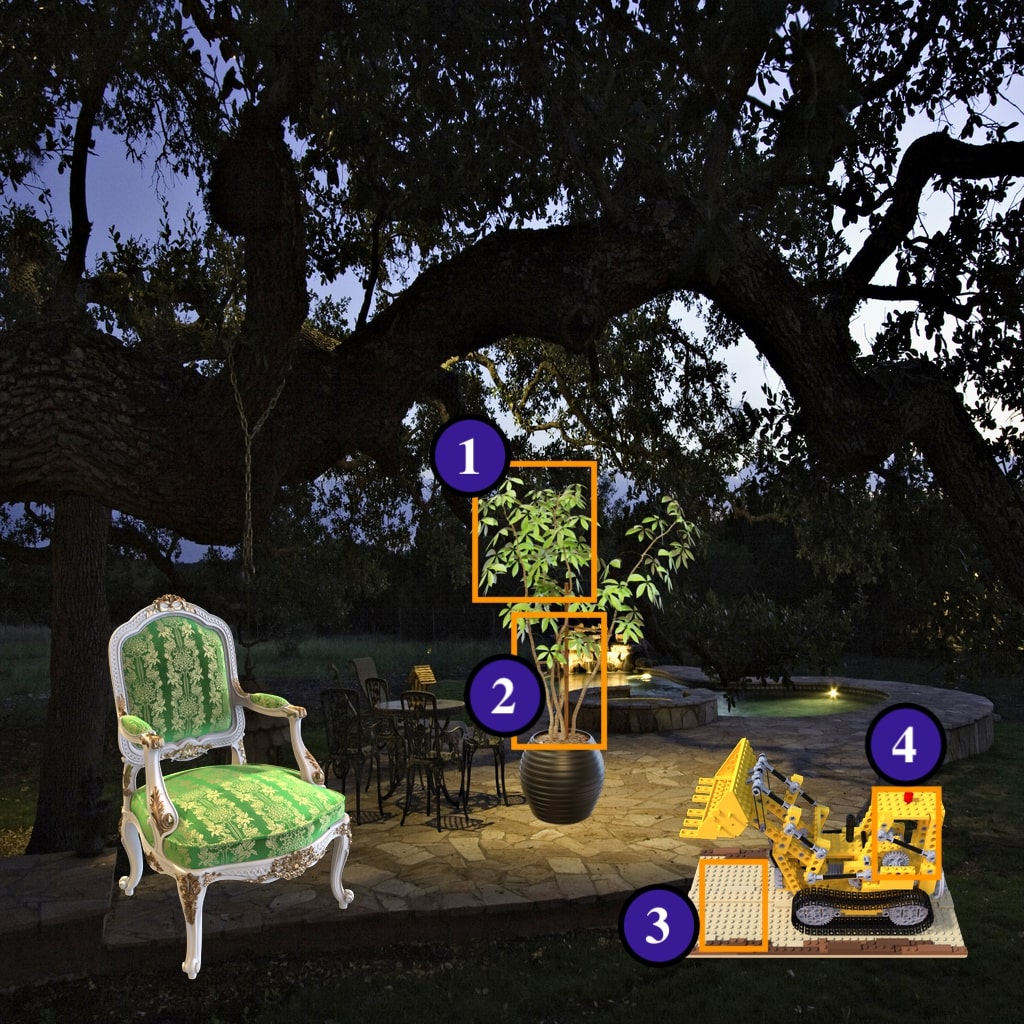} & 
      \includegraphics{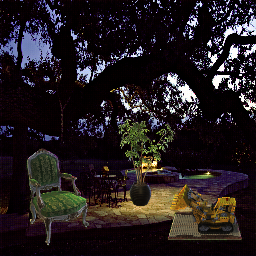} & 
      \includegraphics{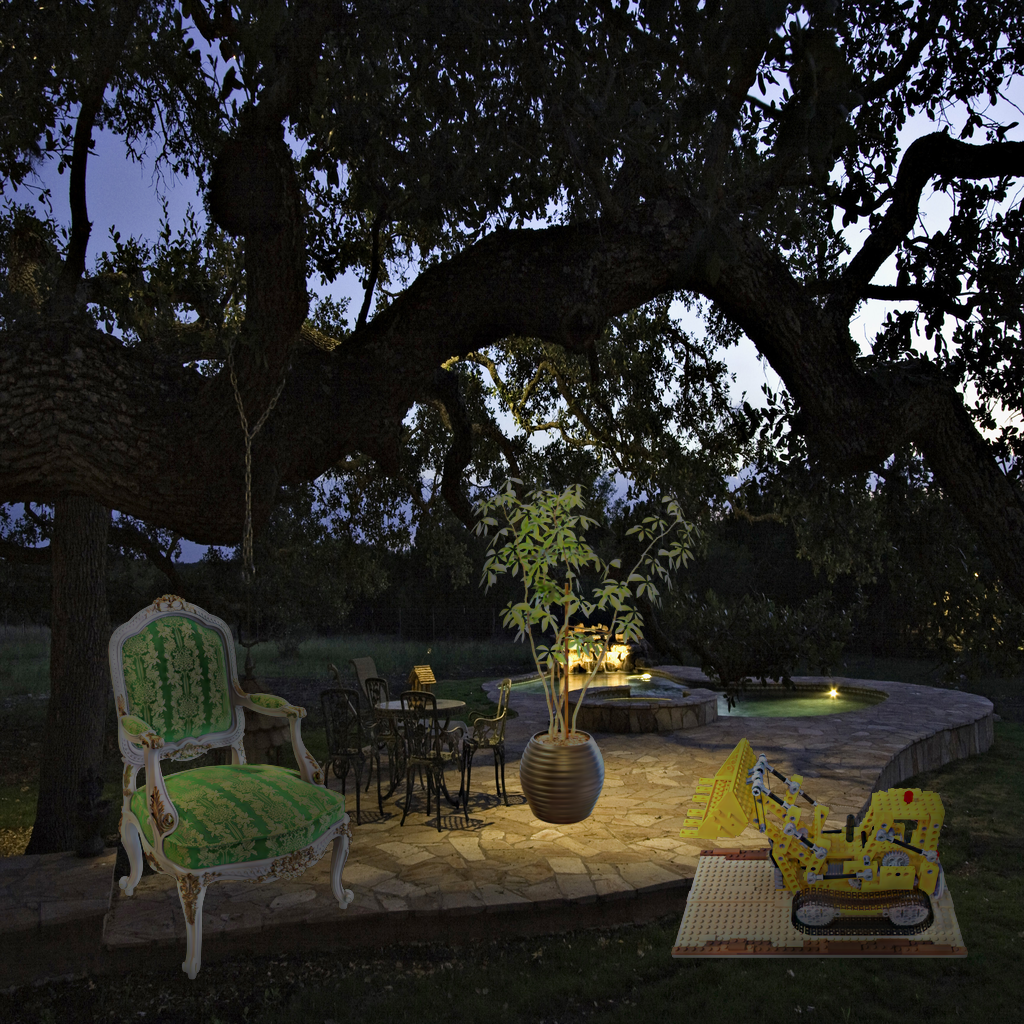}\\
      \includegraphics{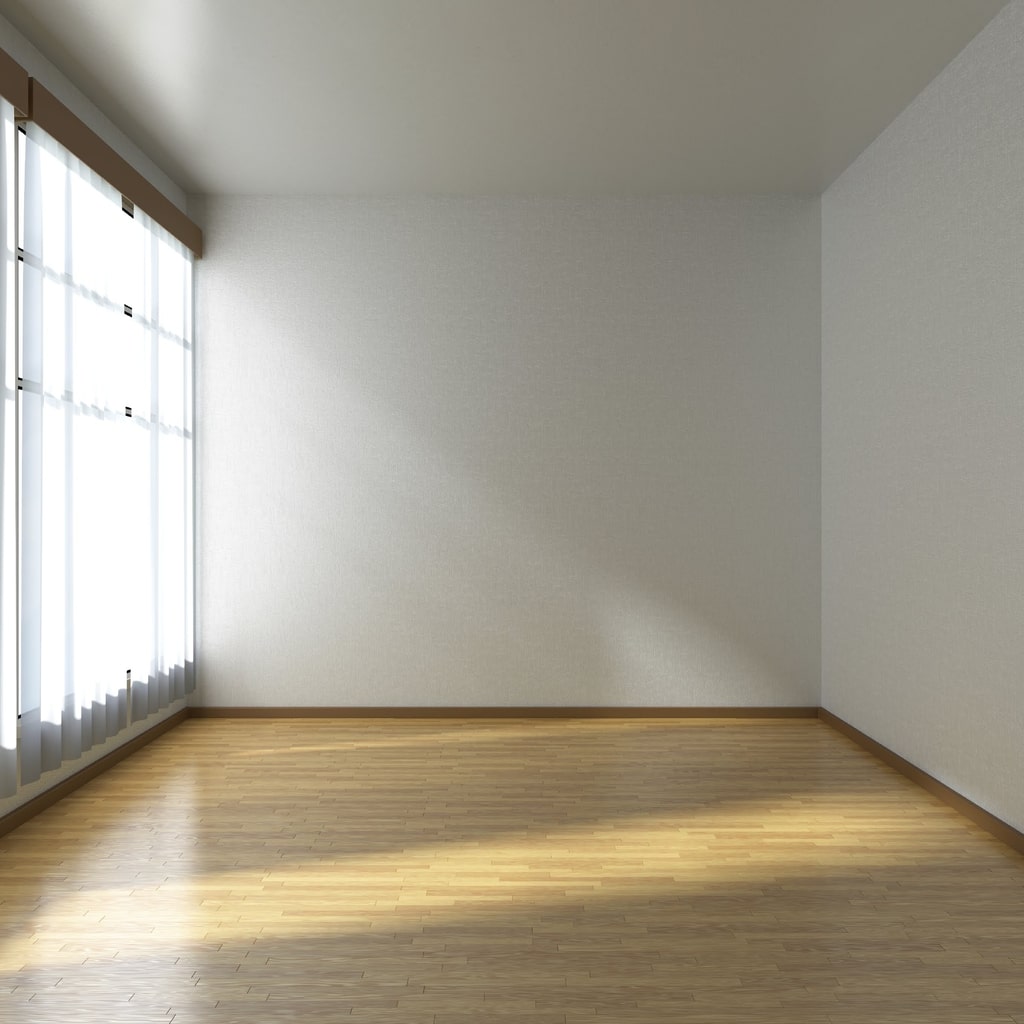} &
      \includegraphics{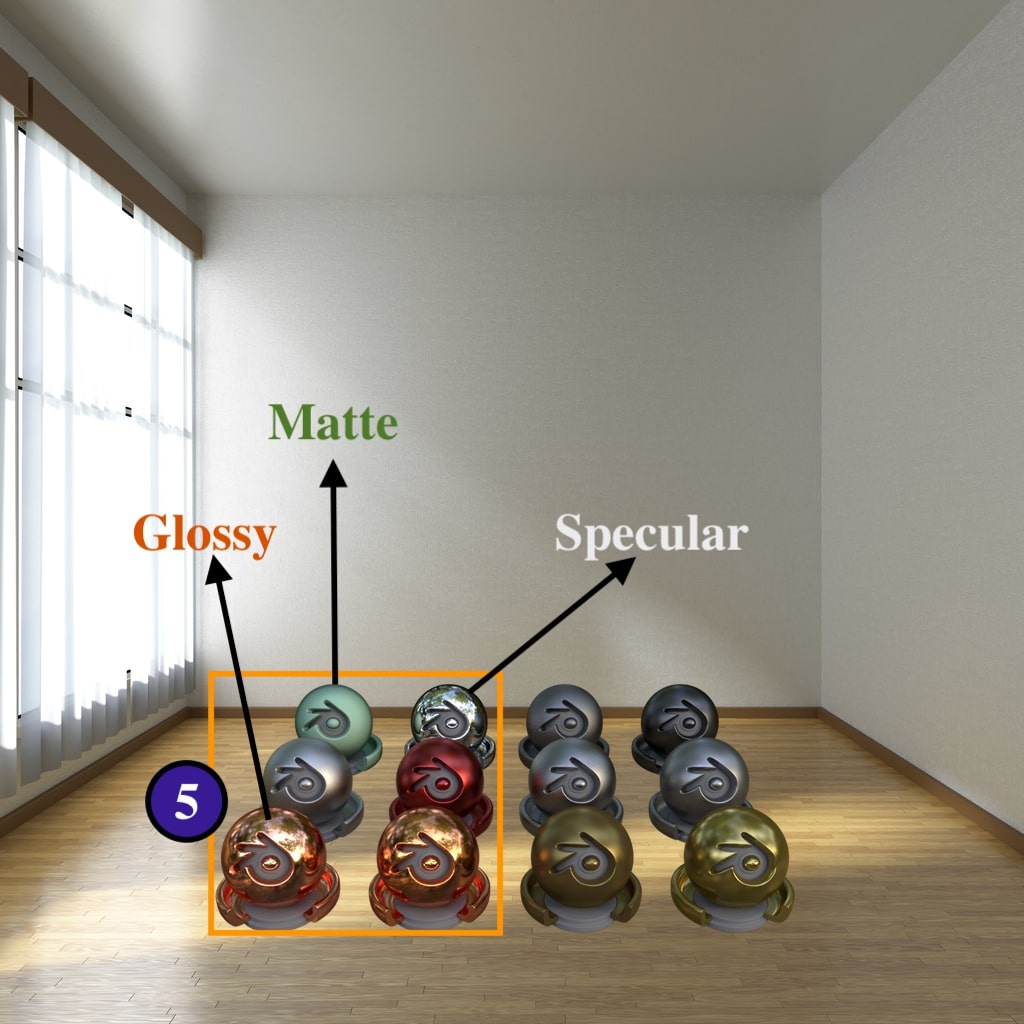} & 
      \includegraphics{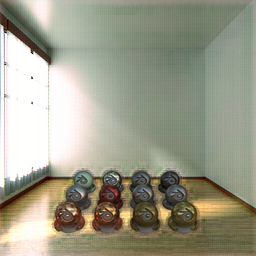} & 
      \includegraphics{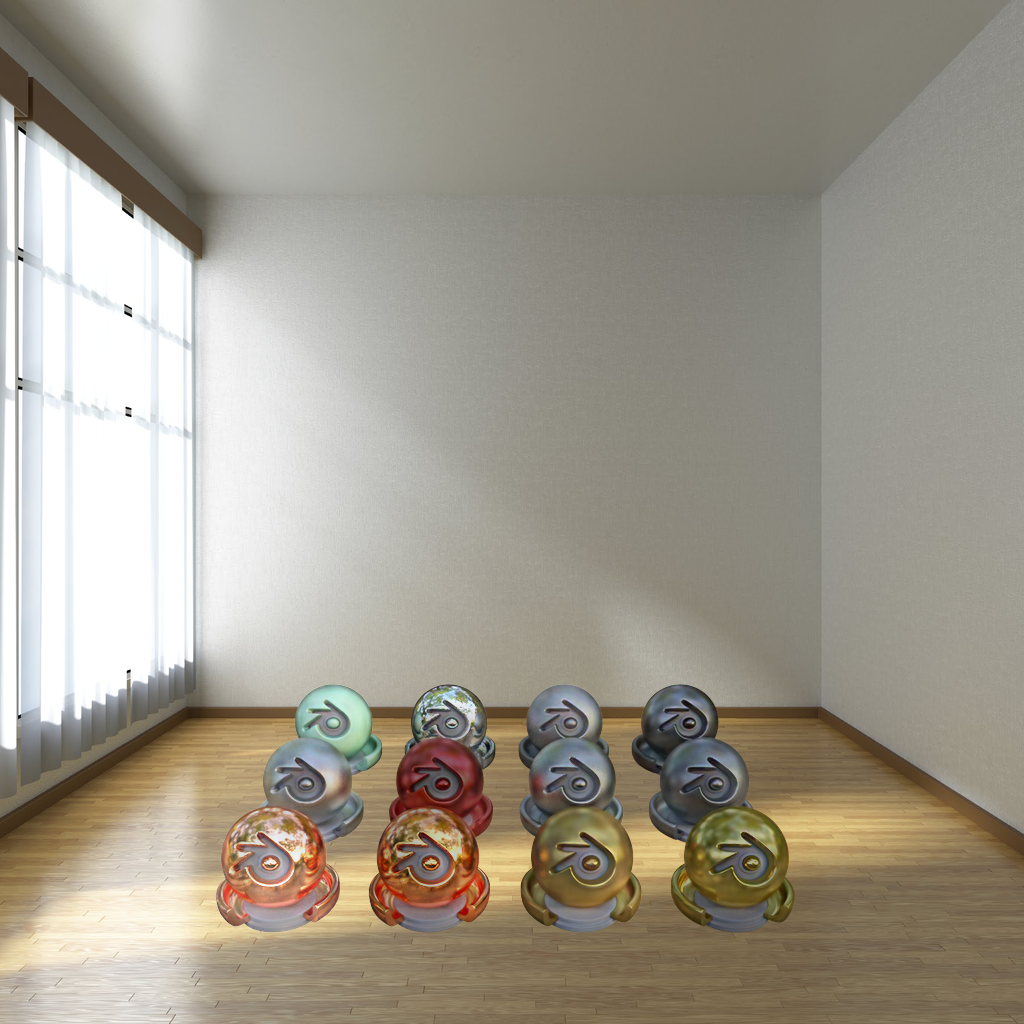}\vspace{2pt}\\
\centering
\small
Target Scene & \centering \small Cut-and-Paste (\textcolor{orange}{CP}) & \centering \small Image Harmonization (\textcolor{red}{IH})~\cite{cong2020dovenet} & \centering \small \textcolor{darkgreen}{DIPR (ours)}
\vspace{-10pt}
\end{tabularx} 
\vspace{-15pt}
}
\includegraphics[width=0.9\linewidth]{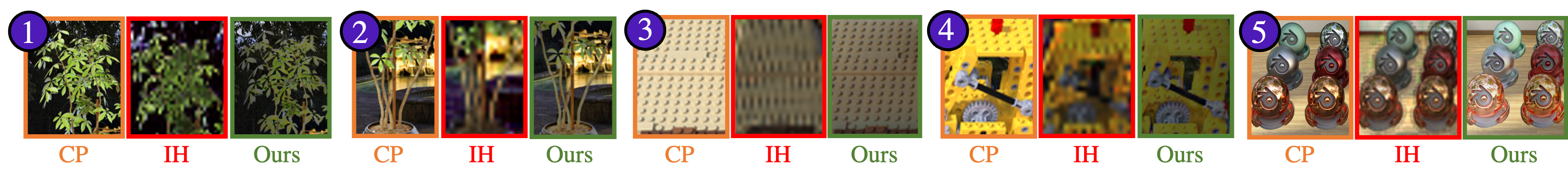}
\vspace{-5pt}
\caption{We synthesize realistic, high-resolution renderings of objects added to scenes -- cut from one image and pasted into another. Our approach, DIPR, is entirely image-based and can convincingly insert objects with complex surface properties (a lego dozer, a plant and a chair in the first row) and matte, glossy and specular objects (a set of 16 different materials in the second row; crop$\#5$ in bottom row) added to spatially varying illuminating scenes (indoor-outdoor, day-night) without requiring the geometry of the inserted objects or the parameters of the target scene. Key findings from our method are (a) it appears to handle complex and subtle interaction with light; for eg., leaves (crop$\#1$) (b) it appears to understand 3D scene and illumination reasonably well (crop$\#2$); CP in crop$\#2$ looks realistic in local context, but it isn't capturing the 3D scene -- the light source behind is far away from the plant, and (c) it preserves material properties because of a carefully designed model; enabling object insertion without any loss of high-frequency details (crop$\#3\ \&\ \#4$).} 
\vspace{-4pt}
\label{fig:overview}
\end{figure*}

Inserting objects into images is an appealingly easy rendering paradigm -- one just moves objects from one image into another. Applications of this task are abundant, ranging from room planners to image editing for artists to training detectors~\cite{liao2012building,dwibedi2017cut}. But most insertions are not realistic because the shading between the inserted object clashes with the target scene and the object sticks out~\cite{lalonde2007photoclipart}. Current state-of-the-art (SOTA) methods recover an environment map and render objects using their geometric and physical model~\cite{li2020inverse,garon2019fast}. Current single-image methods for shape and material recovery cannot accurately reconstruct realistic objects (say, a lego toy)~\cite{li2018learningspat}. Therefore, in this work, we focus on an image-based insertion approach: one takes an object from one image, inserts it into another, and expects a system to correct it. Our method not only synthesizes plausibly realistic renderings of inserted objects with complex surface properties but also does not require a geometric or physical model or an environment map at test time. Furthermore, it does not require rendered images during training.

Our method, DIPR, uses Deep Image Prior~\cite{ulyanov2018deep} for reshading. DIPR adjusts shading so that simple inferences are consistent with cut-and-paste predictions. The rendering process produces an image that is realistic, guaranteed by our use of a deep image prior. The rendering must also produce (a) an albedo that matches the cut-and-paste albedo, (b) a shading that matches the target scene's shading outside the inserted fragment, 
and (c) the rendered shading and the cut-and-paste normals are consistent with each other. The result is a simple procedure that produces convincing shading. Our entire process, including image decomposition, does not require any form of labeled data for training. We use an extension of Retinex\cite{land1977retinex}, build a statistical process for data generation and train variants of image decomposition models for DIPR. In fact, our only use of simulated ground truth is our use of a pre-trained, off-the-shelf, normal estimator~\cite{nekrasov2019real}. 

Our experiments show DIPR convincingly inserts several objects with complex surface properties -- a lego dozer, a plant, a chair and a set of 16 materials with different reflective properties (Fig.~\ref{fig:overview}), cars (Fig.~\ref{fig:cars}) and spherical lampshades (Fig.~\ref{fig:real_spheres}). In qualitative analysis, we show DIPR produces convincing results compared to a SOTA image harmonization baseline~\cite{cong2020dovenet}. We also find that DIPR achieves significantly better PSNR, MSE and LPIPS scores, outperforming a SOTA image harmonization baseline for lampshades renderings (Tab.~\ref{table:quant_metric}). We conduct user studies comparing our renderings against baselines and real images. We also show our method has an implicit notion of 3D shape as an emergent property of our consistent reshading (Fig.~\ref{fig:shape_recons}). 

In summary, {\bf our main contributions are} (1) we enable deep image priors to explicitly reason about the shading of the scene by a new class of image decomposition model. (2) Our method can realistically insert objects without any ground truth labeled data. The only labeled data that our method requires is a surface normal, obtained from a pretrained network. Other than that, our method is completely self-supervised. (3) Our method works for matte, glossy and specular objects with complex surface properties and without using explicit geometric or physical model of the scene. (4) Our method works for diverse (indoor-outdoor, day-night) spatially varying illuminated complex scenes. (5) Our method produces convincing results compared to a SOTA image harmonization baseline and achieves significantly better PSNR, MSE and LPIPS scores. (6) Our method has an implicit notion of 3D shape as an emergent property of our consistent reshading.

\section{Related Work}
\label{relatedwork}
\begin{figure*}[t!]
    \centering
    \vspace{-6pt}
    \includegraphics[width=0.96\linewidth]{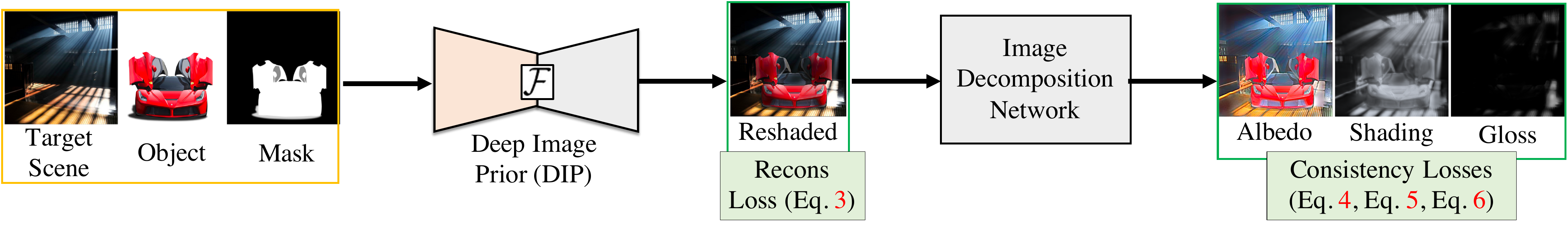}
    \vspace{-3pt}
    \caption{{\bf DIPR overview}. DIPR generates a plausibly realistic rendering of an object inserted from a source image to a target scene. DIPR uses a DIP to generate a reshaded rendering that has consistent image decomposition inferences. The resulting rendering from DIP should have an albedo, same as the cut-and-paste albedo; it should have a shading and gloss field that, outside the inserted fragment, is the same as the target scene’s shading and gloss field. The rendering must have similar spherical harmonic properties as target scene and meet a consistency test everywhere (Sec \ref{subsec:shc}, Fig.~\ref{fig:shc}). This simple procedure inserts objects convincingly in real images. 
    \vspace{-11pt}
    }
    \label{fig:method}
    \end{figure*}
{\bf Object insertion} originated with Lalonde \emph{et al.}~\cite{lalonde2007photoclipart}. They insert objects into target images and control illumination problems by checking objects for compatibility with targets; Bansal \emph{et al.}~\cite{bansal2019shapes}  and Lee \emph{et al.}~\cite{lee2018context} do so by matching contexts. Poisson blending~\cite{perez2003poisson, jia2006drag} can resolve nasty boundary artifacts, but significant illumination and color mismatches will cause cross-talk between target and fragment, producing ugly results. Karsch \emph{et al.}~\cite{karsch2011rendering, karsch2014automatic} how convincing insertions of computer graphics (CG) objects into inverse rendering models.
Inverse rendering trained with rendered images can produce excellent reshading of CG objects~\cite{ramachandran1988perceiving}. 
However, recovering a renderable model from an image fragment is extremely difficult,
particularly if the fragment has an odd surface texture.  Liao {\em et al.}\cite{liao2015cvpr,liao2019IJCV} showed that a weak geometric model of the object can be 
sufficient for correcting shading \emph{if} one has strong the geometric information about the target scene. However, we do not know about geometry of the target scene, except for their normals.

We use {\bf image harmonization} (IH) methods as a strong baseline.  
IH train models to correct corrupted images where a fragment is adjusted by some noise process (made brighter; recolored; etc.) to the original image~\cite{sunkavalli2010multi, tsai2017Deep, cong2020dovenet, ling2021region, guo2021image}, and so could clearly be applied here.  But we find those IH methods very often change the albedo of an inserted object, rather than its shading.  This is because they rely on ensuring consistency of color representations across the image. In contrast, we wish to correct shading alone.

{\noindent \bf Image Relighting:} With appropriate training data, for indoor-scenes, one can predict
multiple spherical harmonic components of illumination~\cite{garon2019fast},
or parametric lighting model~\cite{gardner2019deep} or even full radiance maps
at scene points from images~\cite{song2019neural,srinivasan2020lighthouse}.
For outdoor scenes, the sun's position is predicted in panoramas using a learning-based approach~\cite{hold2019deep}. However, we do not have access to either training data with lighting parameters/environment maps to construct such a radiance field. Recent single-image relighting methods relight portrait faces under directional lighting~\cite{sun2019single, zhou2019deep, nestmeyerlearning}.
Our method can relight matte, gloss and specular objects with complex material properties like cars (Fig.~\ref{fig:cars}) for both indoor and outdoor spatially varying lighting only from a single image and without requiring physics-based BRDF~\cite{li2020inverse}.

Land's Retinex ({\bf image decomposition}) model
assumes effective albedo displays sharp, localized changes (which result in large image gradients), and that shading has small
gradients~\cite{land1959color1,land1959color2,land1977retinex,land1971lightness}. These models require no ground truth.
An alternative is to use CG rendered images for training~\cite{li2018cgintrinsics, bi20151, fan2018revisiting}.  
Current image decomposition evaluation uses the weighted human disagreement rate (WHDR)~\cite{bell2014intrinsic}; current champions are~\cite{fan2018revisiting, forsyth2020intrinsic}.  We use an image decomposition built around approximate statistical models of albedo and shading~\cite{forsyth2020intrinsic} to train our network without requiring real image decompositions. Our method has reasonable, but not SOTA, WHDR; but we show that improvements in WHDR do not result in improvements in reshading (Fig.~\ref{fig:whdr_compare}).

\section{Approach}
\label{sec:method}
DIPR synthesizes a reshaded object transferred from the source image ($s$) into a target scene image ($t$). We use a deep image prior (DIP)~\cite{ulyanov2018deep} as a renderer to produce a reshaded image. We enable DIP to reshade by forcing it to produce consistent image decomposition inferences that meet certain shading consistency tests. We use an image decomposition trained on statistical samples of albedo, shading and gloss; Fig.~\ref{fig:decomp}a and not real images (Sec~\ref{subsec:invrend}), and surface normals inferred by the method of~\cite{nekrasov2019real} to meet the shading consistency tests (Sec~\ref{subsec:shc}). 
The final reshaded image's albedo must be like the cut-and-paste albedo; the reshaded image's shading  must match the shading of the target scene outside the fragment; and  the shading of the reshaded image must have reasonable spherical harmonic properties  and meet a consistency test everywhere
Fig.~\ref{fig:method} summarizes our method.

\subsection{Enabling DIP for object reshading}
Assume we have a noisy image $\imit$, and wish to reconstruct the original.
Write $z$ for a random vector, and $f_\theta$ for a CNN with parameters $\theta$ and
$E(f_\theta(z); \imit)$ for a loss comparing the image $f_\theta(z)$ to $\imit$. 
DIP seeks 
\begin{equation}
  \label{eqn:dip_vanilla}
\hat{\theta}=\mbox{argmin}_\theta E(f_\theta (z); \imit)
\end{equation}
and then reports $f_{\hat{\theta}}(z)$. In this naive setup we find that the DIP always converges to cut-and-paste image. 
This is because the inconsistency we observe in cut-and-paste images are subtle view-dependent lighting effects that are difficult to capture using a simple DIP.

An alternative strategy is to decompose image into two components -- a persistent map, one that is invariant to lighting effects and a transient map, one that changes with lighting. We then have to train a DIP only to make corrections to the extrinsic map and use intrinsic map as it is. 
To this end, we modify Eq.~\ref{eqn:dip_vanilla} by requiring that $E(\cdot; \imit)$ only to adjust the extrinsic properties of the inserted object.  In particular,  write $g_\phi$ for some inference network(s), $t_\psi(\imis, \imit)$ for inferences 
constructed out of $\imit$ and the source image $\imis$.
DIPR seeks 
\begin{equation}
\hat{\theta}=\mbox{argmin}_\theta E(g_{\phi}(f_\theta (z)); t_\psi(\imis, \imit)).
\end{equation}

For us, $g_\phi$ is an image decomposition network, which is pretrained and fixed. 

\begin{figure*}[t!]
  \centering
  \includegraphics[width=\linewidth]{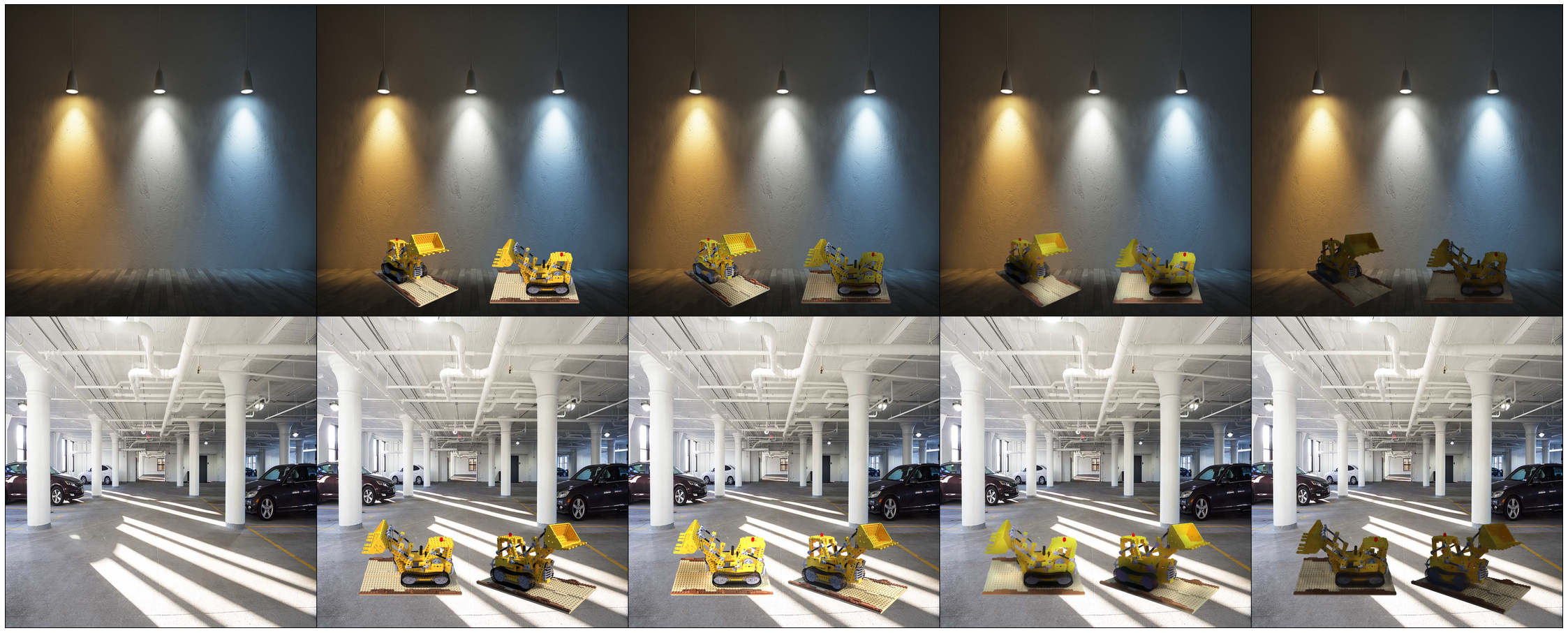}
    \centering
  \resizebox{\linewidth}{!}{
  \begin{tabularx}{\linewidth}{*{5}{>{\vspace{-15pt}\centering\arraybackslash\footnotesize $}X<{$}}} 
    \scalebox{1}{Target \ Scene} & \scalebox{1}{CP} & \scalebox{1}{Our Decomposition} &  \scalebox{1}{Paradigms~\cite{forsyth2020intrinsic}} & \scalebox{1}{CGIntrinsics{\cite{li2018cgintrinsics}}}
    \end{tabularx}
  }
  \vspace{-18pt}
  \caption{{Better WHDR does not mean better reshading. We show reshaded images when using target inferences from different decomposition models. Our decomposition achieves (relatively weak) WHDR of $19\%$; Paradigms~\cite{forsyth2020intrinsic} achieve $17\%$, and a supervised SOTA~\cite{li2018cgintrinsics} achieve $15\%$.  Paradigm ~\cite{forsyth2020intrinsic} decomposition produces worse reshading. Moreover, reshading using a supervised SOTA, CGIntrinsics\cite{li2018cgintrinsics}, is worse than ours and Paradigms decomposition. This reflects that better recovery of albedo, as measured by WHDR, does not produce better reshading. The key issue is that methods that get low WHDR do so by suppressing small spatial details in the albedo field (for example, the surface detail on the lego dozer), and the shading inference method cannot recover these details, and so they do not appear in the final rendering. From the perspective of reshading, it is better to model them as fine detail in albedo than in shading.}
\vspace{-5pt}}
  \label{fig:whdr_compare}
  \end{figure*}

\subsection{Image Decomposition}
\label{subsec:invrend}
  A natural choice for an image decomposition is an albedo map (persistent to lighting) and a shading map (transient to lighting). One could then use a SOTA pretrained image decomposition network as $g_\phi$ only to adjust the shading of the scene and use the cut-and-paste albedo as it is. Next, we train DIP to reshade (DIPR) the inserted object to produce an image with their albedo, same as the cut-and-paste albedo and a shading field, same as the cut-and-paste image only for the background region other than the inserted fragment. DIPR then learns to extrapolate or interpolate shading for the inserted fragment from the background's shading field.
 
\begin{figure}[t]
    \centering
    \includegraphics[width=\linewidth]{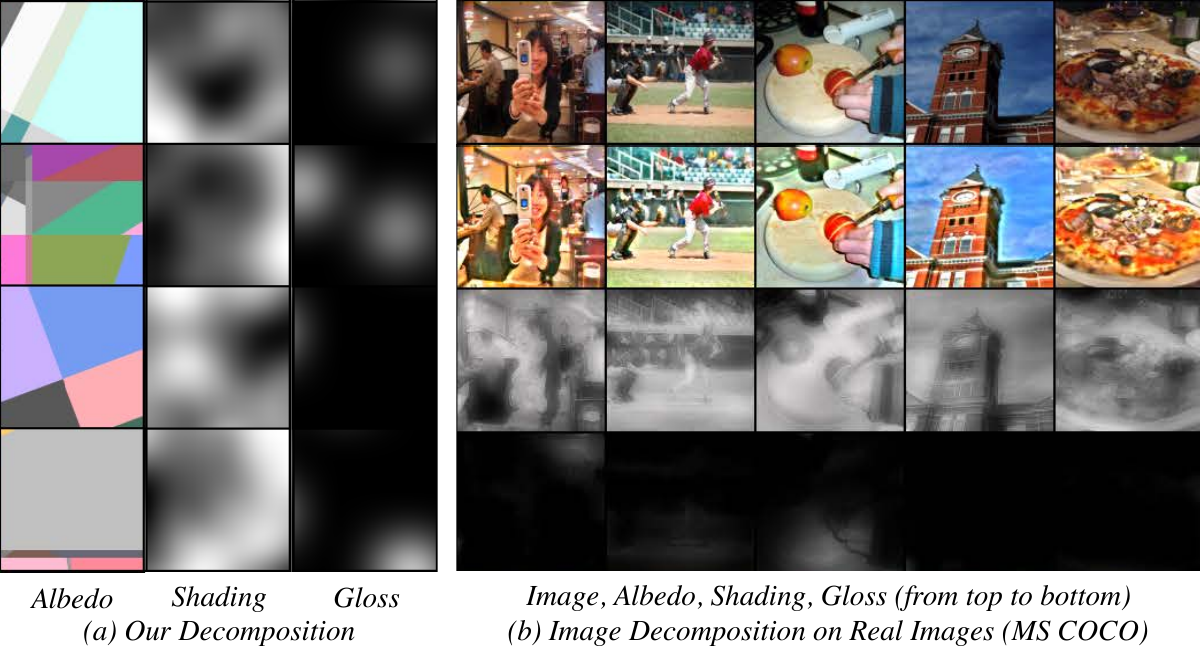}
    \vspace{-15pt}
    \caption{{\bf Image decomposition.} Left: samples of our albedo, shading and gloss used to train our image decomposition network. Right: examples showing MS COCO image decompositions.}
    \label{fig:decomp}
    \vspace{-10pt}
    \end{figure}

  We first evaluated DIPR with two SOTA image decomposition methods (one supervised~\cite{li2018cgintrinsics} and one unsupervised~\cite{forsyth2020intrinsic}) as measured by strong WHDR performance. The supervised decomposition train their models on CG-generated datasets with ground-truth supervision. \cite{forsyth2020intrinsic} uses \emph{Paradigms}, a statistical model of albedo and shading, an extension of the Retinex~\cite{land1977retinex}. Albedo paradigms are Mondrian images. Shading paradigms are Perlin noise~\cite{perlin1985image}. We show these methods result in poor reshading outcomes (Fig ~\ref{fig:whdr_compare}). SOTA albedo-shading decompositions get strong WHDR performance by suppressing fine spatial details in albedo. These methods preserve spatial and geometric details in the shading field and not albedo because they construct albedo as a piece-wise color constant (Mondrian) with no fine details on them. These fine geometric details are hard to recover accurately from a DIP when trained to interpolate the foreground shading field from their background. 
  However,~\cite{forsyth2020intrinsic} offers an interesting feature in their Paradigms construction. The statistical models used are authored. Changing the statistical properties of their models would result in a different class of decompositions. We take advantage of this feature. We change~\cite{forsyth2020intrinsic} and construct an albedo ($A$) -- a persistent map, a diffuse shading ($S$) -- a multiplicative transient map and  a gloss ($G$) -- an aditive transient map using the same statistical process.
  We compose them as $I=A\times S+G$ to form an image and train a network to decompose them back. Fig.~\ref{fig:decomp}a shows samples of our.  Fig.~\ref{fig:decomp}b illustrates the resulting decompositions are satisfactory on MSCOCO~\cite{lin2014microsoft} real images. The main difference between ours and ~\cite{forsyth2020intrinsic} is that we assume shading to be smooth and albedo has all the high-frequency information so that they can be recovered when reshading with a DIP. This is a reasonable assumption for our method because we aim to preserve all the fine-spatial details of inserted fragment when transferring from one image to another. The additional gloss map that we use helps us to extract better lighting representations in scenes with strong lighting effects like shafts. We show our decomposition produces convincing object reshading when compared to other SOTA image decomposition methods (Fig.~\ref{fig:whdr_compare}). 

\subsection{Base Losses}
We first construct the desired target albedo ($A_t$), target shading and gloss ($S_t$ and $G_t$). We then train DIPR to produce an image that has reasonable albedo, shading and gloss properties. For DIPR, the input $z$ is the cut-and-paste image and $f_\theta$ is optimized to inpaint inserted fragment and also to meet satisfactory image decomposition consistency tests. We use U-Net with partial convolution~\cite{liu2018image, shih20203d}. However, we find the standard partial convolution converges to a trivial solution, producing images close to cut-and-paste. To prevent this overfitting, we flip the context for partial convolution. We consider the inserted object(s) as the context and hallucinate/outpaint the entire target scene around it.
We call this {\emph{flipped partial convolution}}. This encourages the network not to overfit to the input cut-and-paste image.

We use $\cp{\imis}{\imit}{s}$ for an operator that cuts the fragment out of the source image 
($\imis$), scales it by $s$,  and places it in the relevant location in the target image ($\imit$).  
$\mask$ for a mask with the size of the target image that is $0$ inside the fragment and $1$ outside. Our reconstruction loss for the background is:
\begin{equation}
  \small
\mathcal{L}_{recons} = \sqnorm{\imit{\odot}\mask- (f_\theta(\cp{\imis}{\imit}{s};\mask))}
\end{equation}
We then pass the DIP rendered image through the image decomposition network $g_\phi$ making 
$A_{r}$, $S_{r}$ and $G_{r}$ for the rendered albedo, shading and gloss maps respectively. 
Our consistent image decomposition inference losses to train DIPR are:
{\small
\begin{multline}
  \mathcal{L}_{decomp} = \sqnorm{A_{\cp{\imis}{\imit}{s}}-A_{r}}  + \sqnorm{S_t{\odot}\mask-S_{r}{\odot}\mask} \\ + \sqnorm{G_t{\odot}\mask-G_{r}{\odot}\mask}  
  \label{eqn:decomp}
\end{multline}
}
\vspace{-15pt}
\subsection{Normal Consistency Losses}
\label{subsec:shc}
\begin{figure}[t]
  \centering
  \includegraphics[width=\linewidth]{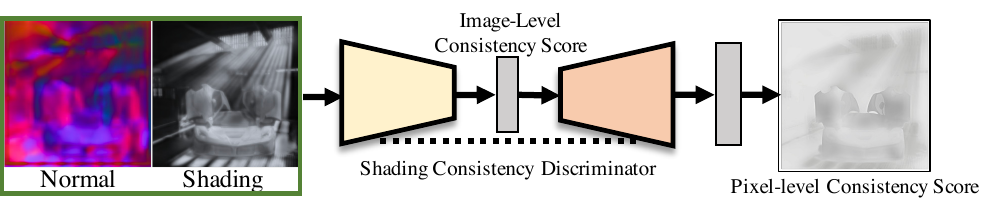}
  \caption{{\bf Shading consistency discriminator} penalizes shading, if it is not consistent with the cut-and-paste normals.}
  \label{fig:shc}
  \end{figure}

We use two normal consistency losses to make the strong structure of a shading field apparent to a DIP's reshading.  There is good evidence that shading (image extrinsic)is tied across surface normals (this underlies 
spherical harmonic models~\cite{li2018learningspat,yu2019inverserendernet}), 
and one should think of a surface normal as a latent variable that explains shading or extrinsic similarities. We assume the resulting illumination approximated with the first 9 spherical harmonics basis coefficients ($Y$) and does not change when an object is inserted into a scene. We get $Y$ by solving the least square regression between normals
(${N}$) and shading ($S$) for both the target scene and the resulting composite image. We then minimize loss ($\mathcal{L}_{Y}$) between the target and rendered image $Y$($S;{N}$) and use  a Huber loss.
\begin{equation} 
  \centering
  \small
  \mathcal{L}_{Y}={\begin{cases}
    {\frac  {1}{2}}({Y}(S_t;{{N}}_{t})-{Y}(S_{r};{{N}_{CP}}))^{2}
    &{\textrm {for}}|Y_t-Y_r|\leq 1 ,
    \\|{Y}(S_t;{{N}}_{t})-{Y}(S_{r};{{N}_{CP}})|-{\frac  {1}{2}}
    &{\textrm  {otherwise.}}
  \end{cases}}
  \label{eqn:spherical_harmonics}
\end{equation}
Spherical harmonic shading fields have some disadvantages: every point with the same normal must have the same shading value, 
which results in poor models of (say) indoor shading on walls. 
To control this effect, we use a novel neural shading consistency loss ($\mathcal{L}_{Z}$) that allows the shading field to depart from a spherical harmonic shading field, but only in ways consistent with past inferences. Our shading consistency discriminator, $\zeta(S;{N})$, is a U-Net~\cite{schonfeld2020u} (Fig.~\ref{fig:shc}); trained to discriminate real and fake shading-normal pairs. $\zeta(S;{N})$ produces two outputs: one a pixel-level map, yielding the first loss term in Eq.~\ref{eqn:nshc}, which measures per-pixel consistency; the other an image-level value, the second term in Eq.~\ref{eqn:nshc}, which measures consistency for the entire image.  The $\mathcal{L}_{Z}$ loss is a binary cross-entropy loss. 
Let $m \times n$ be the resolution of our renderings, then $\mathcal{L}_{Z}$ is given by 
\begin{equation}
  \label{eqn:nshc}
  \small
  \centering
  \mathcal{L}_{Z} = -\sum_{i=1}^{m}\sum_{j=1}^n\log\zeta(S_{r}[i, j]; {N}_{{C}{P}}[i,j]) - log\zeta(S_{r}{{N}_{CP}})  
\end{equation}
In summary, we update DIPR with
\begin{equation}
  \small
  \centering
  \mathcal{L}_{r} = \mathcal{L}_{recons} + \mathcal{L}_{decomp} + \mathcal{L}_{Y} + \mathcal{L}_{Z}
\end{equation}

\section{Experiments}
\label{sec:experiments}
\begin{figure*}[t]
  \centering
  \resizebox{\linewidth}{!}{
  \setkeys{Gin}{width=\linewidth}
  \renewcommand\arraystretch{0}
  \centering
  \begin{tabularx}{\linewidth}{@{}X@{}X@{}X@{}X@{}}
      \includegraphics{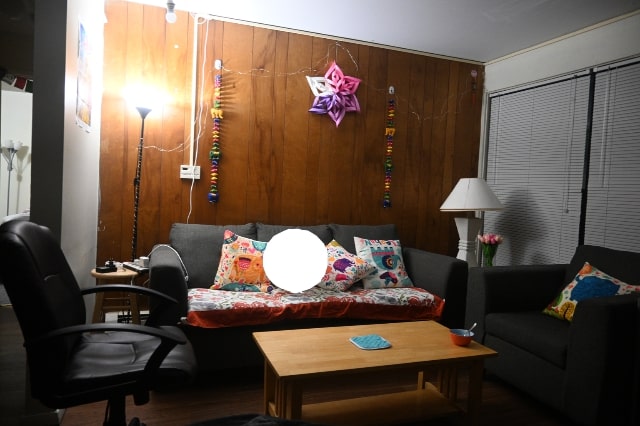} &
      \includegraphics{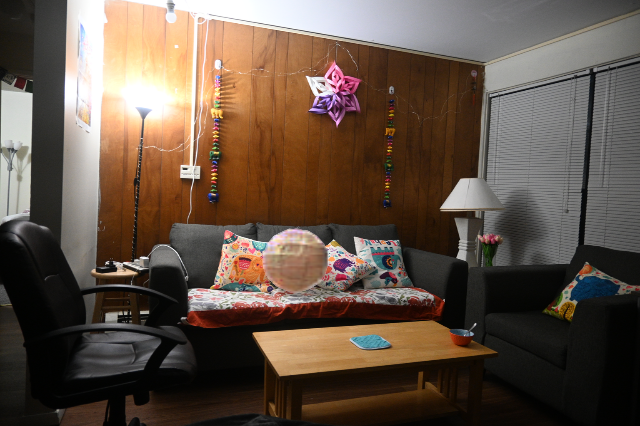} & 
      \includegraphics{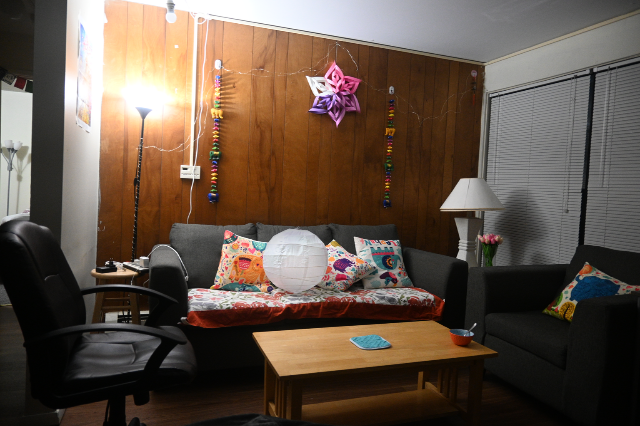} & 
      \includegraphics{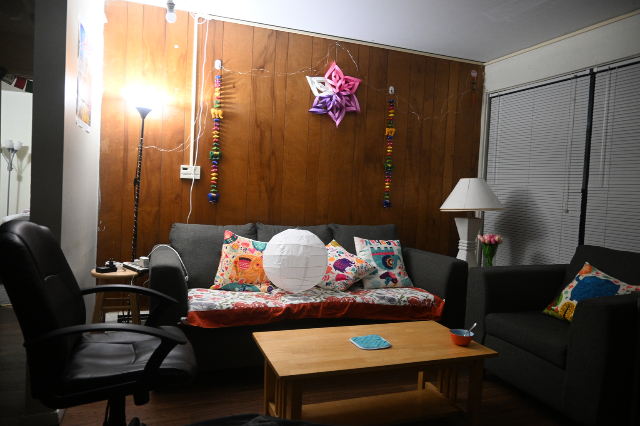}
      \\
      \includegraphics{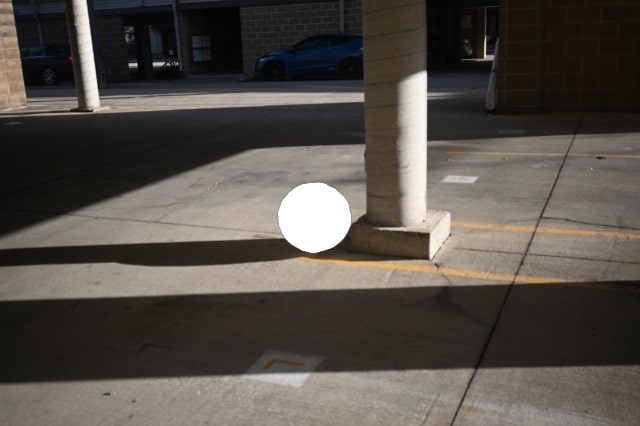} &
      \includegraphics{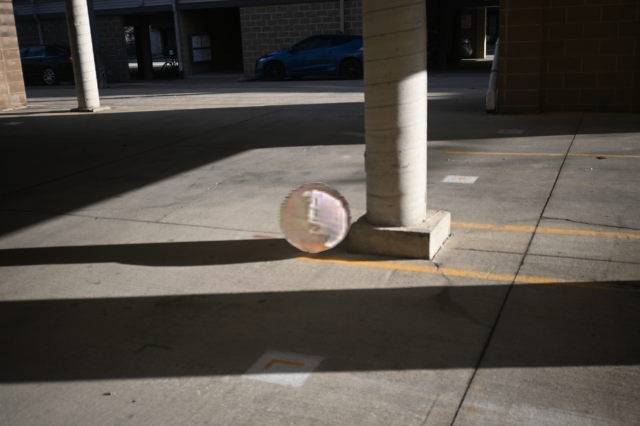} & 
      \includegraphics{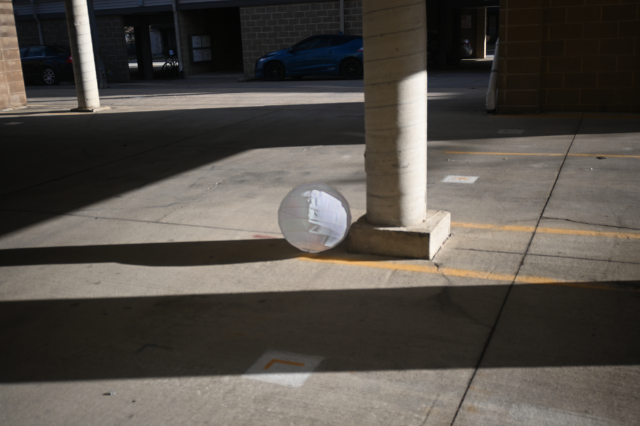} & 
      \includegraphics{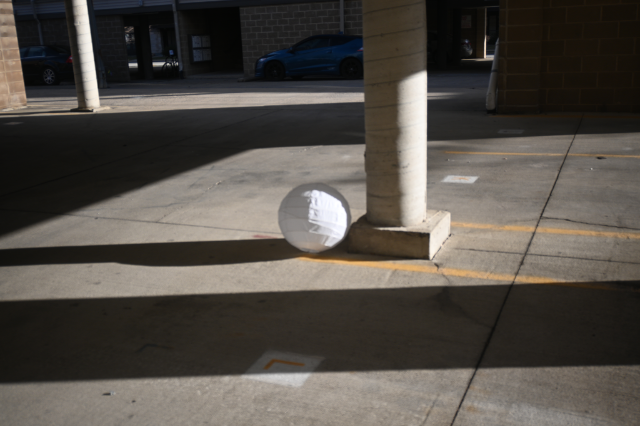}\vspace{2pt}\\
        \centering \scalebox{1}{CP} &\centering \scalebox{1}{IH~\cite{cong2020dovenet}} &\centering \scalebox{1}{DIPR (ours)} &\centering \scalebox{1}{Real Scene}
      \end{tabularx}
  }\vspace{2pt}
  \caption{{\bf Reshading spherical lampshades.} Similar to Fig.~\ref{fig:spheres}, we generate sphere rendering by overlaying 2D white disks over the real sphere from the photograph. We then use DIPR to reshade the overlaid 2D disk and combine that with the actual albedo (as they remain same irrespective of lighting) into a final rendering. IH copies average color. Our renderings are close to the ground truth (also see Tab.~\ref{table:quant_metric}).}
  \label{fig:real_spheres}
  \vspace{-10pt}
\end{figure*}

{\noindent \bf Scenes and objects.} We collected about $100$ diverse images with
spatially varying illumination, both indoors-outdoors, day-night, to act as target scenes in our experiments. $25$ of these scenes are captured by placing a spherical lampshade, which we use as ground truth for quantitative evaluation. We first test DIPR by inserting simple 2D bright disks at various locations in target scenes (Fig.~\ref{fig:spheres}). 
We also show reshading results for real cars inserted into our target scenes (Fig.~\ref{fig:cars}). We also tested DIPR with complex surface properties -- a lego dozer, a plant, a chair, and a set of sixteen materials with different reflective properties used in NeRF~\cite{mildenhall2020nerf} (Fig.~\ref{fig:overview}). Other objects from~\cite{li2018learningspat} are in our Appendix.
We use ADE20K validation set \cite{zhou2017scene} for supplying real residual loss to our image decomposition network and 
also to train our shading consistency network. ADE20K does not have ground truth normals and we use normals from~\cite{nekrasov2019real}.

{ \noindent \bf Network architecture.} 
  Our DIP network is very similar to the U-Net used in 3DPhotoInpainting~\cite{shih20203d}.
  We use their provided model in our implementation as DIP.
  The only difference is our use of flipped partial convolution instead of  standard partial convolution as described in our Sec~\ref{sec:method}.

  {\noindent \bf Training details.} We used U-Net for DIP, Image Decomposition and Shading Consistency Network. 
Network architecture and other training details are in our supplementary. We update our DIP for a fixed $10$k iterations and this takes about $900$ seconds using our image decomposition network 
and $1600$ seconds when using CGIntrinsic~\cite{li2018cgintrinsics}.

{\noindent {\bf IH baseline.} We use Cong \emph{et al.}'s DoveNet~\cite{cong2020dovenet} and their provided pretrained model for our IH baseline.

{\noindent \bf Quantitative comparison to ground truth.}
We compare 25 photographs of a spherical
lampshade with DIPR based insertions. We get insertions
by placing a white disk over the lampshade, reshading
the disk using DIPR, then multiplying by lampshade's albedo
(a favorable case as it correctly rendered cast shadows). We then quantitatively compare results with ground
truth (the real lampshade) using PSNR, LPIPS or MSE (Tab.~\ref{table:quant_metric}; Fig.~\ref{fig:teaser}, Fig.~\ref{fig:real_spheres}). Fig.~\ref{fig:teaser} (b) and (c) are real.

\begin{table}[t]
  \caption{
    {\bf Quantitative Evaluation on Spherical Lampshades} (Fig.~\ref{fig:real_spheres}). Cases: fixed shading fields ($S$ is constant);  fixed albedo fields ($A$ constant); cut-and-paste albedo fields ($A_{\cp{\imis}{\imit}{s}}$); image harmonization (IH); and our reconstruction ($S_r$). 
    Note DIPR reshading wins in all metrics.  Note such comparisons to ground truth occur in circumstances favorable to a method like ours (because the shading around the object is consistent), but we know of no way to avoid this. 
    }
  \label{table:quant_metric}
  \centering
  \vspace{2pt}
  \resizebox{\linewidth}{!}
{
  \begin{tabular}{lccc}
  \toprule
   Shading ($S$) &  Albedo ($A$) & LPIPS\cite{zhang2018perceptual} $\downarrow$ & PSNR $\uparrow$  \\
  \midrule
  $1$ & $1$ & $0.0105$ & $30.81$ \\ %
  $S{_r}$  & $1$ & ${0.0070}$ & ${34.26}$ \\
  \midrule
  DoveNet~\cite{cong2020dovenet}  & $A_{\cp{\imis}{\imit}{s}}$ & ${0.0072}$ & ${35.57}$ \\
  RainNet~\cite{ling2021region}  & $A_{\cp{\imis}{\imit}{s}}$ & ${0.0070}$ & ${36.10}$ \\
  Harmony Transformer~\cite{guo2021image}  & $A_{\cp{\imis}{\imit}{s}}$ & ${0.0069}$ & ${36.52}$ \\
  \midrule
  $0.25$ &$A_{\cp{\imis}{\imit}{s}}$ & $0.0113$ & $26.74$ \\
  $0.50$ &$A_{\cp{\imis}{\imit}{s}}$ & $0.0060$ & $31.28$ \\
  $0.75$ &$A_{\cp{\imis}{\imit}{s}}$ & 0.$0032$ & $38.18$ \\
  $1.0$ &$A_{\cp{\imis}{\imit}{s}}$ & 0.$0043$ & $36.66$ \\
  \midrule
  $S_r$ (ours)  &$A_{\cp{\imis}{\imit}{s}}$ & $\bf{0.0021}$ & $\bf{39.53}$ \\
   \bottomrule
  \end{tabular}
  }
  \vspace{-5pt}
\end{table}

\begin{table}[t]
  \caption{{\bf Ablation} over losses show each helps improve reshading.
  }
\label{table:ablation}
\vspace{2pt}
  \centering
  \resizebox{\linewidth}{!}
  {
  \begin{tabular}{ccccc}
  \toprule
  ${\cal L}_{decomp}$(Eq.~\ref{eqn:decomp})      & $\cal{L}_{Y}$ (Eq.~\ref{eqn:spherical_harmonics})& $\cal{L}_{Z}$(Eq.~\ref{eqn:nshc}) & LPIPS\cite{zhang2018perceptual} $\downarrow$ & PSNR $\uparrow$ \\
  \midrule
   \cmark &  &  &  $0.0026$ & $37.89$ \\
  \cmark & & \cmark  & ${0.0023}$ & ${38.74}$\\ %
  \cmark & \cmark & & ${0.0022}$ & ${39.09}$ \\
  \cmark & \cmark & \cmark & $\bf{0.0021}$ & $\bf{39.53}$ \\
  \bottomrule
  \end{tabular}
  }
  \vspace{-15pt}
\end{table}

{\noindent \bf Fooling users.} The gold standard evaluation here is user studies (after all, the goal is to fool people).  However, user studies are a poor
way to polish a method, and a proxy would be valuable.  For images of the spherical lampshade used in the quantitative evaluation of
Tab.~\ref{table:quant_metric}, we asked users to identify which of the two presented images (showing distinct scenes, but both containing the lampshade) was real.  One image
presented was always a rendering, the other always real.  Users each see
16 pairs, and there were $72$ participants performing this study. Our DIPR renderings are very good at fooling users ($50\%$ is a chance). They pick DIPR $42.7\%$ of the time. We then use logistic regression to predict each rendered image's probability of being marked real against MSE, PSNR and LPIPS.  An accurate regression would mean that we had a score of ``realness". However, such a model explains almost none of the variation of the data (null deviance: $17.87$, residual deviance: $17$).
This means that, while it is pleasant that DIPR has strong MSE, PSNR, and LPIPS scores, these scores can not be used to predict user preferences for our task.

For other object when we did not have ground truth for objects like cars, lego dozer, plant and chair, we conducted another user study to compare DIPR, CP, and IH renderings. Each study comprises a pre-qualifying process, followed by $9$ pair-wise comparisons, where the user is asked which of two images are more realistic. The result is $109$ pre-qualified studies. The comparisons are balanced. Each study is $3$ DIPR-IH pairs, $3$ CP-IH pairs, and $3$ DIPR-CP pairs, in random order.

We collected data from a total of 122 unique users in 500 studies from Amazon Mechanical Turk.  Each study consists of a prequalifying process,
followed by 9 pair-wise comparisons, where the user is asked which of two images are more realistic.  The prequalifying
process presents the user with five tests; each consists of an image with inserted white spheres which are not reshaded (i.e. bright
white disks) and an image with inserted spheres which have been reshaded (see Fig \ref{fig:spheres}).  We ignore any study where the user does not correctly identify all five reshaded images, on the grounds that the difference is very obvious and the user must not have
been paying attention.

The simplest analysis strongly supports DIPR is preferred over both alternatives.  One compares the probability that DIPR is preferred to IH ($.673$, over $327$ comparisons, so standard error is $.026$, and the difference from $0.5$ is clearly significant);  DIPR is preferred to CP ($.645$, over 327 comparisons, so the standard error is $.026$, and the difference from $0.5$ is clearly significant);  IH is preferred to CP ($.511$, over $327$ comparisons, so standard error is $.027$, and there is no significant difference from $0.5$). An alternative is a Bradley-Terry model~\cite{tsai2017Deep, cong2020dovenet} used in IH evaluation, regressing the quality predicted by the Bradley-Terry model against the class of algorithm. This yields coefficients of $0$ for IH, $-0.347$ for CP, and $0.039$ for DIPR, implying again that DIPR is preferred over IH and strongly preferred over CP.

\begin{figure}[t]
  \setkeys{Gin}{width=0.98\linewidth}
  \renewcommand\arraystretch{0}
  \centering
  \resizebox{\linewidth}{!}{
    \centering
  \begin{tabularx}{\textwidth}{@{}X@{}X@{}X@{}X@{}}
    \includegraphics{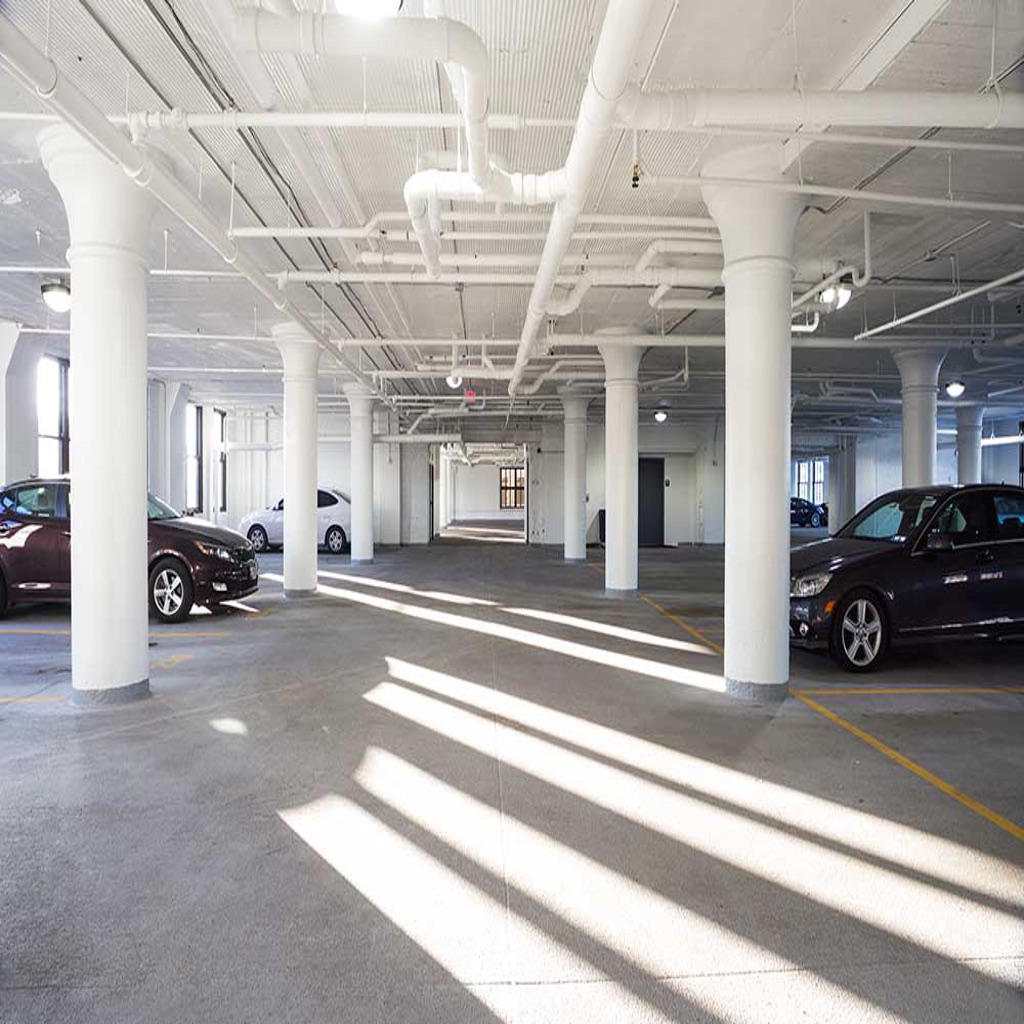} &
    \includegraphics{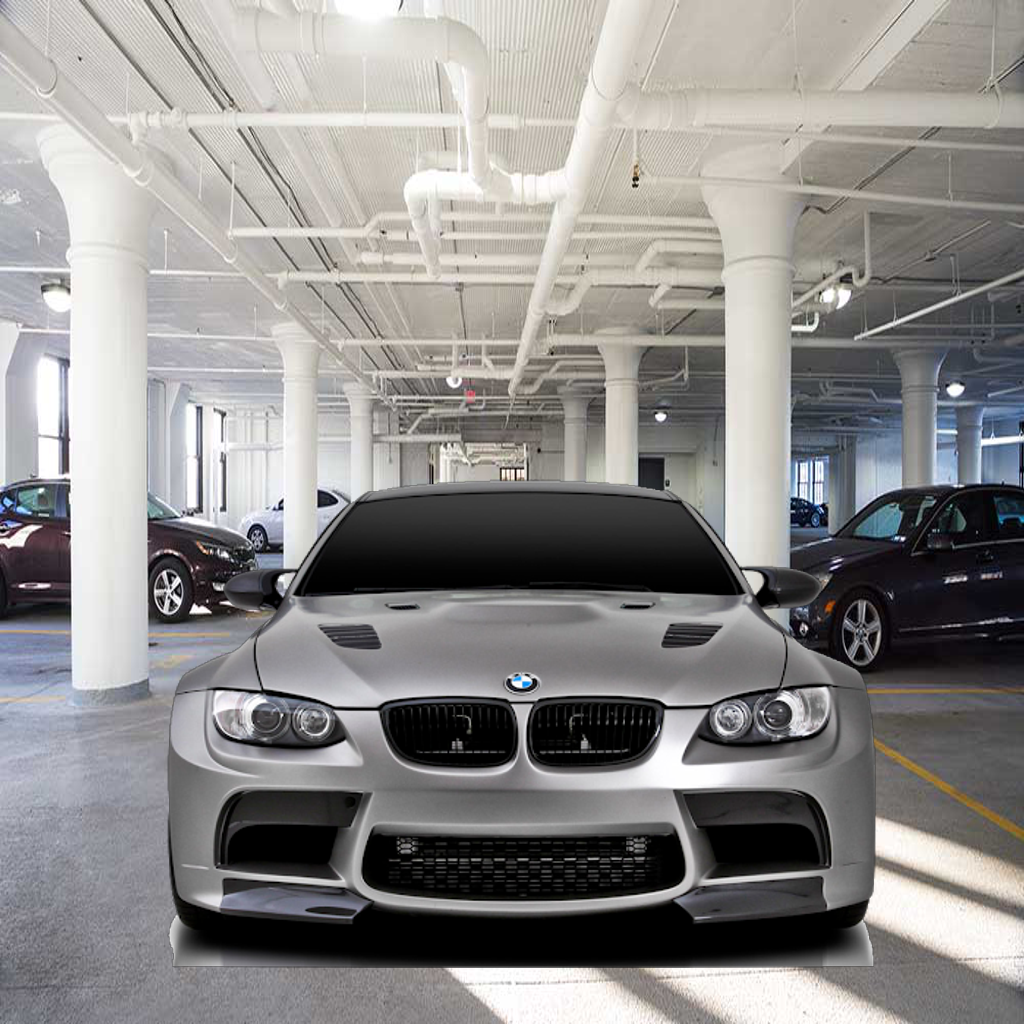} & 
    \includegraphics{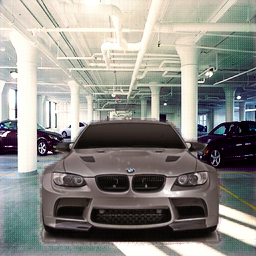} & 
    \includegraphics{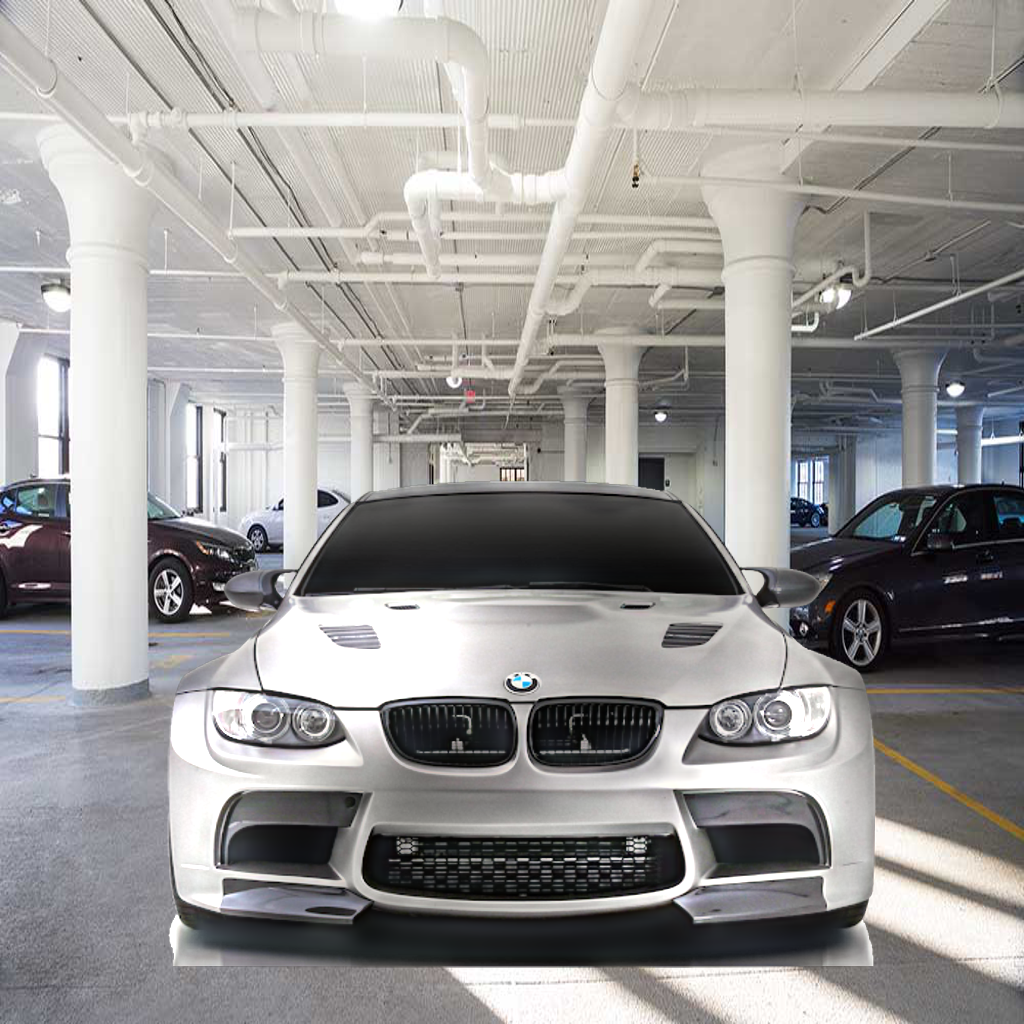} \\ 
    \includegraphics{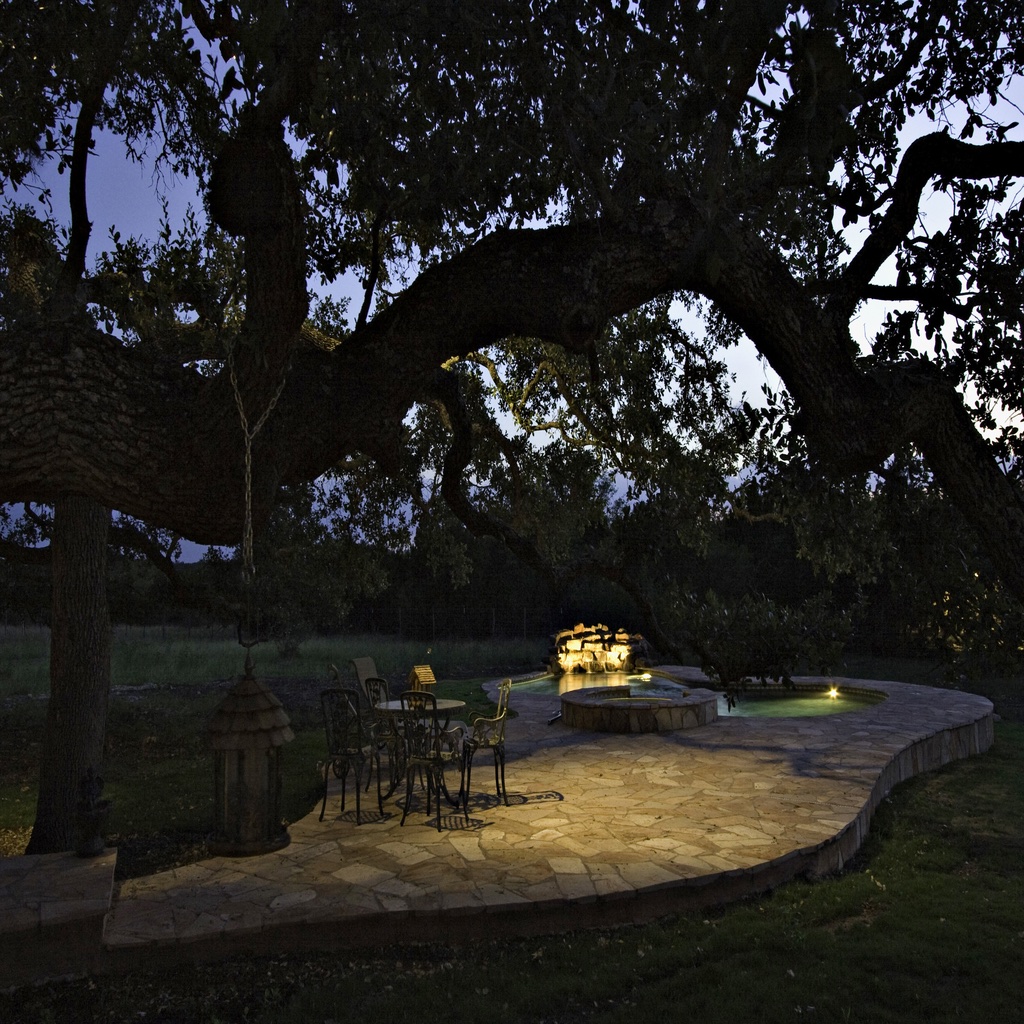} 
     & 
    \includegraphics{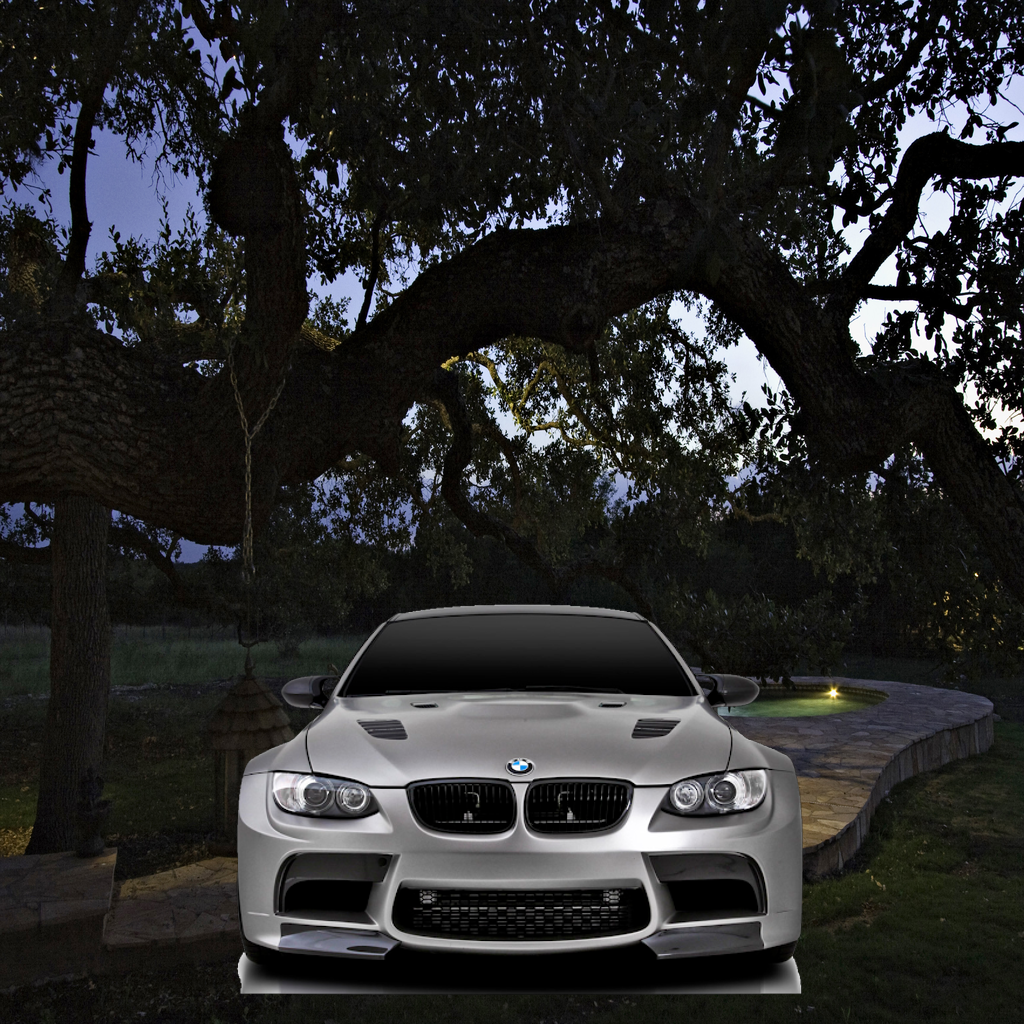}&
    \includegraphics{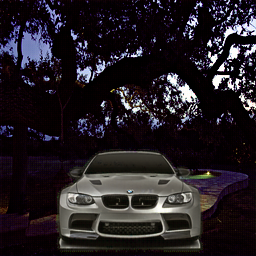}&
    \includegraphics{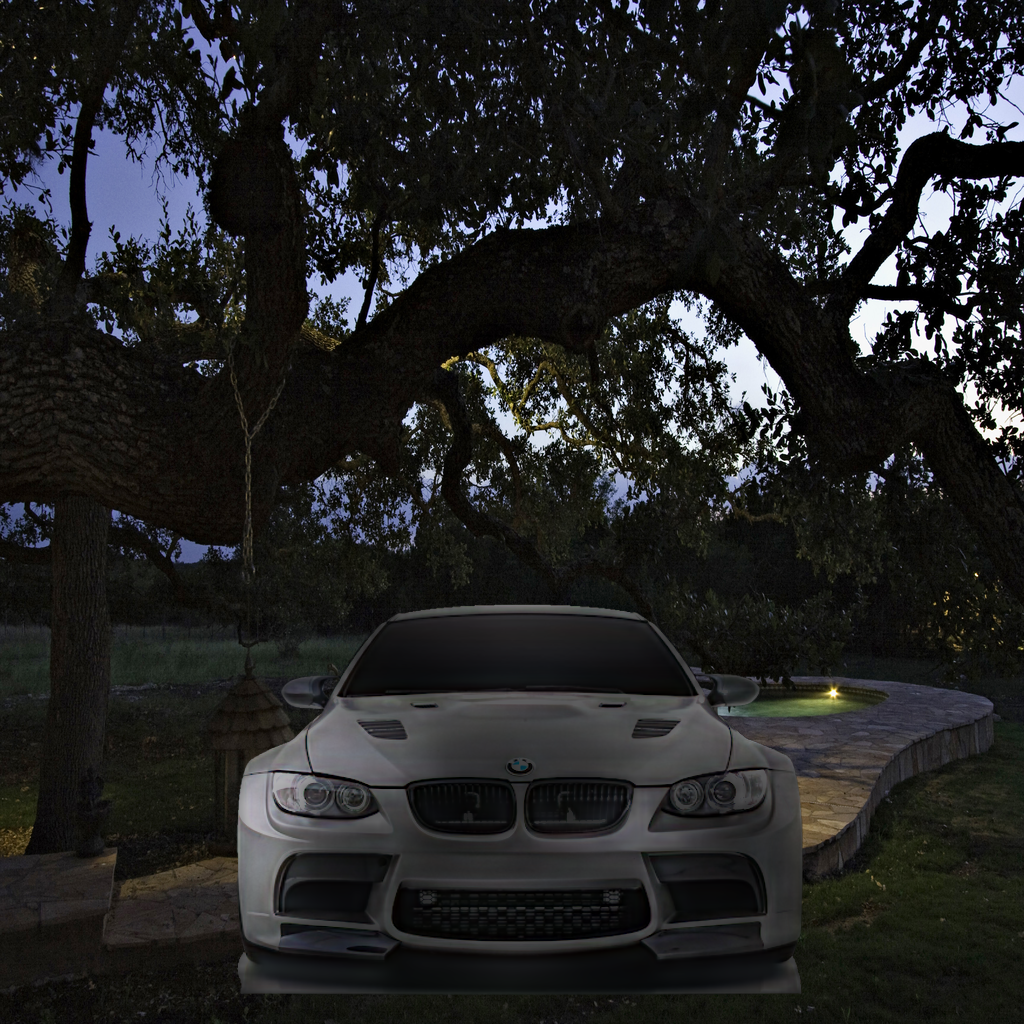}\vspace{3pt}
    \\ 
  \centering\scalebox{2}{Target Scene} & \centering \scalebox{2}{CP} &\centering\scalebox{2}{IH~\cite{cong2020dovenet}}&\centering \scalebox{2}{DIPR (ours)} 
  \end{tabularx}
  }
  \caption{{\bf Reshading real cars.} Glossy effects in car paint with glitter in them make reshading cars a particularly challenging case with their complex reflective properties. DIPR
  successfully reshades cars without a distinct shift in object and background color produced by IH. Note the bright patch on the metallic bonnet of the grey car in the top row that may possibly be because of the light source just above it.
  }
  \label{fig:cars}
  \vspace{-15pt}
\end{figure}
\begin{figure}[t!]
  \includegraphics[width=0.95\linewidth]{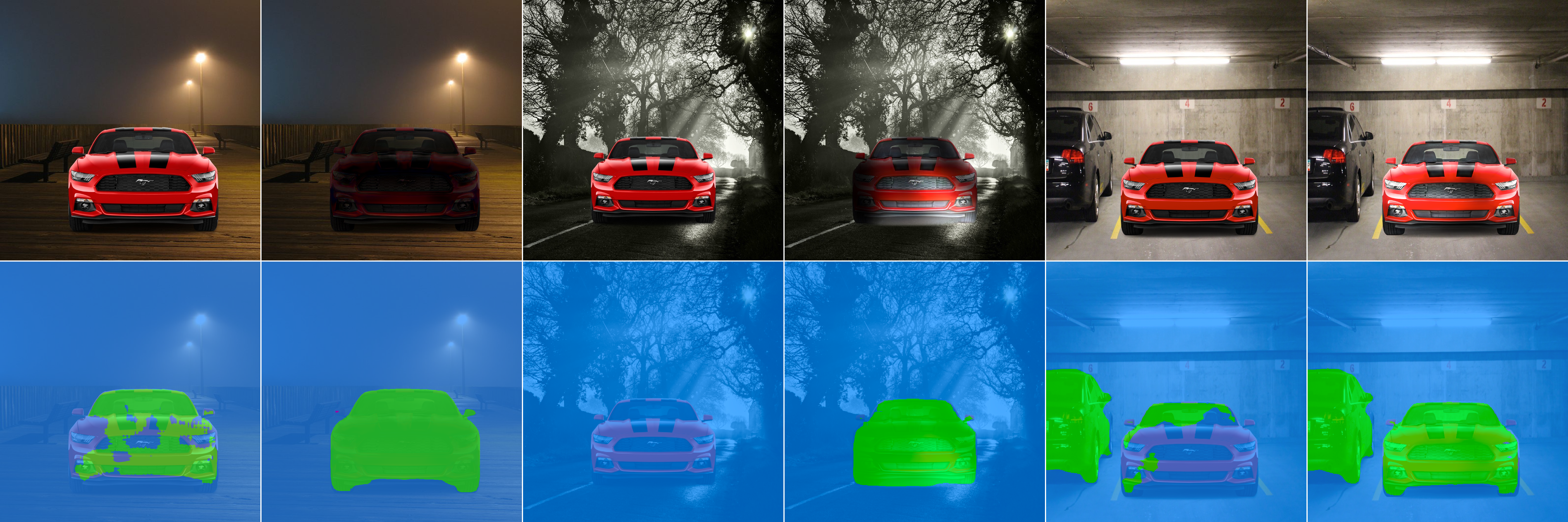}\vspace{-3pt}
  \\ 
  \centering
  \begin{tabularx}{\linewidth}{*{6}{>{\centering\arraybackslash\footnotesize $}X<{$}}}
   \scalebox{1.0}{CP} & \scalebox{1.0}{DIPR} & \scalebox{1.0}{CP} & \scalebox{1.0}{DIPR} & \scalebox{1.0}{CP} & \scalebox{1.0}{DIPR}
  \end{tabularx}
  \caption{{\bf Instance segmentation.}
  DIPR rendered images produce consistent and accurate segmentation maps (bottom row). We observe segmentation fails for cut-and-paste images often. Our key intuition is that the instance segmentation network inherently ``knows" if the object's placement is natural or not and expects the foreground object to have consistent shading with the background. If the object's shading does not match with that of the background then the resulting segmentation fails to segment object completely. \vspace{-2pt}}
\label{fig:consistency}
\vspace{-5pt}
\end{figure}
{\noindent \bf Consistent Instance Segmentation.} We cannot quantitatively evaluate our reshading method when we do not know the ground truth.  But we can test whether standard image tasks (which likely benefit from structural consistency in images) perform better on our images.  We observe image segmentation methods (we used~\cite{chen2017rethinking} %
) seem to prefer our images
(Fig~\ref{fig:consistency}) compared to the naïve cut-and-paste. We believe the instance segmentation network inherently ``knows" if the object's placement is natural or not and hence requires the foreground object to be consistent with the background. If not, the segmentation would produce inconsistent results. Our findings are also consistent with Ghiasi \emph{et al.}\cite{ghiasi2020simple}, who show cut-and-paste is a strong data augmentation method for the instance segmentation. Previous models trained without this augmentation, such as ~\cite{chen2017rethinking} are sensitive to cut-and-paste images if the inserted fragments contradict the background's shading. This suggests downstream standard image analysis tasks can serve as a proxy evaluation to further polish reshading methods and also DIPR data augmentations could further improve various recognition tasks.

  \section{Shape from Shading and its Consistency}
  \begin{figure}[t]
    \setkeys{Gin}{width=0.98\linewidth}
    \renewcommand\arraystretch{0}
    \begin{tabularx}{\linewidth}{@{}X@{}X@{}X@{}X@{}}
      \includegraphics{Figures/spheres/sofa_living_room.jpg} &
      \includegraphics{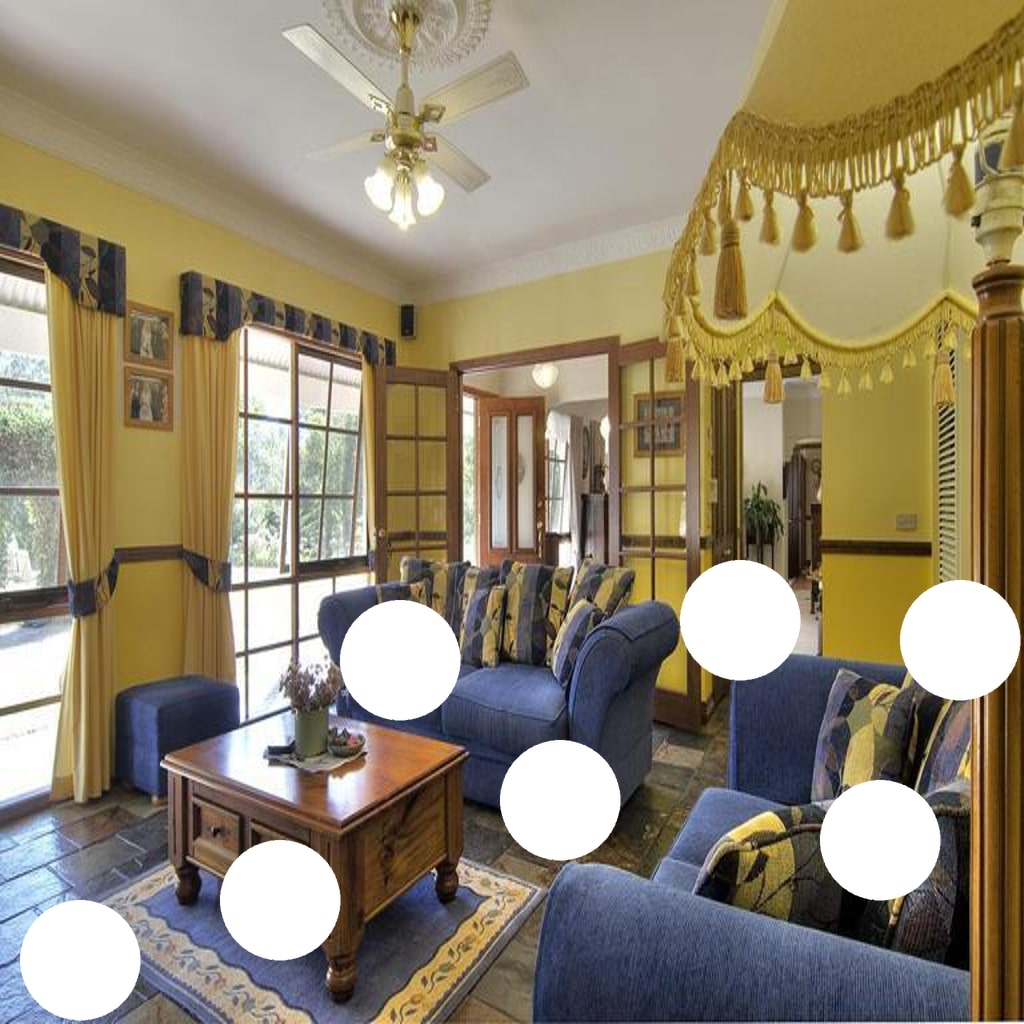} & 
      \includegraphics{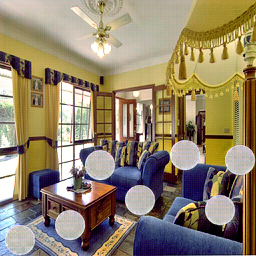} & 
      \includegraphics{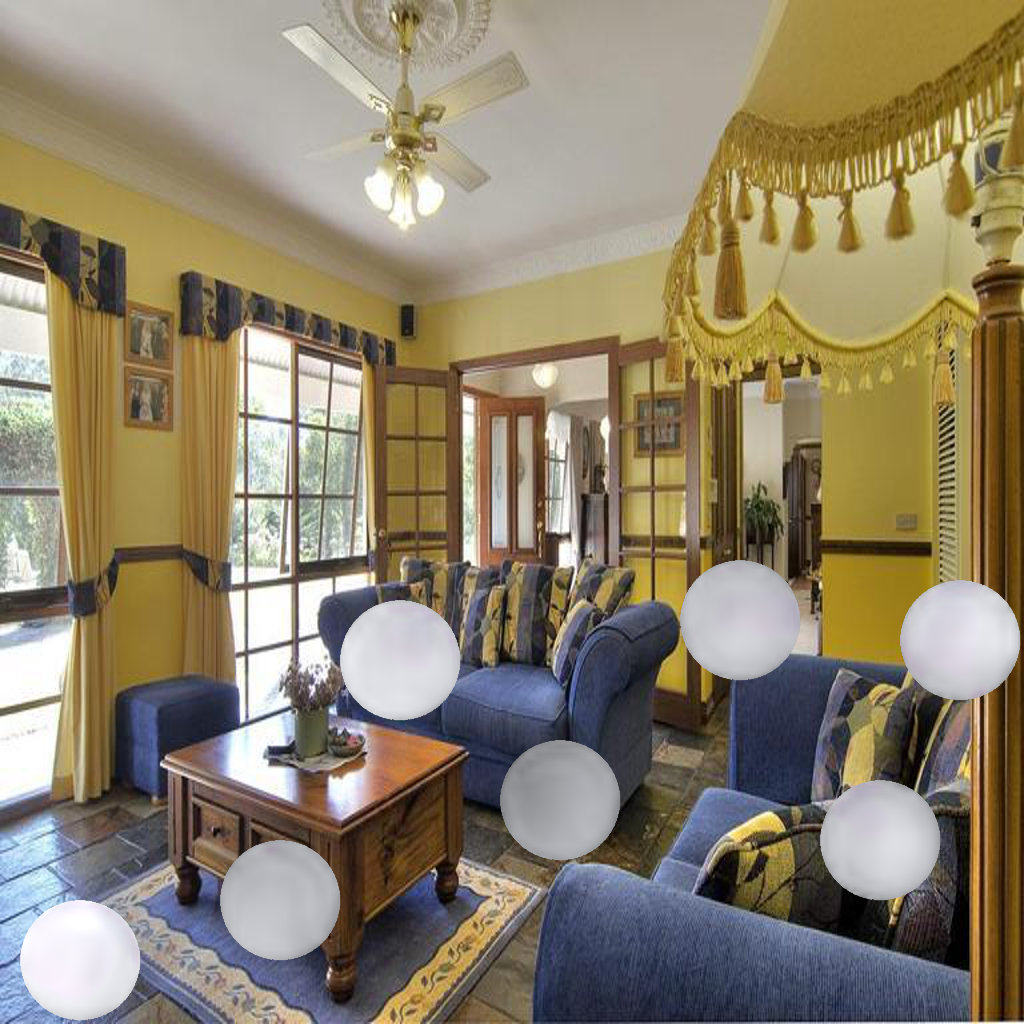}\\
  \includegraphics{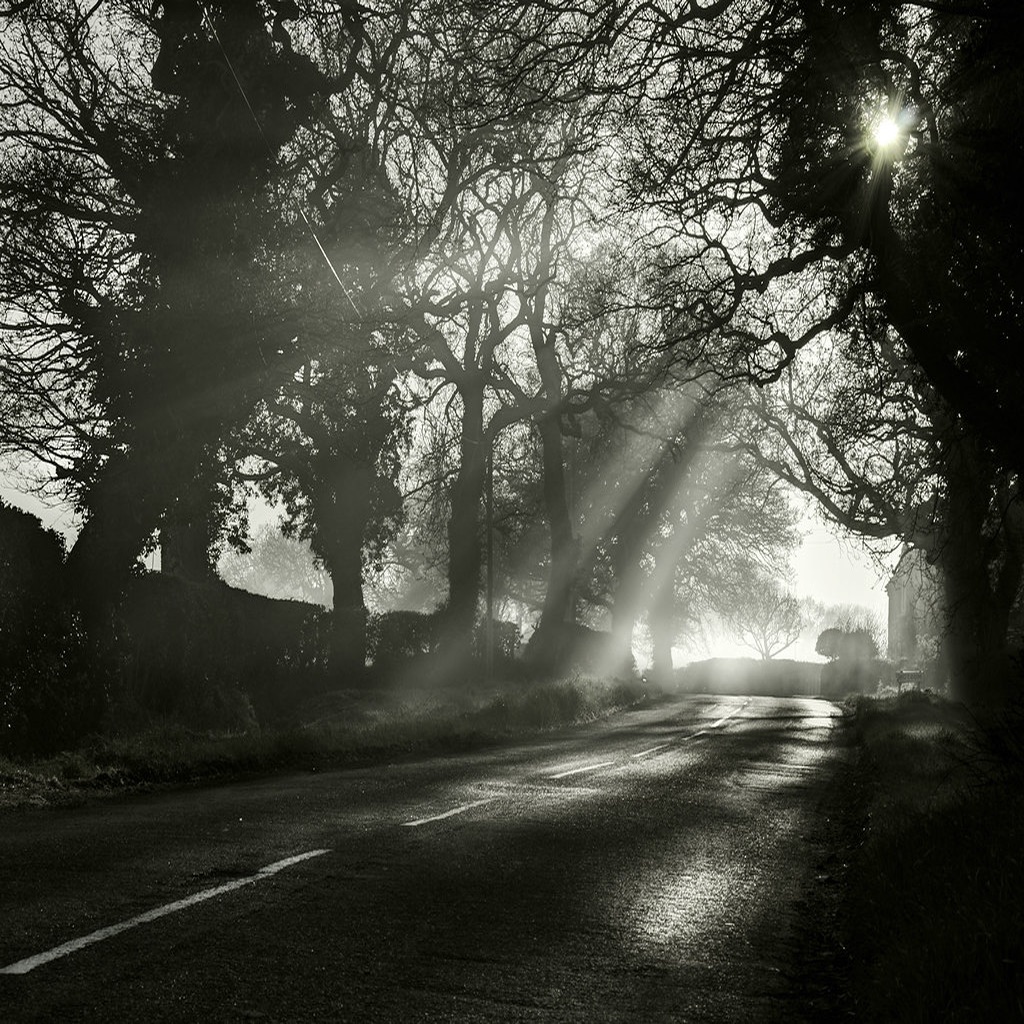} &
  \includegraphics{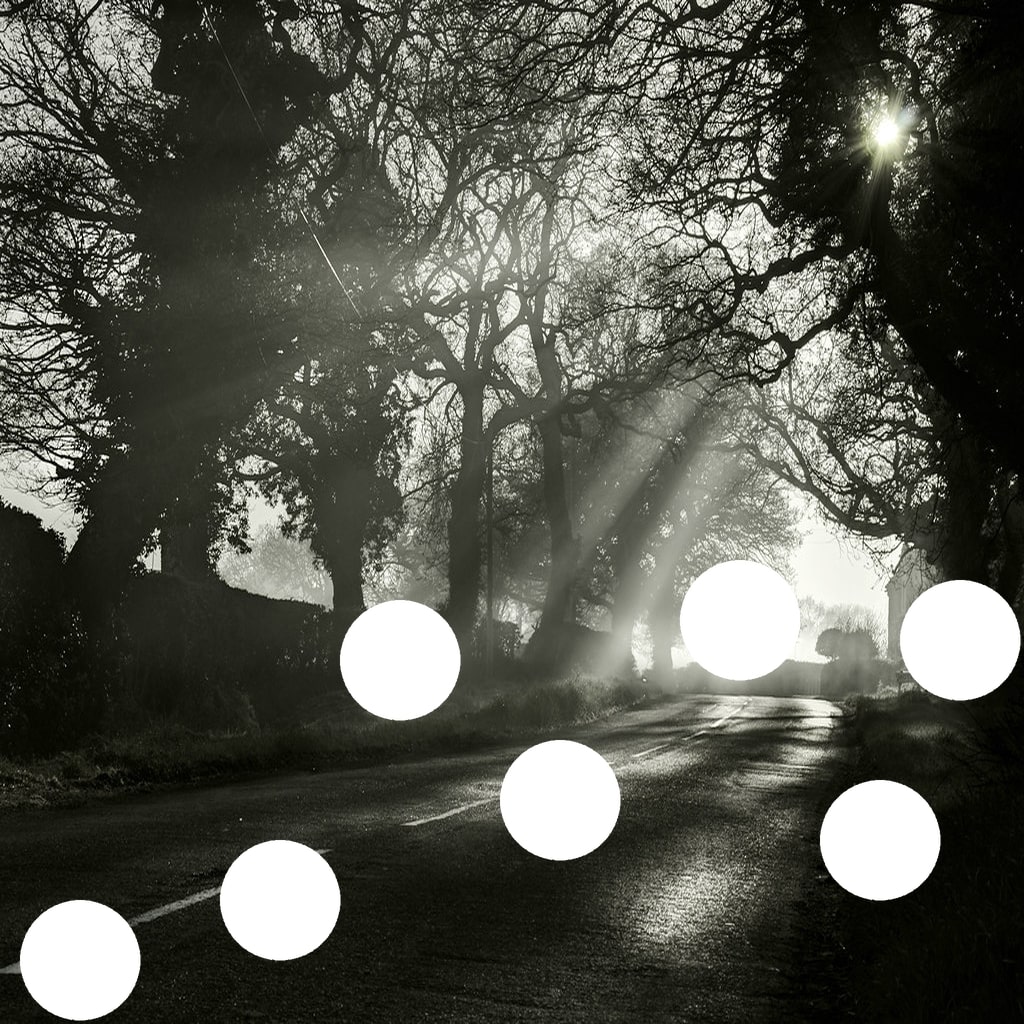} & 
  \includegraphics{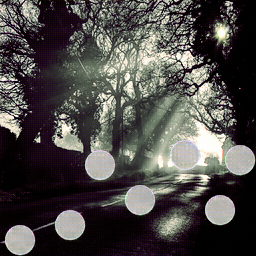} & 
  \includegraphics{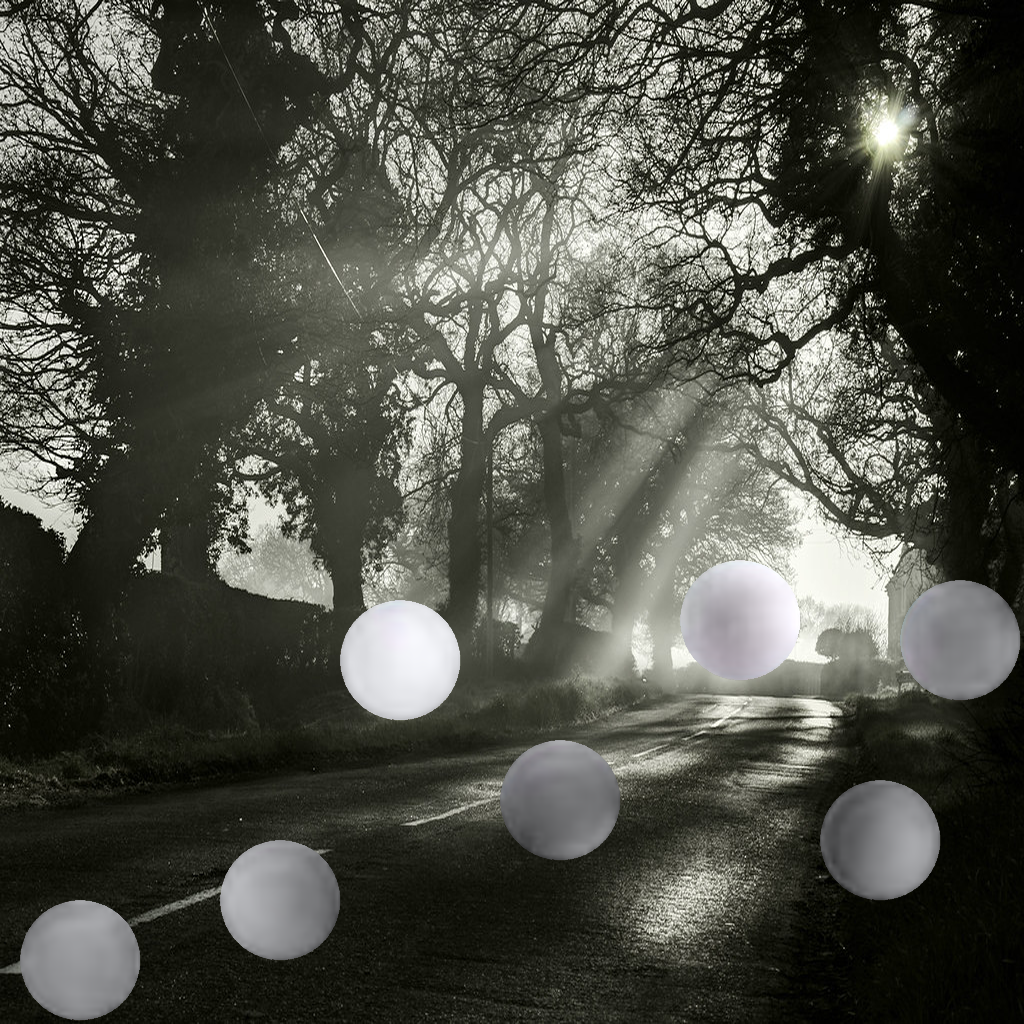}\vspace{2pt}\\
   \centering
   \scalebox{1}{Target Scene} & \centering \scalebox{1}{CP} &\centering \scalebox{1}{IH~\cite{cong2020dovenet}} &\centering \scalebox{1}{DIPR (ours) }\\
  \end{tabularx}
  \vspace{-10pt}
  \caption{{\bf Rendering spheres.} DIPR has some implicit notion of the 3D layout of the scene, which is required to choose
  the appropriate shading.  DIPR shades the white discs as spheres (rather better than IH,
  implying it ``knows" about shape; also see Fig.~\ref{fig:shape_recons}).
  \vspace{-15pt}}
  \label{fig:spheres} 
  \end{figure}
  {\noindent Reshadings are derived from consistent shapes.} DIPR renderings of circles look like spheres (see Fig~\ref{fig:spheres}), suggesting the method has some  notion of shape. We test if our shadings are consistent, using two procedures:  a large scale using singular values and explicit reconstruction (expensive in compute)  at a small scale.   Results suggest there is indeed some emergent notion of shape, obtained with no shape annotated data.
  
  {(a) \bf Large scale:} Imagine we have many different reshadings $S_{i,k}$ of a particular inserted shape (we use white circles in albedo).
  It is known that multiple shadings of the same geometry have important similarities~\cite{belhumeur1998set}.  Assume that the shading value $S_{i,k}(\vect{x})$ is a slowly changing spatial function of the (unknown) normal $\vect{N}(\vect{x})$, so that $S_{i,k}(\vect{x})=\sum_{m=1}^{N_s} a_{m,k}(\vect{x}) \phi_m(\vect{N}(\vect{x})$.
  Assume here that $k$ is small, and $a_{m,k}(\vect{x})$ are spatially slow.  These assumptions apply to, for example,
  spherical harmonic shading of diffuse surfaces~\cite{ramamoorthi2001relationship}) and shading using the spherical gaussian basis of~\cite{li2020inverse}.  Now
  straighten each image into a vector $\vect{S}_{i,k}$. Then these vectors span a $N_s$ dimensional space.  
  Form ${\cal D}_i=\left[\vect{S}_{i,1}^T, \ldots, \vect{S}_{i,N_o}\right]$ (where $N_o>N_s$).
  We expect the singular values $\sigma_r$ of ${\cal M}_i$ to be small for $r>N_s$ if all the $\vect{S}_{i,j}$ are
  shadings of the same scene, and large if they are not.  Using these observations to make a test of consistency requires knowing what is a ``small'' singular value, and what
  $N_s$ and $N_o$ should be.    We use $N_s=14$ and $200$ sample points on each shading field.  We now draw $10000$ sets of $N_s$
  reshadings from our examples, and look at the test statistic $T{=}\frac{\Pi_{r=10}^{14}\sigma_r}{\sigma_0^5}$.  
  We see a mean of 0.21 and a standard deviation of $7.4{\times}10^{-2}$ for $T$, which appears to be normally distributed. We compare this with a baseline of randomly chosen shading from natural scenes (cropped to the 2D disk). This yields a mean of 0.66 and a standard deviation of $5.9{\times}10^{-2}$.   We conclude DIPR reshadings of spheres are strongly different from random shading and display a degree of shape consistency.
  
  { \bf (b) Explicit shape reconstruction:} 
  We produce explicit shape of our inserted
  circles from their predicted shadings. We take $7$ reshaded circles, each
  from two different images. From this pool of $14$, we draw $7$ at random, use them to drive a
  shape reconstruction procedure (see Appendix; a typical reconstruction in Fig.~\ref{fig:shape_recons}). We then compute the spherical harmonic reshading of the resulting reconstruction that is
  closest to each of the $7$ held out images and record
  the mean squared error of the shading residual for each. A small residual means that
  the shape is consistent across the reshaded circles. Box plots
  of the resulting residuals for $7$ different splits of the data
  in Fig.~\ref{fig:shape_recons}. To calibrate, we compare this with $3$ baselines; we reshade:  (1) a sphere using actual spherical
  harmonic shading, reconstruct from reshadings, then predict
  the held out reshadings (`SphHarm’), (2) a constant height surface (`Const’) and (3) a surface reconstructed from smoothed random
  noise shadings (`Rand’). These residuals and singular value analysis suggest the reshading network has some form of shape theory as an emergent property.

  \begin{figure}[t]
    \setkeys{Gin}{width=0.95\linewidth}
    \renewcommand\arraystretch{0}
    \begin{tabularx}{\linewidth}{@{}X@{}X@{}}
  
      \includegraphics[width=\linewidth]{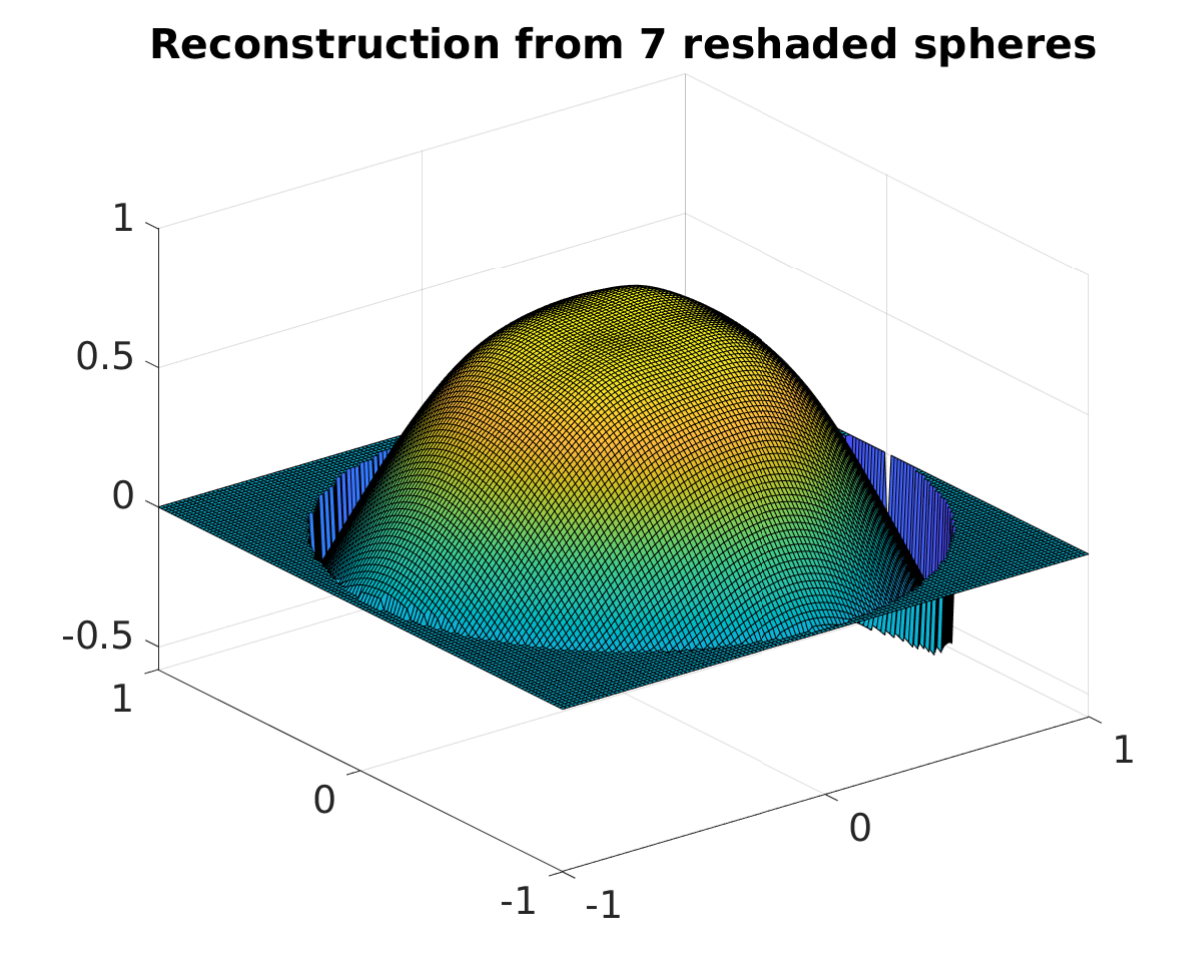} &
      \includegraphics[width=\linewidth]{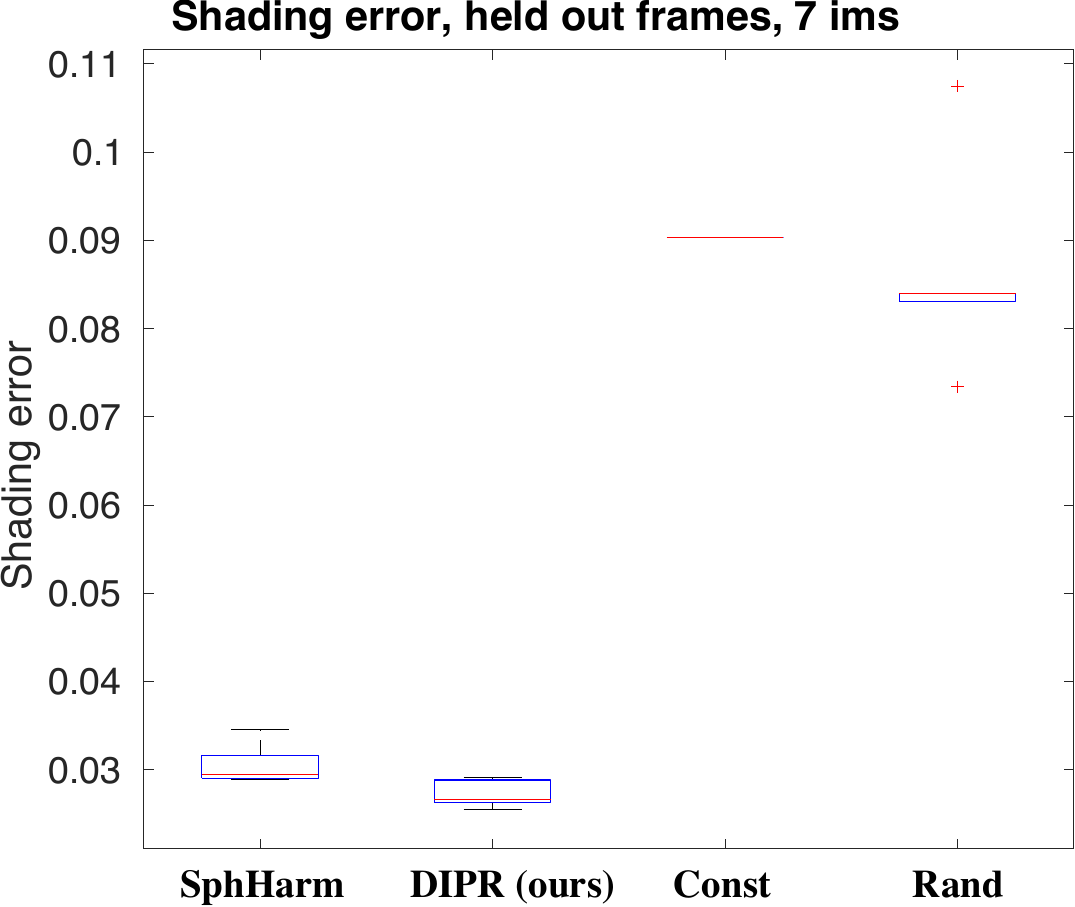}\\
  \end{tabularx}
  \vspace{-3pt}
  \caption{{\bf Shape theory and shading consistency:} On the {\bf left} a sample surface reconstruction produced by using $7$ reshaded spheres (Fig. \ref{fig:spheres}).  On the right,
  predicted shading residuals for held out spheres' shadings using our reconstructed shapes. The residuals suggest:  there is a
  consistent underlying shape. Our DIPR residuals are small. The reconstruction process is reliable but the underlying
  surface is not quite a sphere. `SphHarm' residuals are also small, but not as small as DIPR. `Const' residuals are large. Therefore, underlying surface is not flat. Smooth `Rand' residuals are
  large, that is, random shadings cannot explain this consistency.}
  \label{fig:shape_recons} 
  \vspace{-5pt}  
\end{figure}

\section{Discussion and Conclusions}

\begin{figure}[t]
  \centering
  \includegraphics[width=0.95\linewidth]{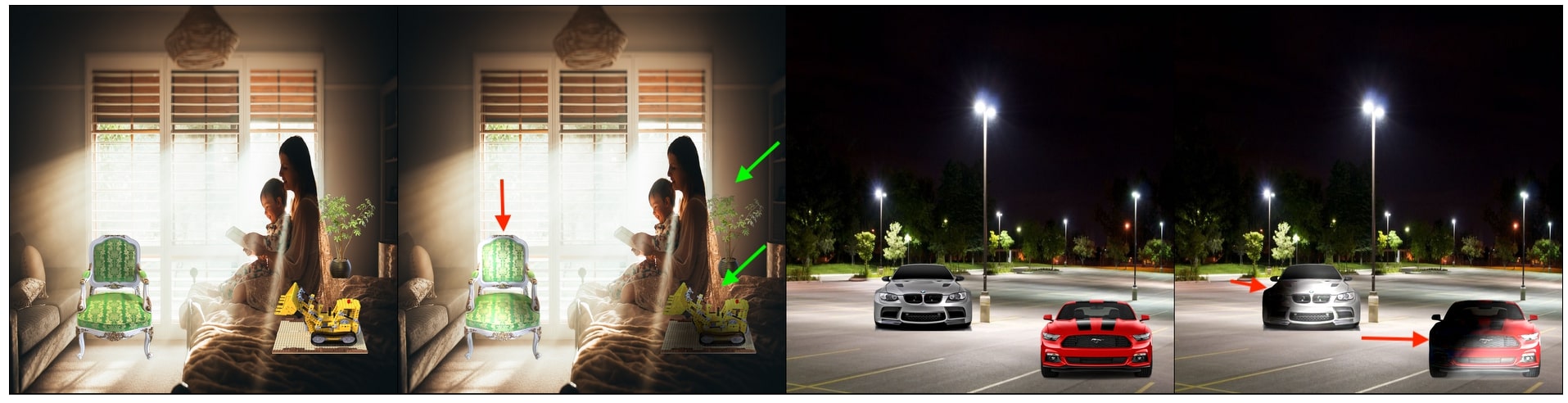}\\
  \vspace{-3pt}
  \centering
  \small{
    \begin{tabularx}{\linewidth}{*{4}{>{\centering\arraybackslash$}X<{$}}}
      \scalebox{1}{CP} &  \scalebox{1}{DIPR (ours)} &  \scalebox{1}{CP}&  \scalebox{1}{DIPR (ours)}
    \end{tabularx}
    }\hfill
    \vspace{-12pt}
  \caption{{\bf Failure cases.} Red arrows point to shading errors.
  In first image, DIPR aggressively copies background shading onto the chair. However, lego's and plant's shading looks plausible. In the second scene, it has two dominant normals -- the ground (upwards) and the sky (towards viewer). The lack of third direction results in copying shading either from the sky or the ground.}
  \vspace{-15pt}
  \label{fig:failure}
\end{figure}
{\noindent \bf Limitations.} DIPR is slow because DIP takes $15$ mins to render. DIPR cannot cast shadows; very hard and we leave that for our future work. DIPR likely copies shading, so has problems when there is little shading variation and responds poorly when there are “few” normals in the scene (Fig. \ref{fig:failure}).

{\noindent \bf Why DIPR works?} Corrections to object shading cannot be veridical.~\cite{liao2015cvpr,liao2019IJCV} finds corrected shading often fool humans more effectively than physically accurate lighting,
likely because humans attend to complex materials much more than to consistent lighting~\cite{berzhanskaya2005remote}.
The alternative physics theory~\cite{cavanagh2005tracking} argues that the
brain employs a set of rules that are convenient, but not strictly
physical and a violation leads to 
perception alarm or affects recognition negatively~\cite{gauthier1998training}. Otherwise, the scene “looks
right”. This means humans may tolerate a fair degree of
error, as long as it is of the right kind. By requiring image to produce consistent inferences, we appear to be forcing
errors to be ``of the right kind''.

\section*{Acknowledgments}
\vspace{-2pt}
\noindent This material is based upon work supported in part by the National Science Foundation under Grant Nos. 2106825 and 1718221, in part by ONR MURI Award N00014-16- 1-2007, and in part by an award from FutureWei. We thank Derek Hoiem, Kevin Karsch, Zhizhong Li and Dominic Roberts for their feedback and suggestions on the initial draft of the paper.\appendix
\section*{\Large Supplementary Material}

\begin{figure}[htpb!]
  \setkeys{Gin}{width=\linewidth}
  \renewcommand\arraystretch{0}
  \centering
  \resizebox{0.98\linewidth}{!}{
    \centering
    \begin{tabularx}{\textwidth}{@{}X@{}X@{}X@{}X@{}}
  \includegraphics{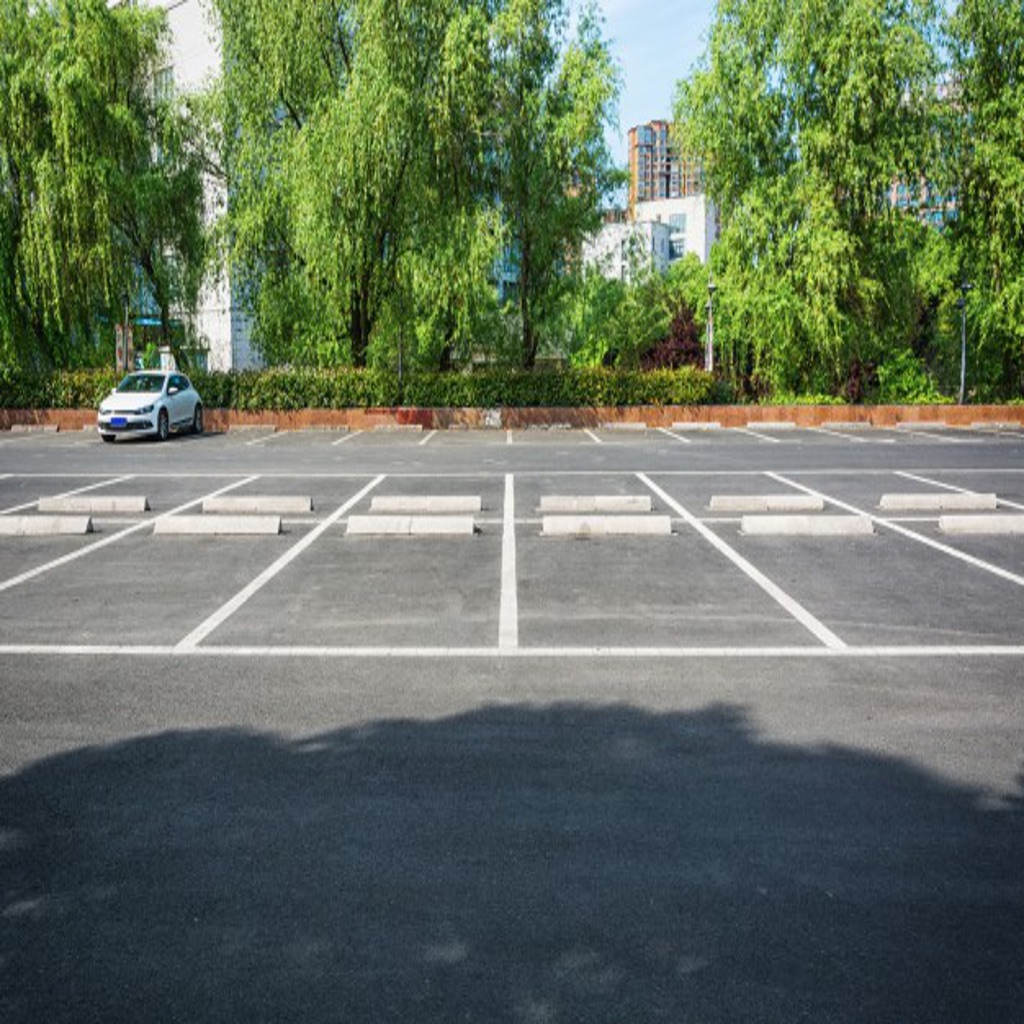} &
  \includegraphics{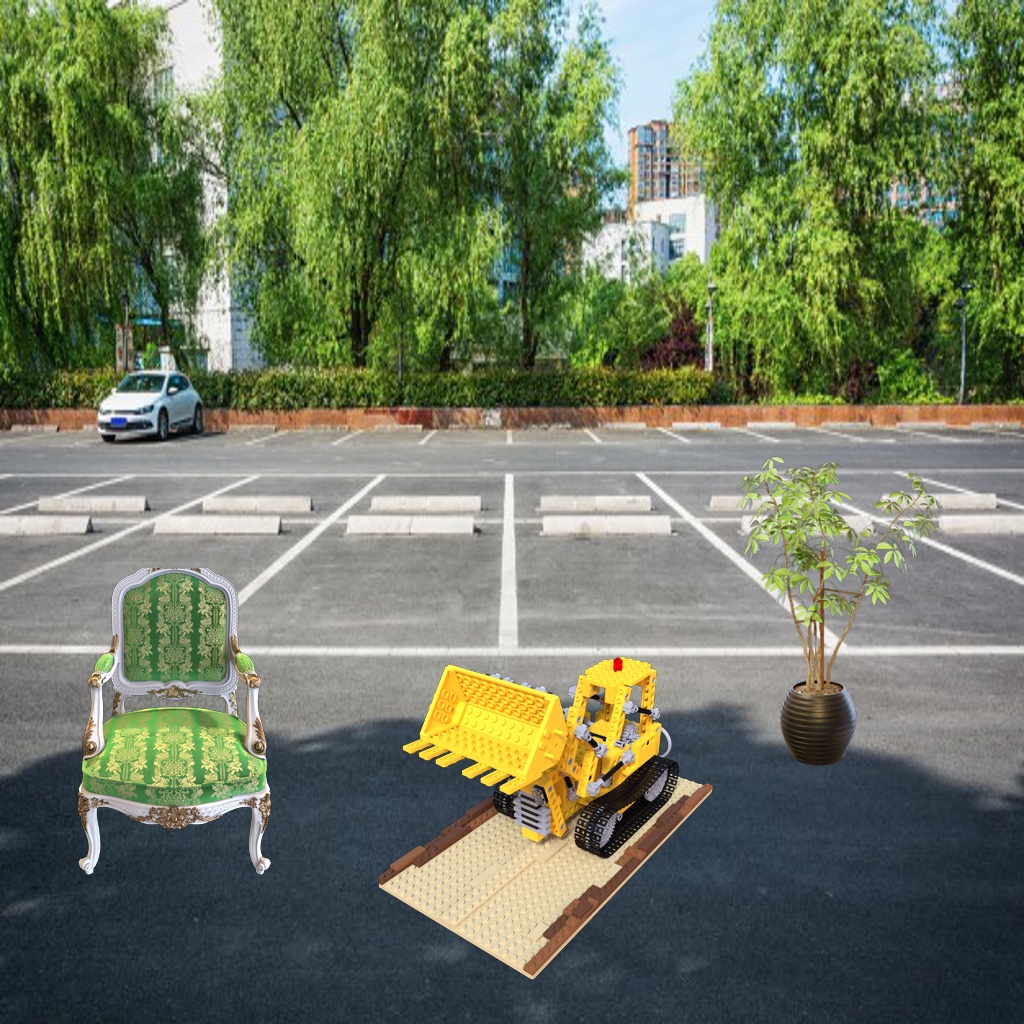} & 
  \includegraphics{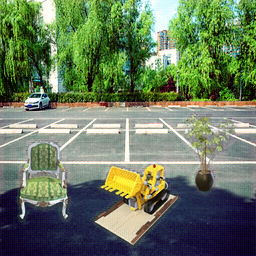} & 
  \includegraphics{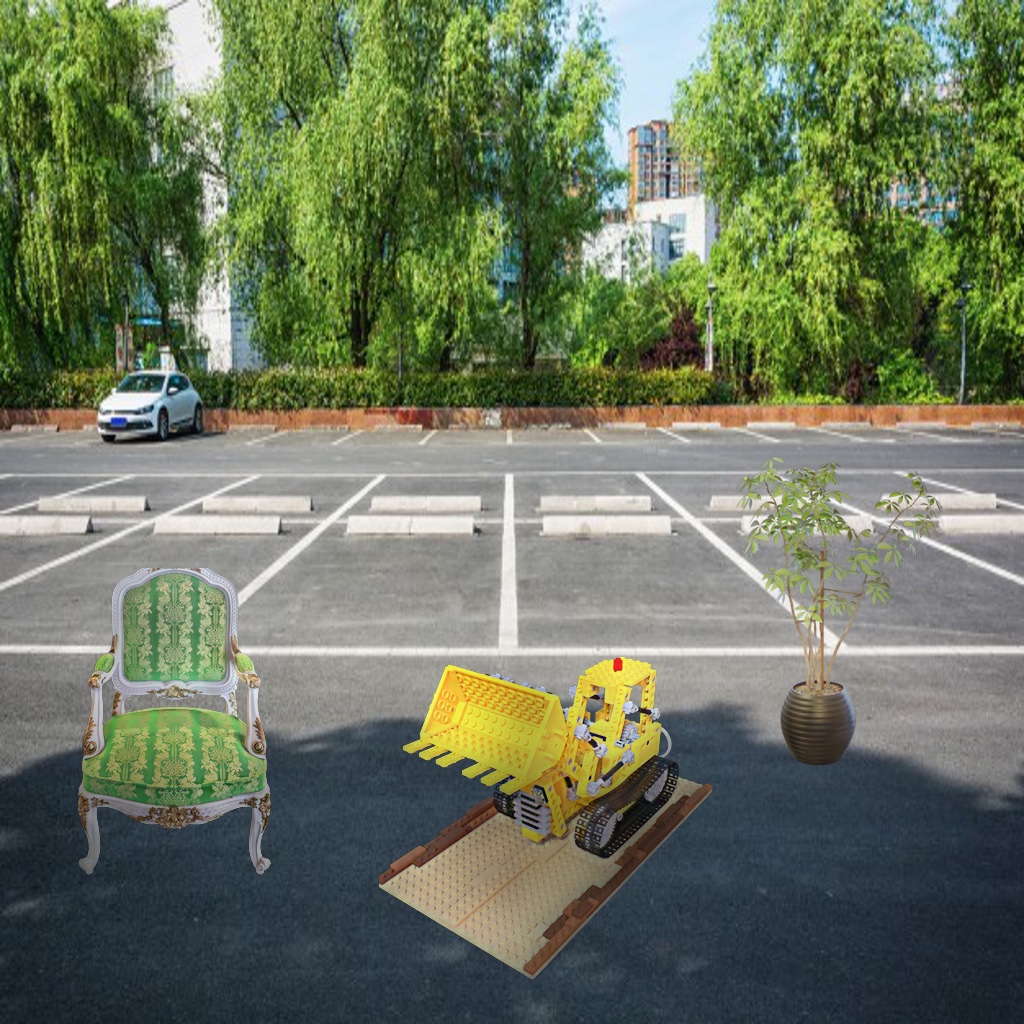} 
  \\ & 
  \includegraphics{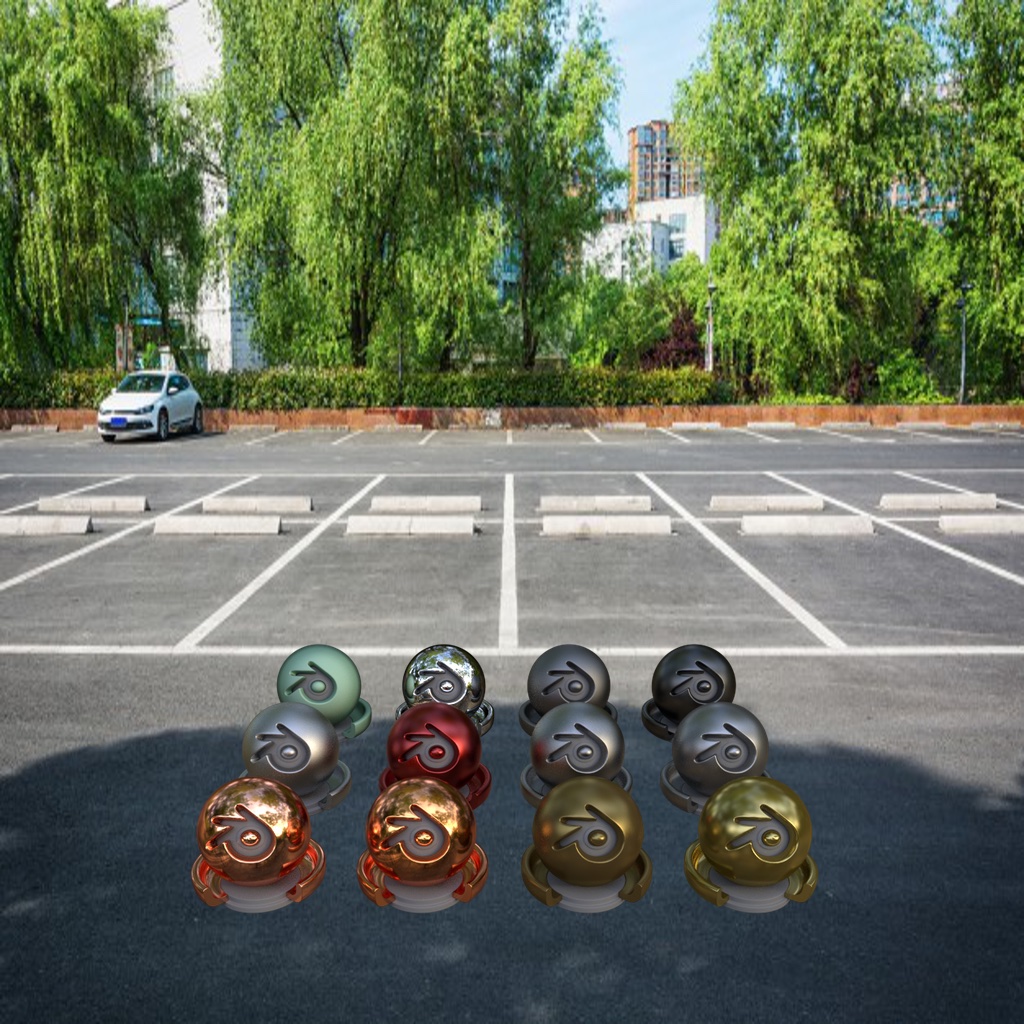}&
  \includegraphics{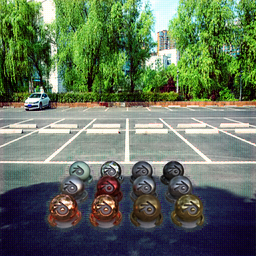}&
  \includegraphics{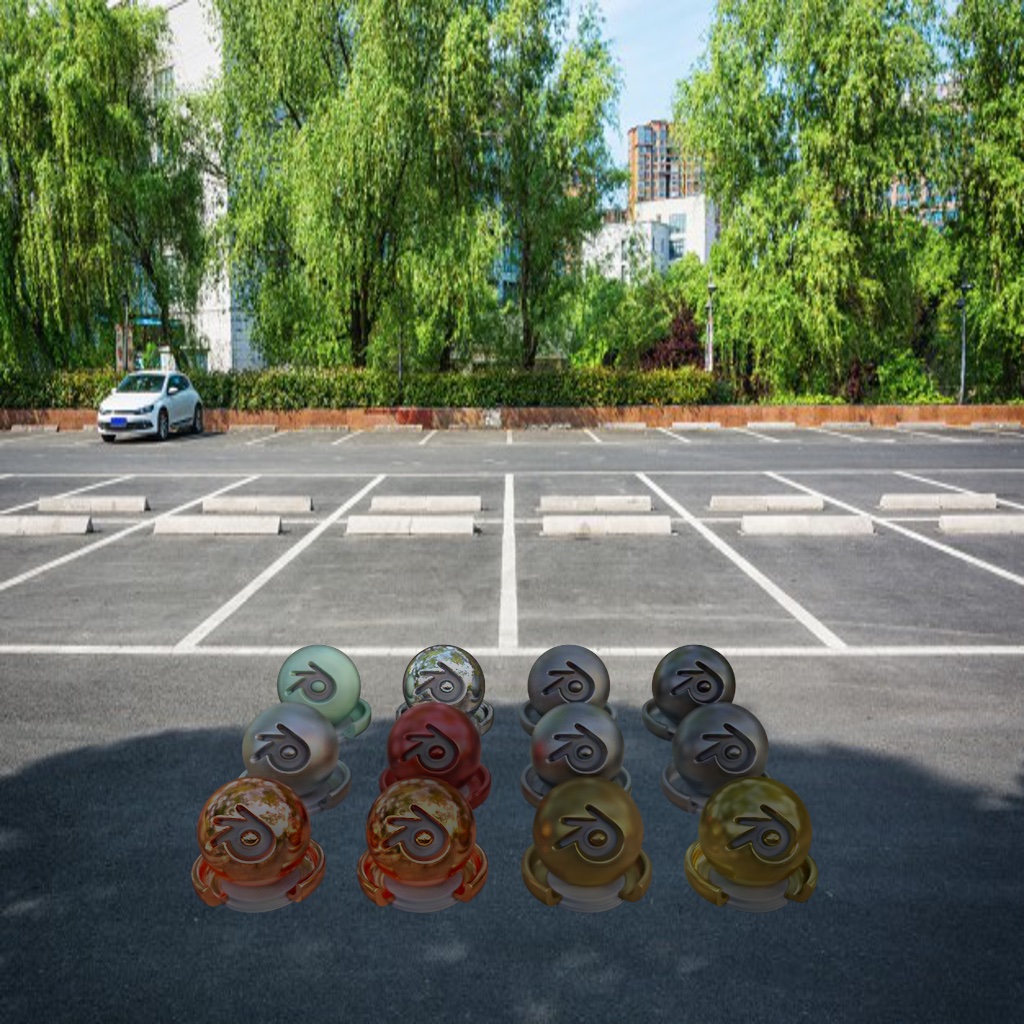}\\

  \includegraphics{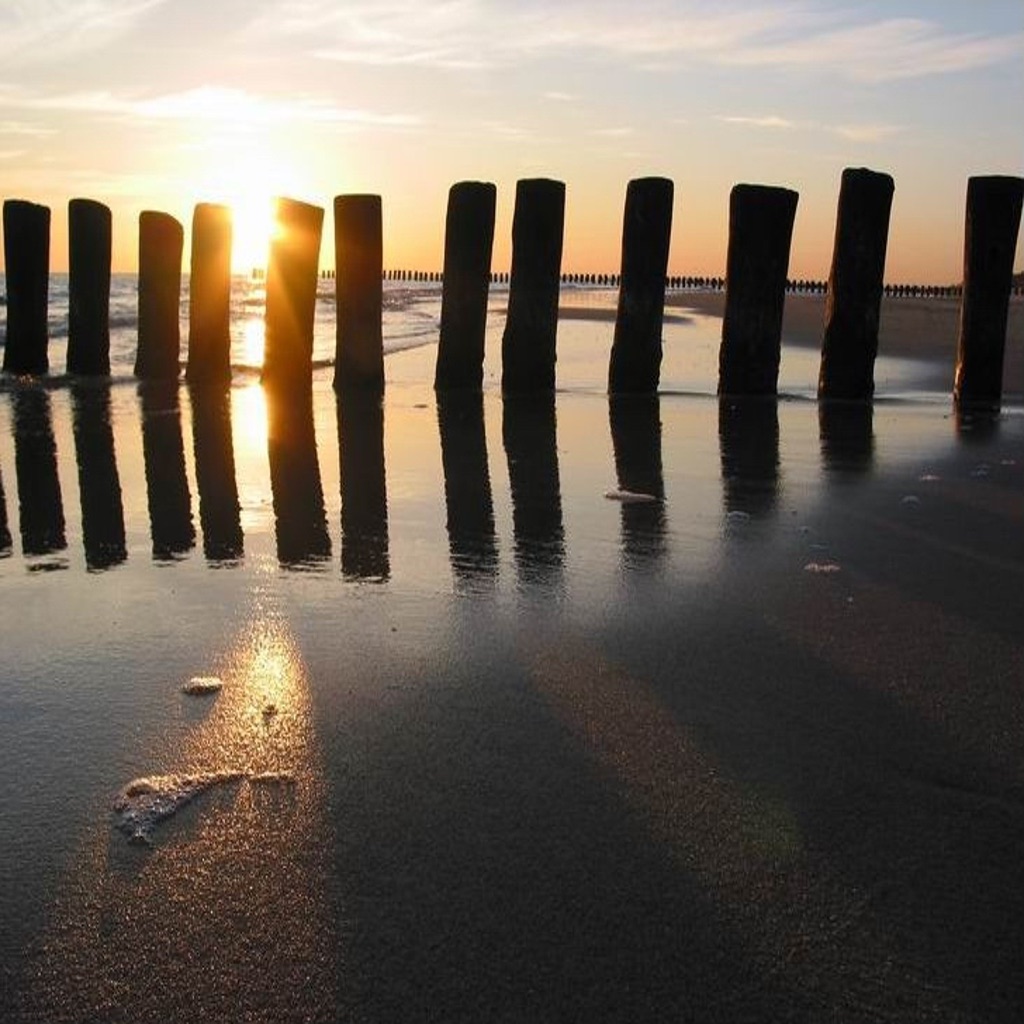} &
  \includegraphics{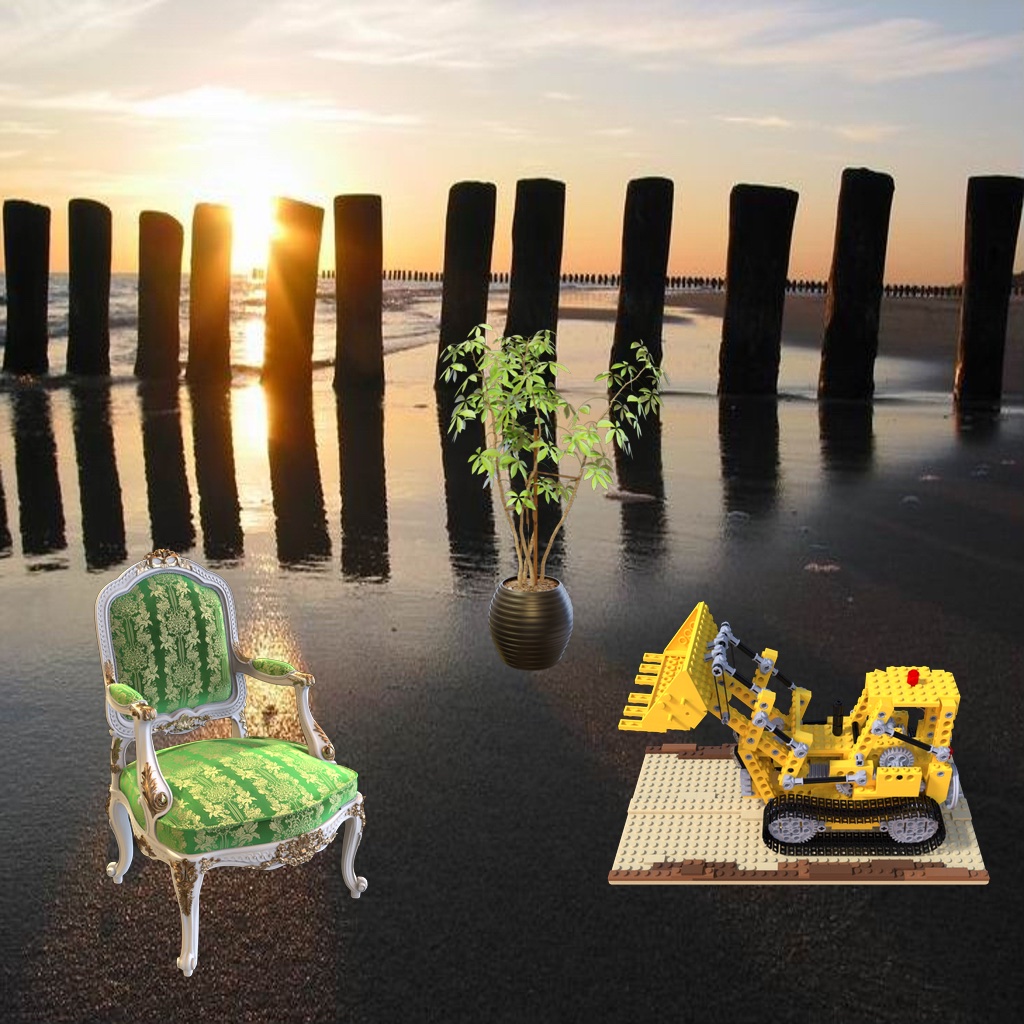} & 
  \includegraphics{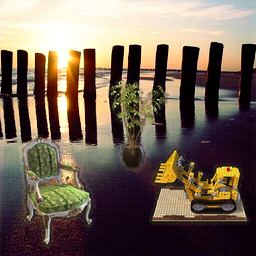} & 
  \includegraphics{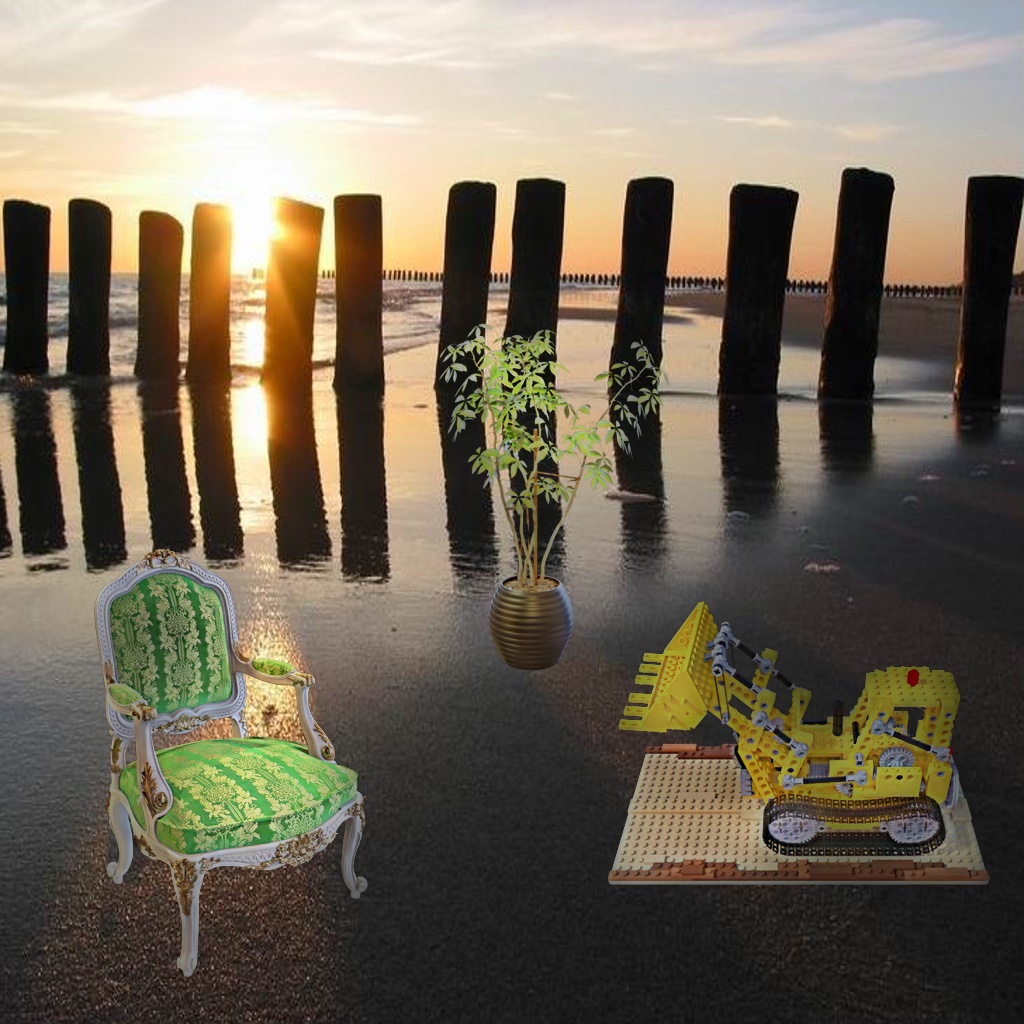} 
  \\ & 
  \includegraphics{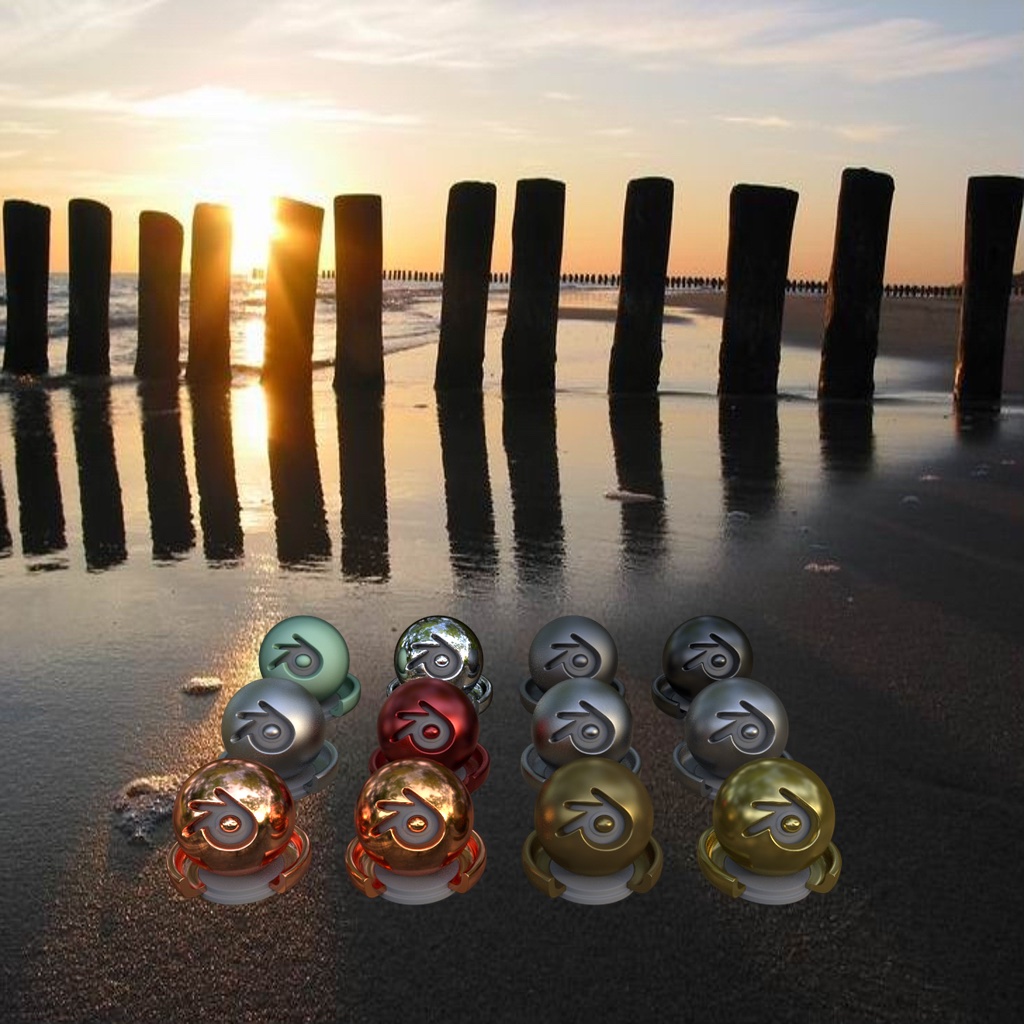}&
  \includegraphics{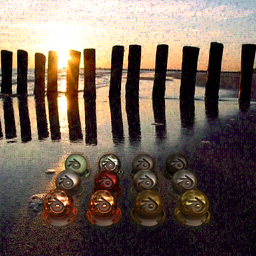}&
  \includegraphics{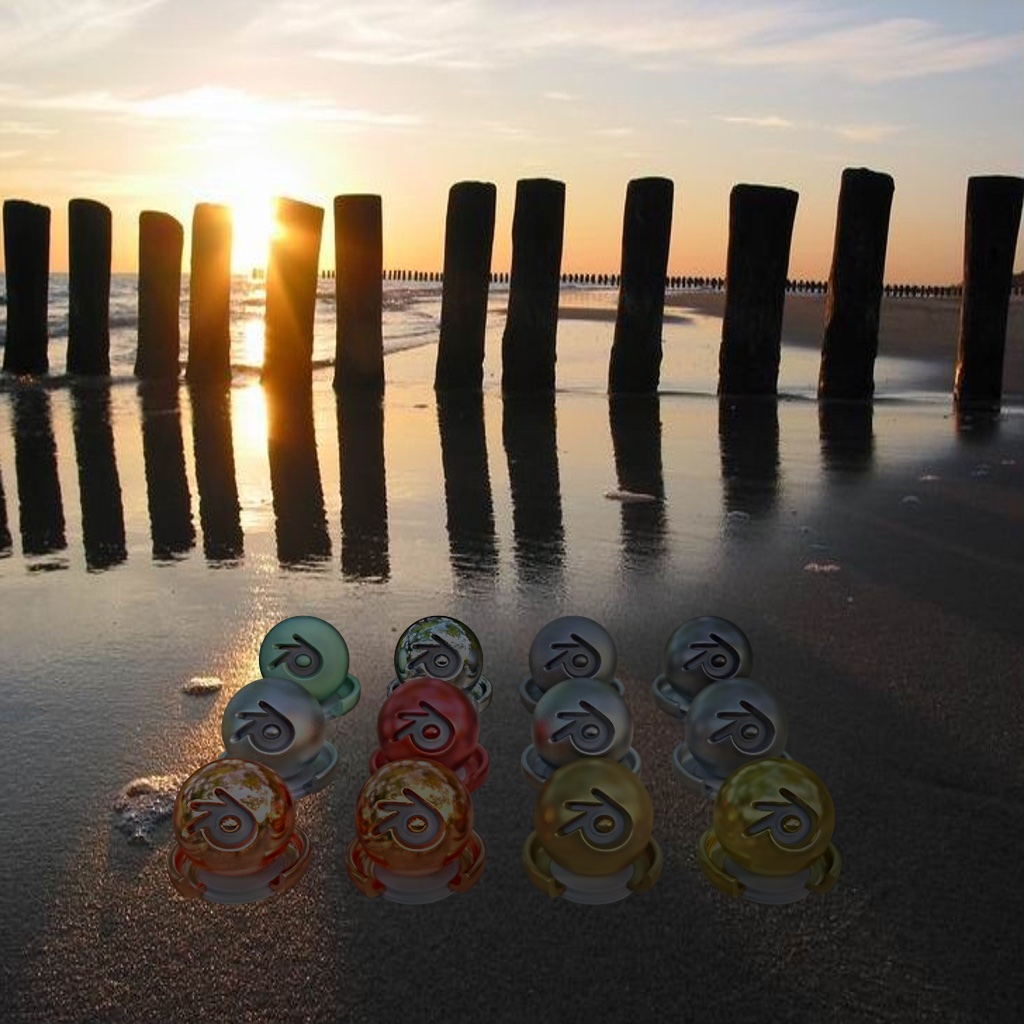}\vspace{3pt}\\
\centering\scalebox{0.8}{Target Scene} & \centering \scalebox{0.8}{CP} &\centering\scalebox{0.8}{IH~\cite{cong2020dovenet}}&\centering \scalebox{0.8}{DIPR (ours)} 
  \end{tabularx}
  }
  \vspace{1pt}
\caption{{\bf Reshading complex sufaces.} DIPR reshades objects with complex surface geometry and complex material properties convincingly, and appears to preserve the identity of the material without a distinct shift in object and
background color produced by IH.}
\label{fig:synth}
\end{figure}

\begin{table}[htpb!]
  \centering
    \resizebox{0.9\linewidth}{!}
{
  \begin{tabular}{cccc}
    \toprule
    Images & NIQE~\cite{mittal2012making}$\downarrow$ & CNN-D~\cite{wang2019cnngenerated}$\downarrow$ & B-T~\cite{cong2020dovenet}$\uparrow$\\
    \midrule 
    IH & 38.39 & 0 & 0\\
    CP & \bf{9.67} & 0 & -0.35 \\
    DIPR (ours) &  9.78 & 0 & \bf{0.04}\\
    \midrule
    Target Scene & 12.51 & 0 & NA\\
    \bottomrule
  \end{tabular}
}
  \vspace{6pt}
  \caption{ Additional quantitative Analysis for plants, lego, 16-set materials and cars when no ground truth is available. We report numbers for a no-reference image quality estimator (NIQE), a CNN-based tampering detector (CNN-D) and Bradley-Terry (BT) model's analysis from the user-study. Surprisingly, NIQE is worse for target scenes, CNN-D cannot detect any image as fake, and from the user study between three methods, BT analysis shows our reshaded images are preferred over CP and IH. }
  \label{tbl:user}
  \vspace{-8pt}
\end{table}

\begin{figure*}[htp]
  \setkeys{Gin}{width=\linewidth}
  \renewcommand\arraystretch{0}
  \centering
  \resizebox{0.98\linewidth}{!}{
    \centering
  \begin{tabularx}{\textwidth}{@{}X@{}X@{}X@{}X@{}X@{}X@{}}
  \includegraphics{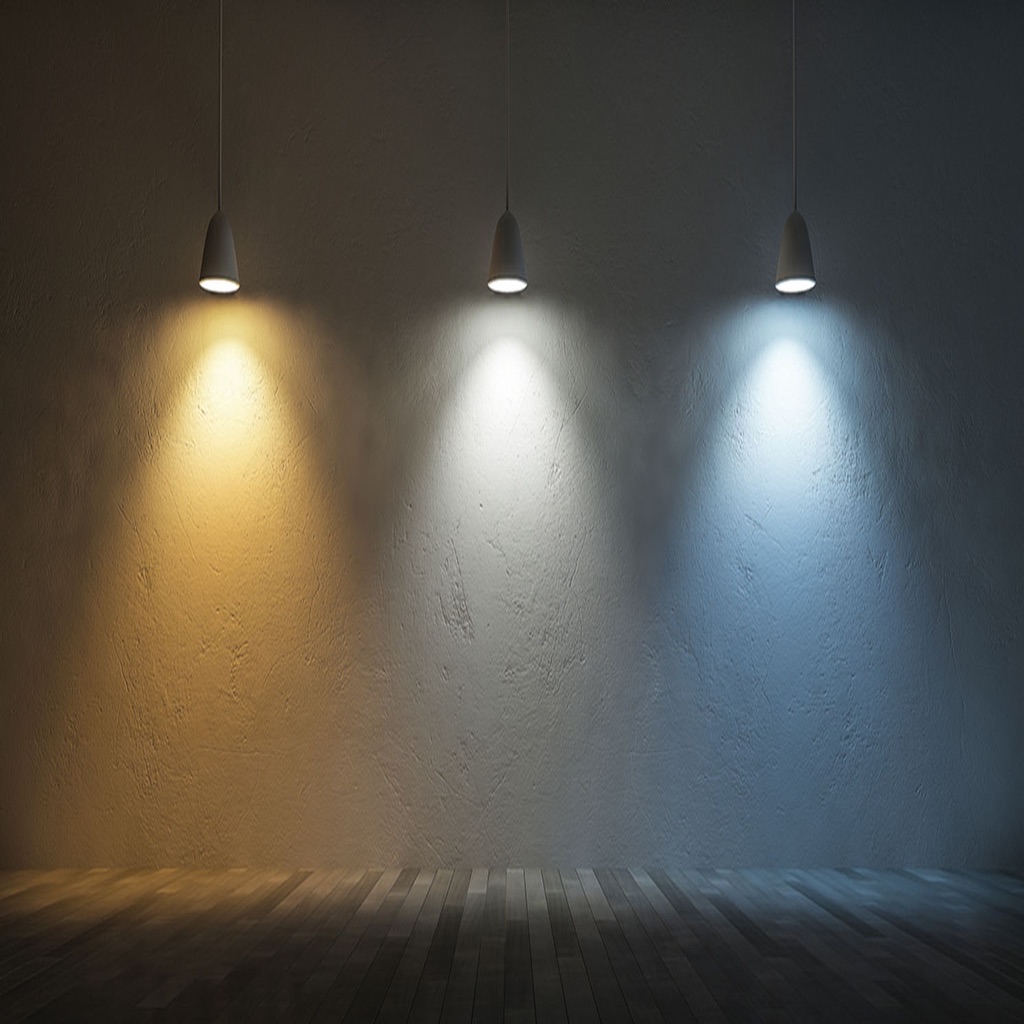} &
  \includegraphics{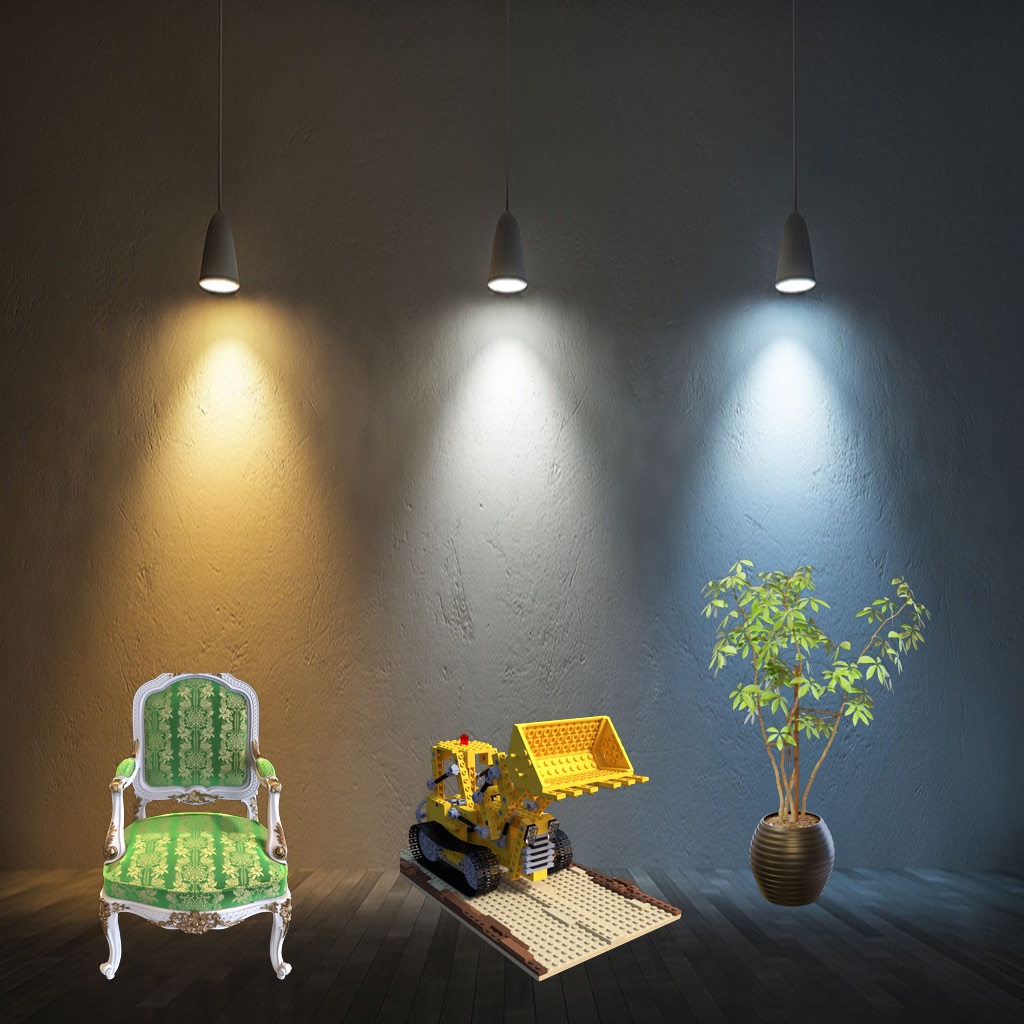} & 
  \includegraphics{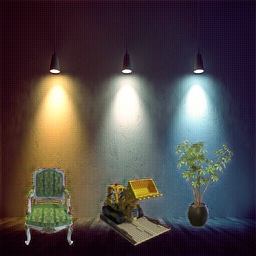} & 
  \includegraphics{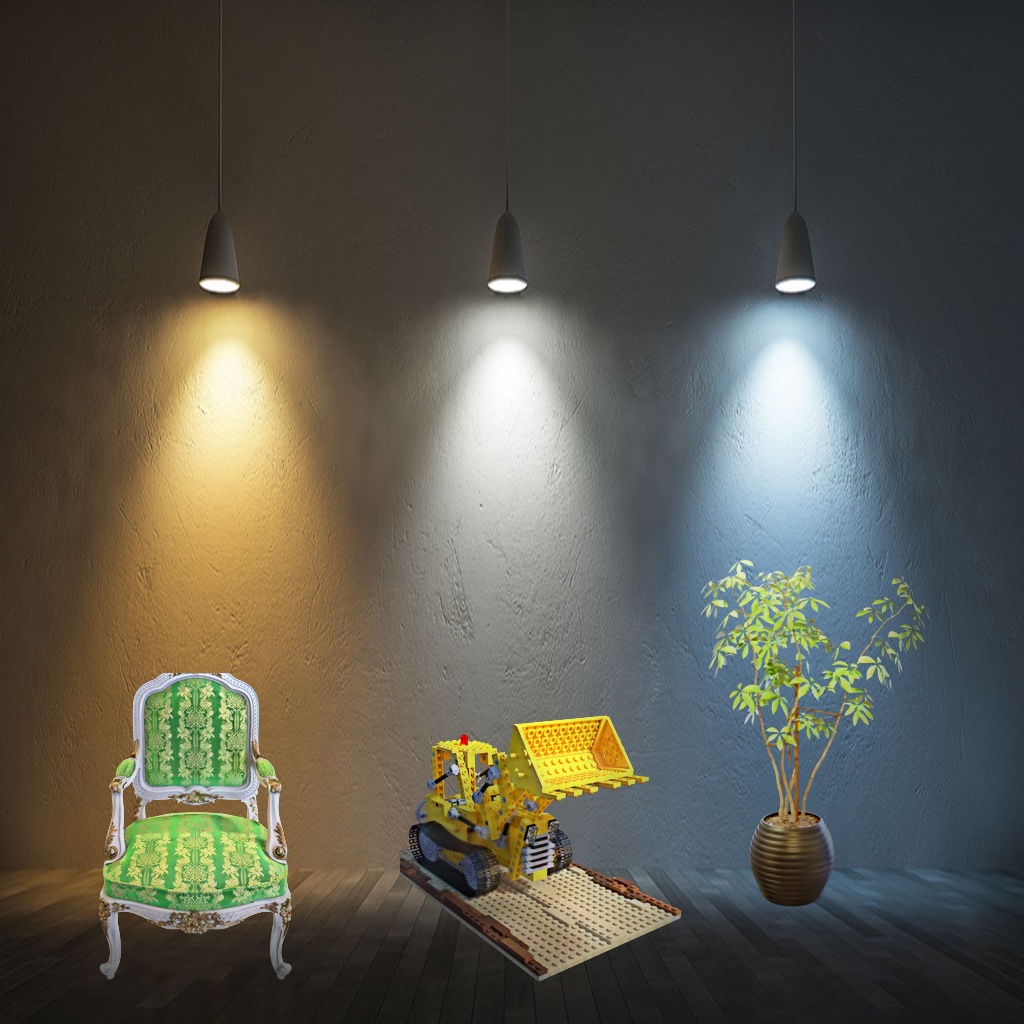} & 
  \includegraphics{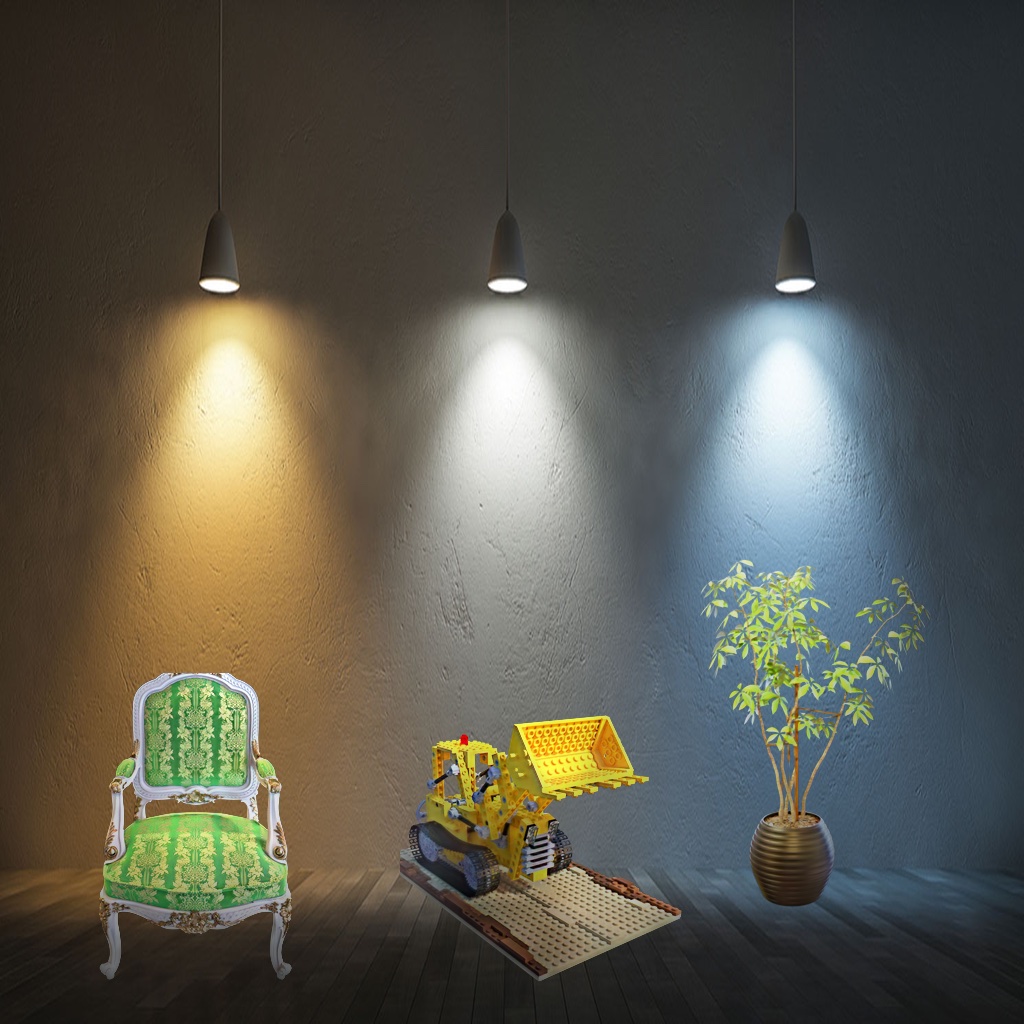}&
  \includegraphics{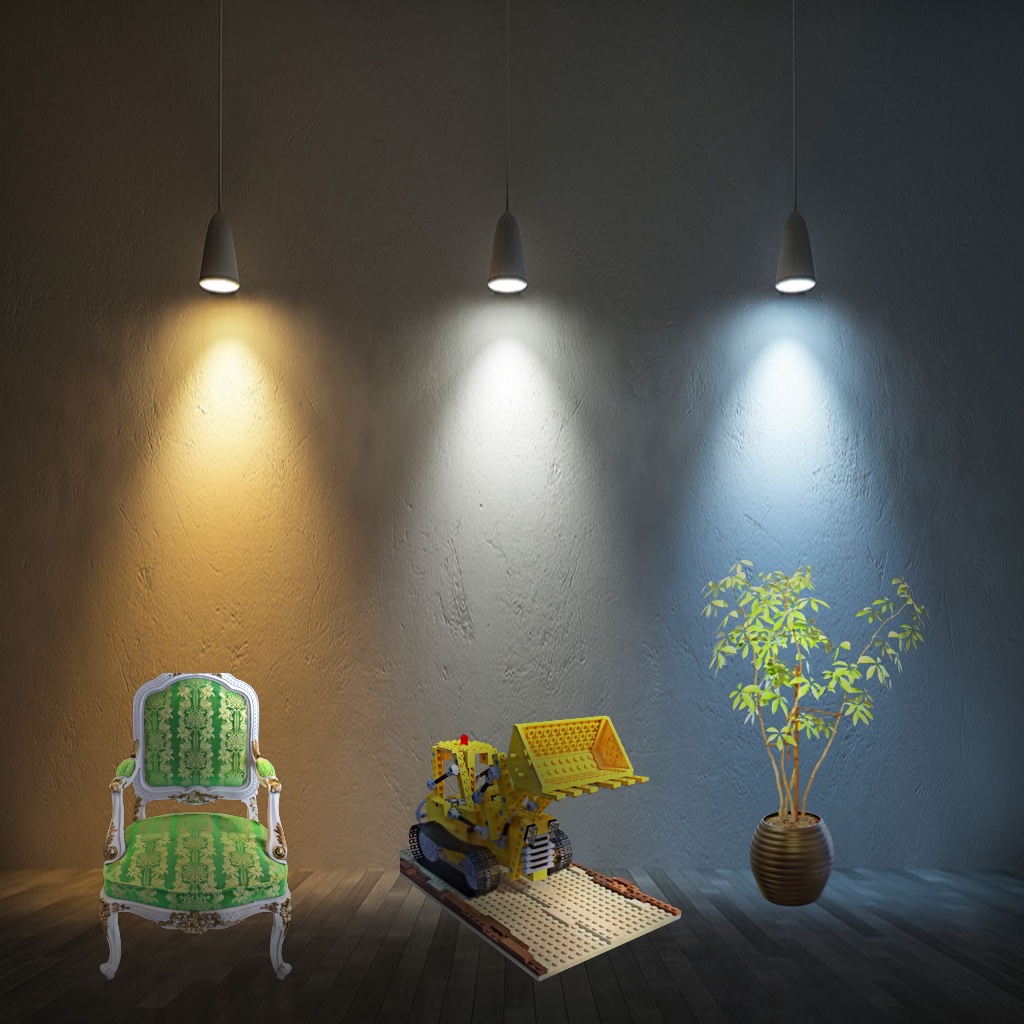}\vspace{2pt}\\
  \centering
  \scalebox{0.8}{Target Scene} & \centering \scalebox{0.8}{CP} &\centering\scalebox{0.8}{IH~\cite{cong2020dovenet}}&\centering \scalebox{0.8}{$\mathcal{L}_{recons}+\mathcal{L}_{decomp}$} &\centering \scalebox{0.8}{+$\mathcal{L}_{Z}$} &\centering \scalebox{0.8}{+$\mathcal{L}_{Z}+\mathcal{L}_{Y}$}\\
  \end{tabularx}
  }
  \vspace{-8pt}
  \caption{{\bf Ablation Study.} We show the role of each of the loss components. Results improve from left to right (see the lego dozer). We find neural shading consistency loss ($\mathcal{L}_{Z}$) helps in improving local changes based on the near-by surroundings and spherical harmonics matching loss ($\mathcal{L}_{Y}$) considers the overall lighting providing complementary solutions. Tab. 2 in the main text has quantitative analysis for losses.}
  \label{fig:ablation}
  \vspace{8pt}
\end{figure*}

\begin{figure}[htp]
  \setkeys{Gin}{width=\linewidth}
  \renewcommand\arraystretch{0}
  \centering
  \resizebox{\linewidth}{!}{
    \centering
  \begin{tabularx}{\textwidth}{@{}X@{}X@{}X@{}X@{}}
    \includegraphics{Figures/cars/target/target_parking_shaft.jpg} &
    \includegraphics{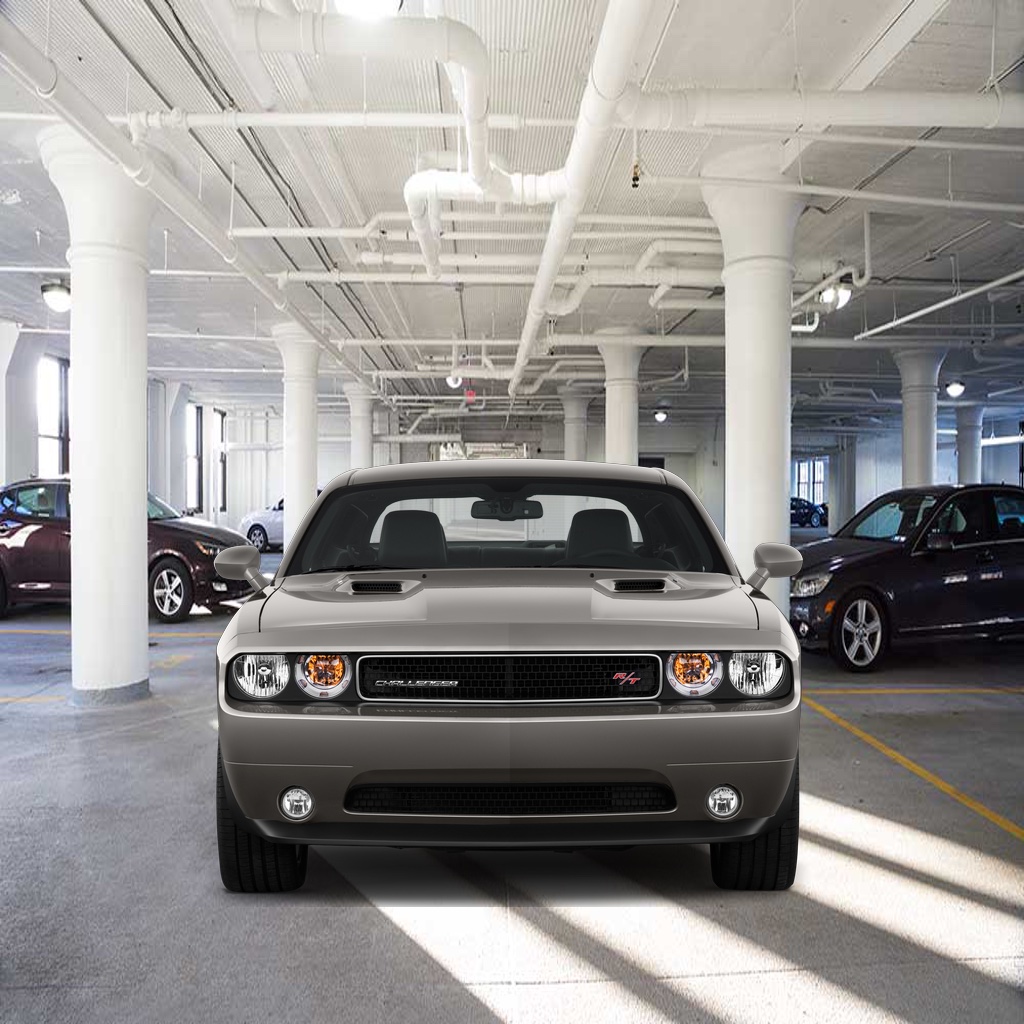}&
    \includegraphics{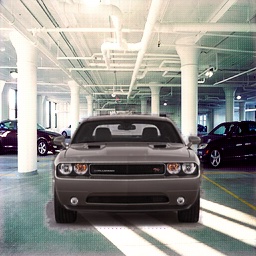}&
    \includegraphics{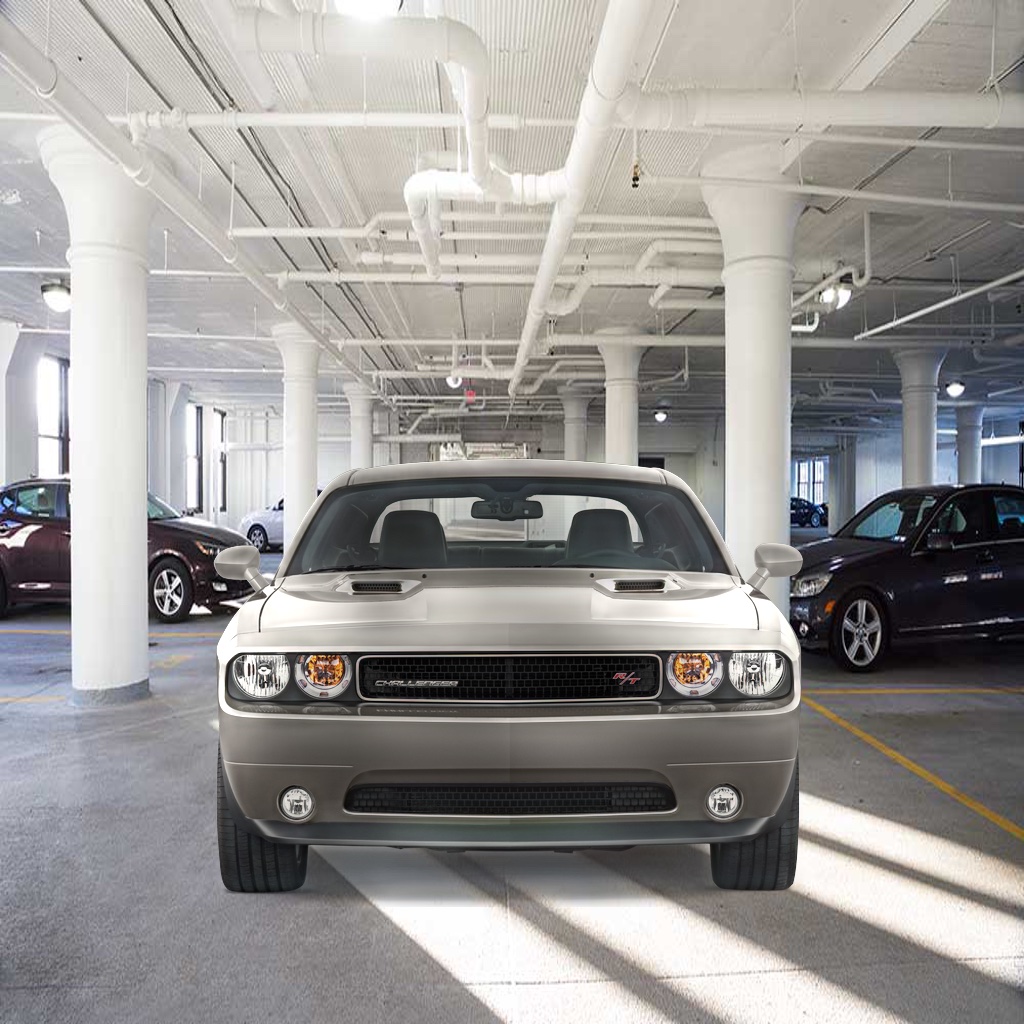}\\
    \includegraphics{Figures/cars/target/target_jersey_night.jpg} 
     &  
   \includegraphics{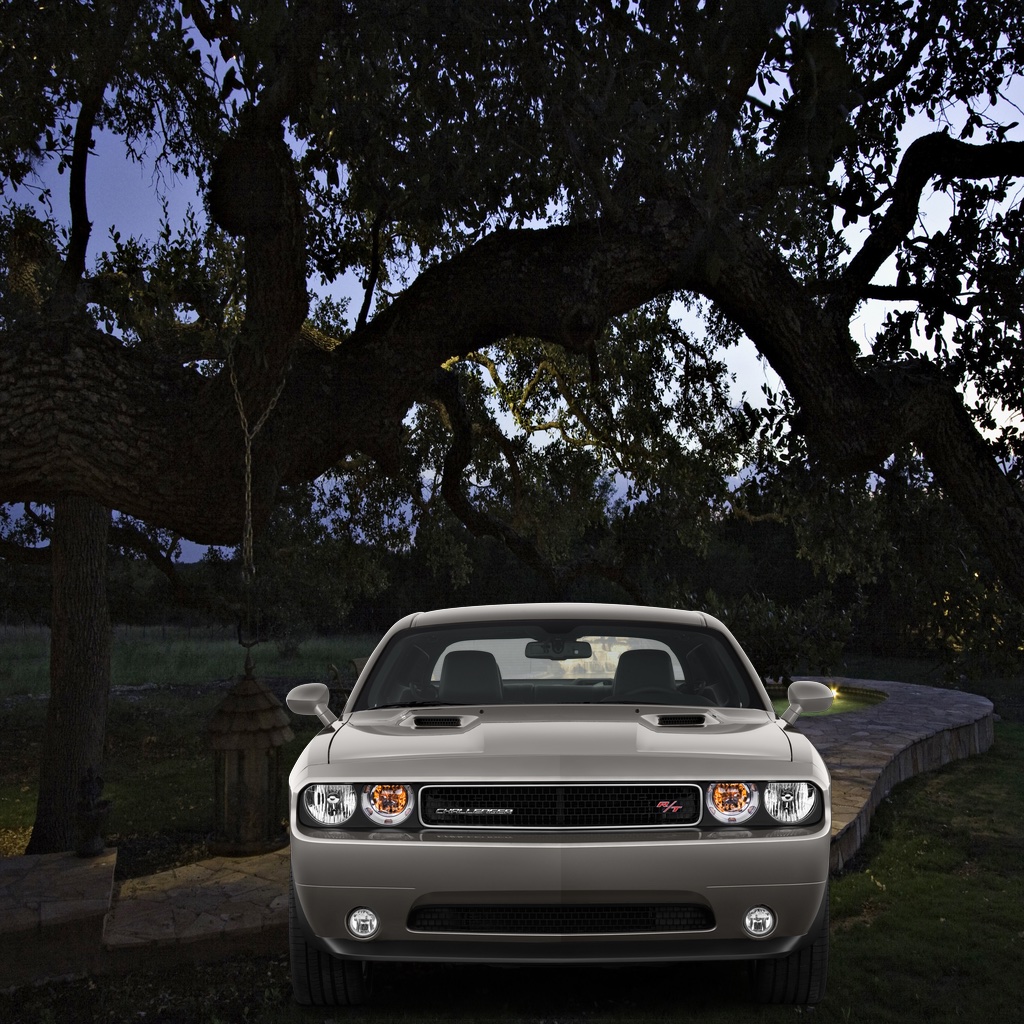}&
   \includegraphics{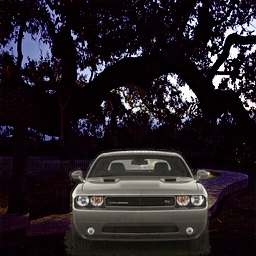}&
   \includegraphics{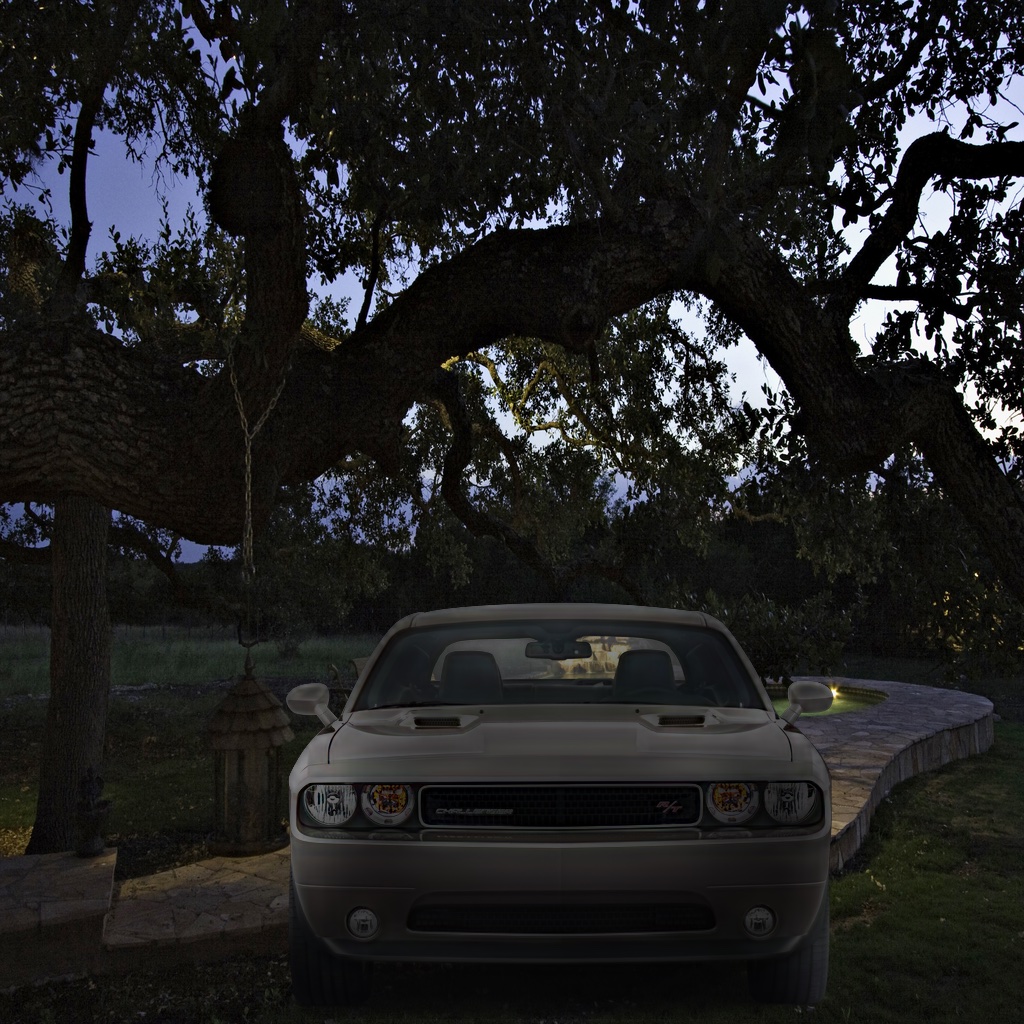}\vspace{3pt}\\
  \centering\scalebox{0.8}{Target Scene} & \centering \scalebox{0.8}{CP} &\centering\scalebox{0.8}{IH~\cite{cong2020dovenet}}&\centering \scalebox{0.8}{DIPR (ours)} 
  \end{tabularx}
  }
  \vspace{2pt}
  \caption{{\bf Additional examples of reshading real cars.} Glossy effects in car paint with glitter in them make reshading cars a particularly challenging case with their complex reflective properties. DIPR
  successfully reshades cars without a distinct shift in object and background color produced by IH. Note the bright patch on the metallic bonnet of the grey car in the top row that may possibly be because of the light source just above it.
  }
  \label{fig:cars}
\end{figure}

\begin{figure}[b!]
  \centering
  \includegraphics[width=\linewidth]{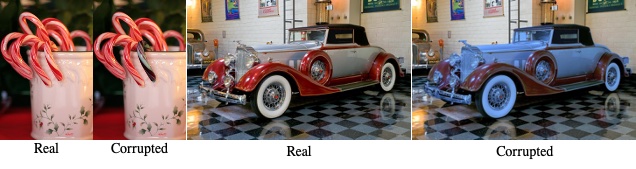}
  \caption{{Samples from iharmony dataset~\cite{cong2020dovenet}.} Image Harmonization methods mostly deals  with albedo or color correction and for object reshading we expect only shading corrections.}
  \label{fig:materials}
  \vspace{-0.25cm}
  \end{figure} 

  \section{3D Shape Reconstruction}
  We wish to reconstruct a shape from $N$ images, each showing that
  shape illuminated by an unknown spherical harmonic illumination field.  Illuminants are restricted to the
  first nine spherical harmonics.  We choose $N=7$, because preliminary experiments showed that this was
  the smallest $N$ that led to reasonably reliable reconstructions.  We ignore boundary conditions, and
  reconstruct by constructing a surface $(x, y, f(x, y))$ and a set of spherical harmonic coefficients $a_{ij}$ to minimize
  \begin{equation}
    \small
    \sum_{j=1}^{7} \mid\!\mid\!(\mbox{shading predicted by $f$ and $a_{\cdot j}$}-I_i)\mid\!\mid\!^2 \\
    +1e-5 \mbox{smoothing
      term}
\end{equation}
  where the smoothing term is the squared magnitude of the surface gradient. Results described construct a height map on a
  128 x 128 grid. We use an approximate Newton method (LBFGS for the Hessian approximation), and use multigrid to
  accelerate convergence.  This reconstruction procedure can be dependent on the start point.  We obtain a start point
  from a random initialization by building a very coarse scale reconstruction (we use 30x30) of a normal field that
  minimizes shading error, meets boundary conditions (Augmented Lagrangian Method) and is integrable (ALM). We then
  integrate that normal field to yield a start point. 
\section{Least Square Formulation for Spherical Harmonics Loss $L_Y$}
Consider we have $k=m \times n$ pixels in an image, our $\mathcal{N} \in {R}^{k \times 3}$ 
and $\mathcal{S} \in {R}^{k \times 1}$. 
We estimate  first 9 basis ($\mathcal{B}(\mathcal{N})\in {R}^{k \times 9}$) from $\mathcal{N}$.
We can now write $\mathcal{S} = Y \times \mathcal{B}(\mathcal{N})$.
The solution for $Y$ is then $\mathcal{B}(\mathcal{N})^\dagger\mathcal{S}$. {$\mathcal{B}(\mathcal{N})^\dagger$ is pseudo-inverse of  $\mathcal{B}(\mathcal{N})$.}
\section{Our Decomposition Model}
Our  decomposer works as follows.  An encoder, with
a resnet structure, produces an image code.  This is decoded (again
with a resnet structure) into an albedo field, a shading field and a gloss field.
The albedo field is colored.  Shading and gloss field are colorless.
The training process accepts albedo, shading and gloss paradigms (generated as
samples in advance and cached) and real images from~\cite{zhou2017scene}.  
The real images are given without any of their ground truth. We use them to ensure our image decomposition inferences when recomposed back are able to produce an image that very much looks like real images. We update our decomposer with this residual loss after training with about 50k paradigm samples. 

There are four losses.  The decomposer must decompose a fake
image (constructed out of randomly selected albedo, shading and gloss
paradigms) into its correct, known paradigms (mixed L1/L2 loss).
The albedo, shading and gloss fields constructed from a real image using our decomposer
must combine into that real image (mixed L1/L2 loss).  The
remaining two losses are adversarial.  The albedo
field constructed from a real image must fool a classifier trained
to distinguish between fake albedo fields and those constructed
from real images. Similarly, the shading field
constructed from a real image must fool a classifier
trained to distinguish between fake shading fields
and those constructed from real images. 

\begin{figure}[htpb!]
  \centering
  \includegraphics[width=\linewidth]{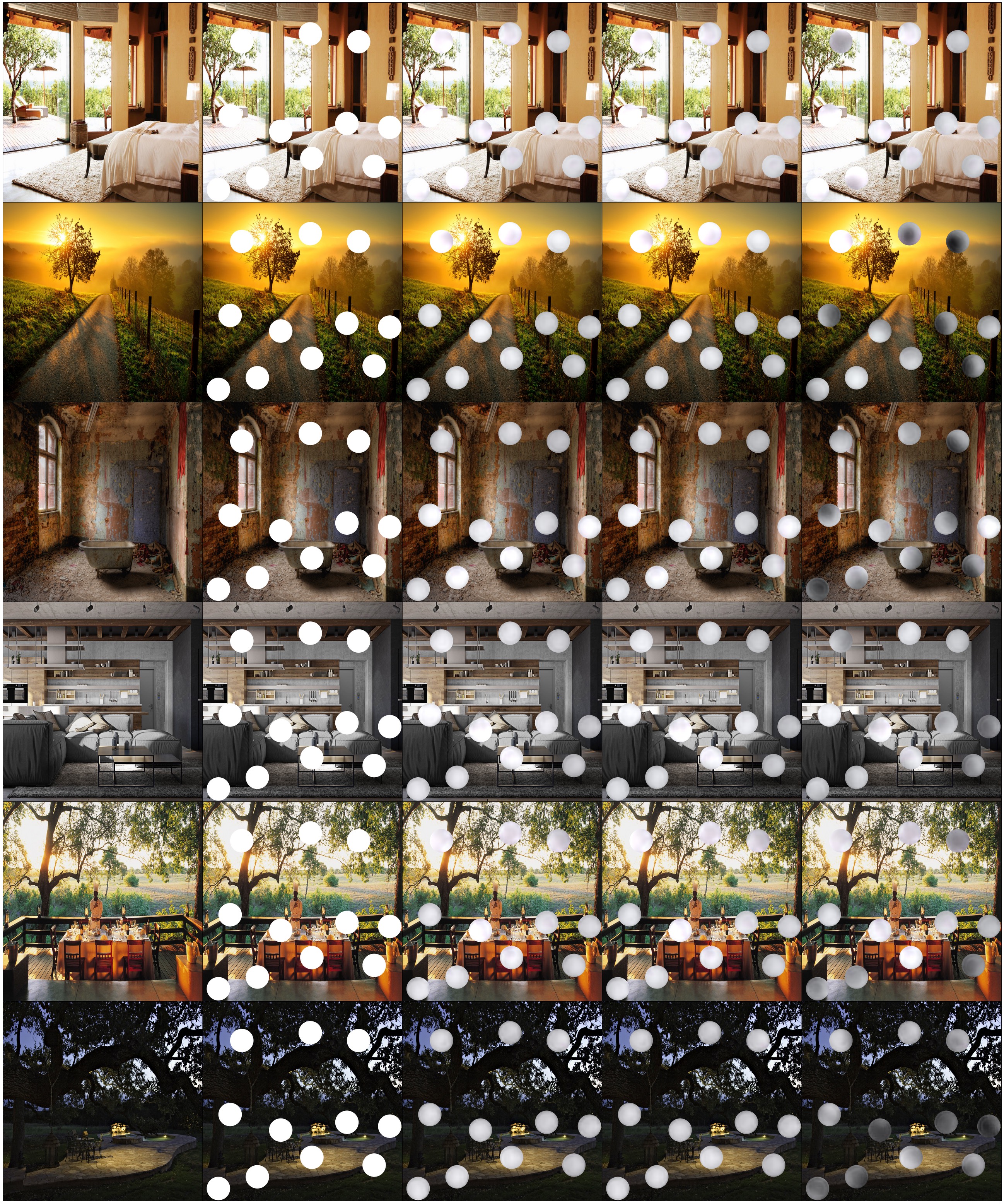}
  \begin{tabularx}{\linewidth}{*{6}{>{\centering\arraybackslash\footnotesize $}X<{$}}}
    \scalebox{1.0}{Target \ Scene} & \scalebox{1.0}{CP} & \scalebox{1.0}{$\mathcal{L}_{decomp}$} & \scalebox{1.0}{$+\mathcal{L}_{Z}$} & \scalebox{1.0}{$+\mathcal{L}_{Z}+\mathcal{L}_{Y}$}
  \end{tabularx}
  \caption{{\bf Ablation.} We show more renderings of spheres from circle cut-outs (see Fig 10 in main text) in scenes exhibiting complex lighting.
  Results improve moving from left to right.}
  \label{fig:spheres_supp}
  \end{figure}

  \section{Post-Processing}
  \label{subsec:postprocess}
  {\noindent \bf Removing DIP artifacts.} 
  The input to DIPR are both the target scene and the cut-and-paste image. We let DIP inpaint the shading of the original target image {\em and} also the composited image for the masked region.  
  This is analogous to rendering scenes twice {\em with} and {\em without} inserted object~\cite{debevec1998RSO, karsch2011rendering}. Doing so means the target scene's shading field acts like a regularizer 
  that prevents DIP copying cut-and-paste. Furthermore, we can remove DIP (network) specific noisy artifacts. We write $\imi_{obj}$ for the target image rendered by DIP with the object and
  $\imi_{noobj}$ for the target scene rendered without object, we then form\vspace{-3pt}
  \begin{equation}
    \imi_{final} = (1-\mask) \odot \imi_{obj} + \mask \odot (\imi_t + \imi_{obj} - \imi_{noobj}) 
  \end{equation}
  {\bf High-resolution rendering.} Our decomposition network can decompose high-resolution intrinsics reliably for real images when trained only on $128$p albedo like intrinsic maps. Since our extrinsic maps are both locally smooth and does not have fine details in them, we can easily upsample them to high-resolution without loss of spatial details. Thus we can render a very high-resolution ($1024$p) composites, without significant loss of geometric detail.

    \begin{figure*}[t!]
      \centering
      \includegraphics[width=\linewidth]{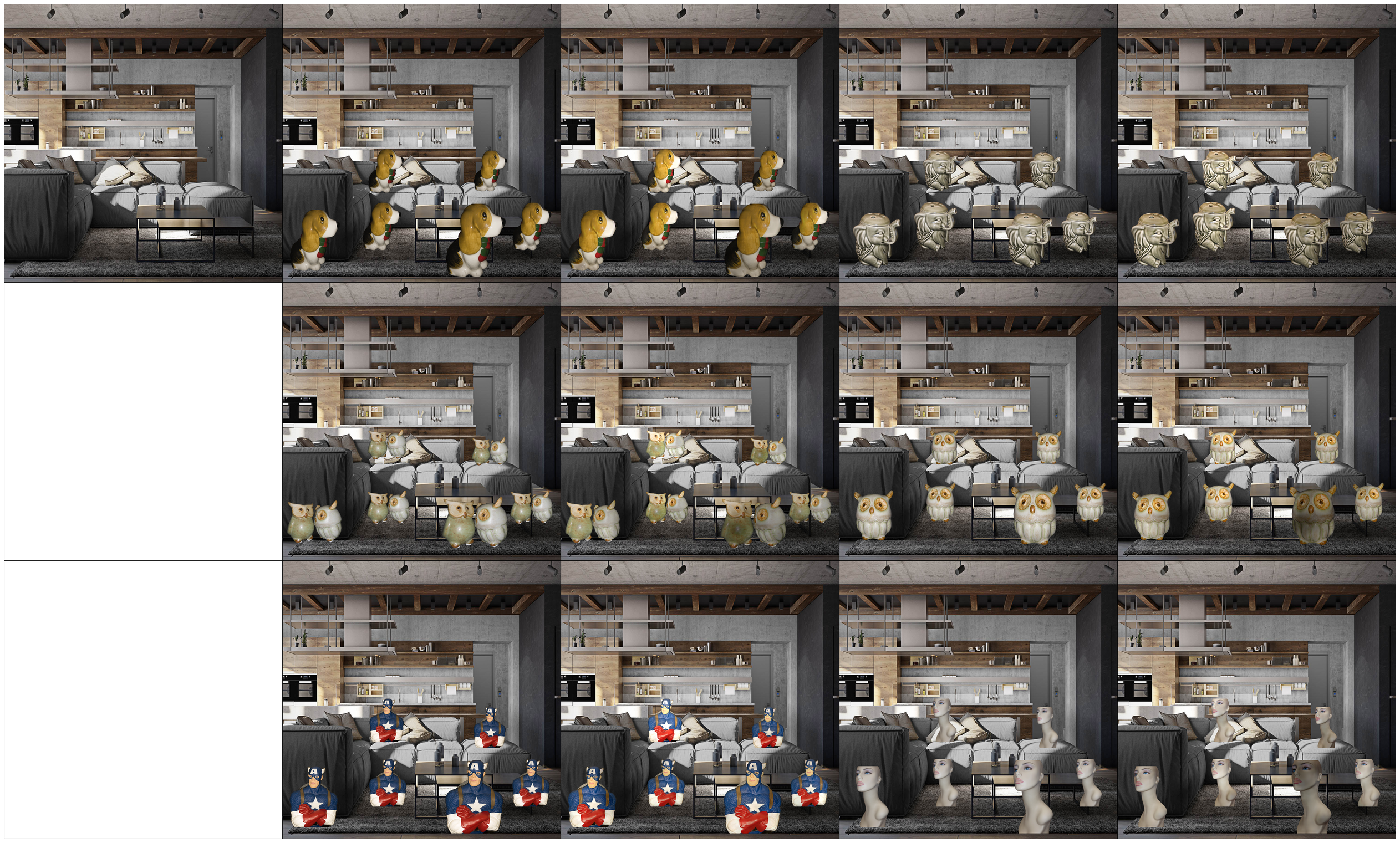}
      \vspace{-5pt}
      \begin{tabularx}{\linewidth}{*{5}{>{\centering\arraybackslash\footnotesize $}X<{$}}}
        \scalebox{1.0}{Target \ Scene} & \scalebox{1.0}{Cut\ \&\ Paste} & \hspace{-2pt}\scalebox{1.0}{DIPR (Ours)} & \scalebox{1.0}{Cut\ \&\ Paste} & \hspace{-2pt}\scalebox{1.0}{DIPR (Ours)}
      \end{tabularx}
      \vspace{-10pt}
      \caption{{\bf Multiple Objects.} We show multiple objects from~\cite{li2018learning} added to a scene at different location produce consistent shading that conforms to the target scene's lighting.}
      \label{fig:other_objects}
      \end{figure*}

\begin{figure*}[t!]
\centering
\includegraphics[width=\linewidth]{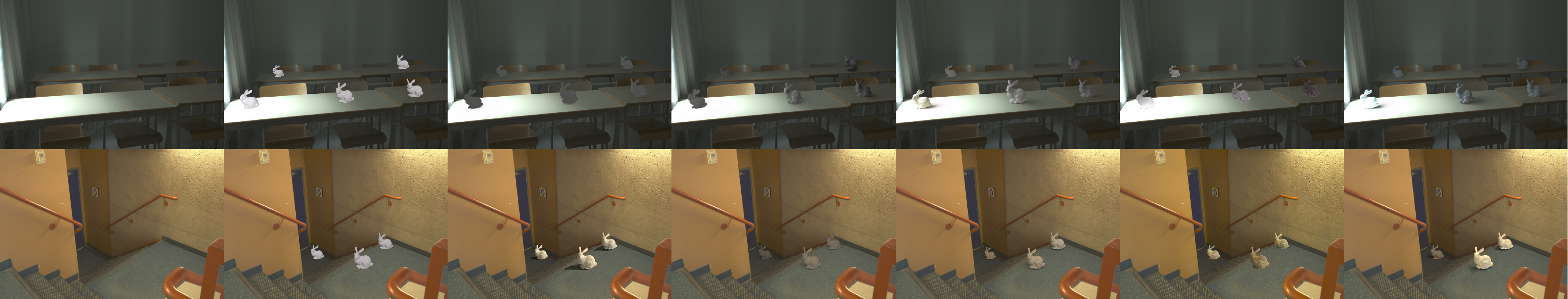}
\centering
\begin{tabularx}{\linewidth}{*{7}{>{\centering\arraybackslash$}X<{$}}} 
\centering
\scalebox{0.95}{Target Scene} & \centering \scalebox{0.95}{Barron \emph{et al.}~\cite{barron2014shape}} &\centering\scalebox{0.95}{Gardner \emph{et al.}~\cite{gardner2019deep}}&\centering \scalebox{0.95}{Garon \emph{et al.}~\cite{garon2019fast}} &\centering \scalebox{0.95}{Li \emph{et al.}~\cite{li2020inverse}} &\centering \scalebox{0.95}{DIPR (ours)} & \centering \scalebox{0.95}{Simulated GT}
\end{tabularx}
\vspace{-10pt}
\caption{Qualitative comparison with SOTA image insertion methods with heavy supervision (which require a geometric and physical model of the inserted objects -- the
bunnies -- and the predicted lighting map for target scene). For the top DIPR image we use Barron's rendering; for the bottom, we use Gardner's rendering.   DIPR produces plausible shadings in each case, but is outperformed by methods that have access to detailed models of illumination and surface models.}
\label{fig:3Dmethodscompare}
\end{figure*}

{\small
\bibliographystyle{ieee_fullname}
\bibliography{main}

\begin{thebibliography}{10}\itemsep=-1pt

\bibitem{bansal2019shapes}
Aayush Bansal, Yaser Sheikh, and Deva Ramanan.
\newblock Shapes and context: In-the-wild image synthesis \& manipulation.
\newblock In {\em Proceedings of the IEEE Conference on Computer Vision and
  Pattern Recognition}, pages 2317--2326, 2019.

\bibitem{barron2014shape}
Jonathan~T Barron and Jitendra Malik.
\newblock Shape, illumination, and reflectance from shading.
\newblock {\em IEEE transactions on pattern analysis and machine intelligence},
  2014.

\bibitem{belhumeur1998set}
Peter~N Belhumeur and David~J Kriegman.
\newblock What is the set of images of an object under all possible
  illumination conditions?
\newblock {\em International Journal of Computer Vision}, 1998.

\bibitem{bell2014intrinsic}
Sean Bell, Kavita Bala, and Noah Snavely.
\newblock Intrinsic images in the wild.
\newblock {\em ACM Transactions on Graphics (TOG)}, 2014.

\bibitem{berzhanskaya2005remote}
Julia Berzhanskaya, Gurumurthy Swaminathan, Jacob Beck, and Ennio Mingolla.
\newblock Remote effects of highlights on gloss perception.
\newblock {\em Perception}, 2005.

\bibitem{bi20151}
Sai Bi, Xiaoguang Han, and Yizhou Yu.
\newblock An l 1 image transform for edge-preserving smoothing and scene-level
  intrinsic decomposition.
\newblock {\em ACM Transactions on Graphics (TOG)}, 2015.

\bibitem{cavanagh2005tracking}
Patrick Cavanagh and George~A Alvarez.
\newblock Tracking multiple targets with multifocal attention.
\newblock {\em Trends in cognitive sciences}, 9(7):349--354, 2005.

\bibitem{chen2017rethinking}
Liang-Chieh Chen, George Papandreou, Florian Schroff, and Hartwig Adam.
\newblock Rethinking atrous convolution for semantic image segmentation.
\newblock {\em arXiv preprint arXiv:1706.05587}, 2017.

\bibitem{cong2020dovenet}
Wenyan Cong, Jianfu Zhang, Li Niu, Liu Liu, Zhixin Ling, Weiyuan Li, and Liqing
  Zhang.
\newblock Dovenet: Deep image harmonization via domain verification.
\newblock In {\em Proceedings of the IEEE/CVF Conference on Computer Vision and
  Pattern Recognition}, pages 8394--8403, 2020.

\bibitem{debevec1998RSO}
Paul Debevec.
\newblock Rendering synthetic objects into real scenes: bridging traditional
  and image-based graphics with global illumination and high dynamic range
  photography.
\newblock In {\em Proceedings of the 25th annual conference on Computer
  graphics and interactive techniques}, 1998.

\bibitem{dwibedi2017cut}
Debidatta Dwibedi, Ishan Misra, and Martial Hebert.
\newblock Cut, paste and learn: Surprisingly easy synthesis for instance
  detection.
\newblock In {\em Proceedings of the IEEE International Conference on Computer
  Vision}, pages 1301--1310, 2017.

\bibitem{fan2018revisiting}
Qingnan Fan, Jiaolong Yang, Gang Hua, Baoquan Chen, and David Wipf.
\newblock Revisiting deep intrinsic image decompositions.
\newblock In {\em Proceedings of the IEEE conference on computer vision and
  pattern recognition}, 2018.

\bibitem{forsyth2020intrinsic}
DA Forsyth and Jason~J Rock.
\newblock Intrinsic image decomposition using paradigms.
\newblock {\em arXiv preprint arXiv:2011.10512}, 2020.

\bibitem{gardner2019deep}
Marc-Andr{\'e} Gardner, Yannick Hold-Geoffroy, Kalyan Sunkavalli, Christian
  Gagn{\'e}, and Jean-Fran{\c{c}}ois Lalonde.
\newblock Deep parametric indoor lighting estimation.
\newblock In {\em Proceedings of the IEEE International Conference on Computer
  Vision}, 2019.

\bibitem{garon2019fast}
Mathieu Garon, Kalyan Sunkavalli, Sunil Hadap, Nathan Carr, and
  Jean-Fran{\c{c}}ois Lalonde.
\newblock Fast spatially-varying indoor lighting estimation.
\newblock In {\em Proceedings of the IEEE Conference on Computer Vision and
  Pattern Recognition}, 2019.

\bibitem{gauthier1998training}
Isabel Gauthier, Pepper Williams, Michael~J Tarr, and James Tanaka.
\newblock Training ‘greeble’experts: a framework for studying expert object
  recognition processes.
\newblock {\em Vision research}, 38(15-16):2401--2428, 1998.

\bibitem{ghiasi2020simple}
Golnaz Ghiasi, Yin Cui, Aravind Srinivas, Rui Qian, Tsung-Yi Lin, Ekin~D Cubuk,
  Quoc~V Le, and Barret Zoph.
\newblock Simple copy-paste is a strong data augmentation method for instance
  segmentation.
\newblock {\em arXiv preprint arXiv:2012.07177}, 2020.

\bibitem{guo2021image}
Zonghui Guo, Dongsheng Guo, Haiyong Zheng, Zhaorui Gu, Bing Zheng, and Junyu
  Dong.
\newblock Image harmonization with transformer.
\newblock In {\em Proceedings of the IEEE/CVF International Conference on
  Computer Vision}, pages 14870--14879, 2021.

\bibitem{hold2019deep}
Yannick Hold-Geoffroy, Akshaya Athawale, and Jean-Fran{\c{c}}ois Lalonde.
\newblock Deep sky modeling for single image outdoor lighting estimation.
\newblock In {\em Proceedings of the IEEE Conference on Computer Vision and
  Pattern Recognition}, 2019.

\bibitem{jia2006drag}
Jiaya Jia, Jian Sun, Chi-Keung Tang, and Heung-Yeung Shum.
\newblock Drag-and-drop pasting.
\newblock {\em ACM Transactions on graphics (TOG)}, 25(3):631--637, 2006.

\bibitem{karsch2011rendering}
Kevin Karsch, Varsha Hedau, David Forsyth, and Derek Hoiem.
\newblock Rendering synthetic objects into legacy photographs.
\newblock {\em ACM Transactions on Graphics (TOG)}, 2011.

\bibitem{karsch2014automatic}
Kevin Karsch, Kalyan Sunkavalli, Sunil Hadap, Nathan Carr, Hailin Jin, Rafael
  Fonte, Michael Sittig, and David Forsyth.
\newblock Automatic scene inference for 3d object compositing.
\newblock {\em ACM Transactions on Graphics (TOG)}, 2014.

\bibitem{lalonde2007photoclipart}
Jean-Fran{\c{c}}ois Lalonde, Derek Hoiem, Alexei~A Efros, Carsten Rother, John
  Winn, and Antonio Criminisi.
\newblock Photo clip art.
\newblock {\em ACM transactions on graphics (TOG)}, 2007.

\bibitem{land1959color1}
Edwin~H Land.
\newblock Color vision and the natural image. part i.
\newblock {\em Proceedings of the National Academy of Sciences of the United
  States of America}, 1959.

\bibitem{land1959color2}
Edwin~H Land.
\newblock Color vision and the natural image part ii.
\newblock {\em Proceedings of the National Academy of Sciences of the United
  States of America}, 1959.

\bibitem{land1977retinex}
Edwin~H Land.
\newblock The retinex theory of color vision.
\newblock {\em Scientific american}, 1977.

\bibitem{land1971lightness}
Edwin~H Land and John~J McCann.
\newblock Lightness and retinex theory.
\newblock {\em Josa}, 1971.

\bibitem{lee2018context}
Donghoon Lee, Sifei Liu, Jinwei Gu, Ming~Yu Liu, Ming~Hsuan Yang, and Jan
  Kautz.
\newblock Context-aware synthesis and placement of object instances.
\newblock In {\em Advances in Neural Information Processing Systems}, 2018.

\bibitem{li2020inverse}
Zhengqin Li, Mohammad Shafiei, Ravi Ramamoorthi, Kalyan Sunkavalli, and
  Manmohan Chandraker.
\newblock Inverse rendering for complex indoor scenes: Shape, spatially-varying
  lighting and svbrdf from a single image.
\newblock In {\em Proceedings of the IEEE Conference on Computer Vision and
  Pattern Recognition}, 2020.

\bibitem{li2018cgintrinsics}
Zhengqi Li and Noah Snavely.
\newblock Cgintrinsics: Better intrinsic image decomposition through
  physically-based rendering.
\newblock In {\em Proceedings of the European Conference on Computer Vision
  (ECCV)}, 2018.

\bibitem{li2018learning}
Zhengqi Li and Noah Snavely.
\newblock Learning intrinsic image decomposition from watching the world.
\newblock In {\em Proceedings of the IEEE Conference on Computer Vision and
  Pattern Recognition}, 2018.

\bibitem{li2018learningspat}
Zhengqin Li, Zexiang Xu, Ravi Ramamoorthi, Kalyan Sunkavalli, and Manmohan
  Chandraker.
\newblock Learning to reconstruct shape and spatially-varying reflectance from
  a single image.
\newblock {\em ACM Transactions on Graphics (TOG)}, 2018.

\bibitem{liao2012building}
Zicheng Liao, Ali Farhadi, Yang Wang, Ian Endres, and David Forsyth.
\newblock Building a dictionary of image fragments.
\newblock In {\em 2012 IEEE Conference on Computer Vision and Pattern
  Recognition}, pages 3442--3449. IEEE, 2012.

\bibitem{liao2015cvpr}
Zicheng Liao, Kevin Karsch, and David Forsyth.
\newblock An approximate shading model for object relighting.
\newblock In {\em Proceedings of the IEEE Conference on Computer Vision and
  Pattern Recognition}, 2015.

\bibitem{liao2019IJCV}
Zicheng Liao, Kevin Karsch, Hongyi Zhang, and David Forsyth.
\newblock An approximate shading model with detail decomposition for object
  relighting.
\newblock {\em International Journal of Computer Vision}, 2019.

\bibitem{lin2014microsoft}
Tsung-Yi Lin, Michael Maire, Serge Belongie, James Hays, Pietro Perona, Deva
  Ramanan, Piotr Doll{\'a}r, and C~Lawrence Zitnick.
\newblock Microsoft coco: Common objects in context.
\newblock In {\em European conference on computer vision}. Springer, 2014.

\bibitem{ling2021region}
Jun Ling, Han Xue, Li Song, Rong Xie, and Xiao Gu.
\newblock Region-aware adaptive instance normalization for image harmonization.
\newblock In {\em Proceedings of the IEEE/CVF Conference on Computer Vision and
  Pattern Recognition}, 2021.

\bibitem{liu2018image}
Guilin Liu, Fitsum~A Reda, Kevin~J Shih, Ting-Chun Wang, Andrew Tao, and Bryan
  Catanzaro.
\newblock Image inpainting for irregular holes using partial convolutions.
\newblock In {\em Proceedings of the European Conference on Computer Vision
  (ECCV)}, 2018.

\bibitem{mildenhall2020nerf}
Ben Mildenhall, Pratul~P Srinivasan, Matthew Tancik, Jonathan~T Barron, Ravi
  Ramamoorthi, and Ren Ng.
\newblock Nerf: Representing scenes as neural radiance fields for view
  synthesis.
\newblock {\em arXiv preprint arXiv:2003.08934}, 2020.

\bibitem{mittal2012making}
Anish Mittal, Rajiv Soundararajan, and Alan~C Bovik.
\newblock Making a “completely blind” image quality analyzer.
\newblock {\em IEEE Signal processing letters}, 20(3):209--212, 2012.

\bibitem{nekrasov2019real}
Vladimir Nekrasov, Thanuja Dharmasiri, Andrew Spek, Tom Drummond, Chunhua Shen,
  and Ian Reid.
\newblock Real-time joint semantic segmentation and depth estimation using
  asymmetric annotations.
\newblock In {\em 2019 International Conference on Robotics and Automation
  (ICRA)}. IEEE, 2019.

\bibitem{nestmeyerlearning}
Thomas Nestmeyer, Jean-Fran{\c{c}}ois Lalonde, Iain Matthews, Epic Games,
  Andreas Lehrmann, and AI Borealis.
\newblock Learning physics-guided face relighting under directional light.
\newblock 2020.

\bibitem{perez2003poisson}
Patrick P{\'e}rez, Michel Gangnet, and Andrew Blake.
\newblock Poisson image editing.
\newblock In {\em ACM SIGGRAPH 2003 Papers}, pages 313--318, 2003.

\bibitem{perlin1985image}
Ken Perlin.
\newblock An image synthesizer.
\newblock {\em ACM Siggraph Computer Graphics}, 1985.

\bibitem{ramachandran1988perceiving}
Vilayanur~S Ramachandran.
\newblock Perceiving shape from shading.
\newblock {\em Scientific American}, 1988.

\bibitem{ramamoorthi2001relationship}
Ravi Ramamoorthi and Pat Hanrahan.
\newblock On the relationship between radiance and irradiance: determining the
  illumination from images of a convex lambertian object.
\newblock {\em JOSA A}, 2001.

\bibitem{schonfeld2020u}
Edgar Sch{\"o}nfeld, Bernt Schiele, and Anna Khoreva.
\newblock A u-net based discriminator for generative adversarial networks.
\newblock {\em arXiv preprint arXiv:2002.12655}, 2020.

\bibitem{shih20203d}
Meng-Li Shih, Shih-Yang Su, Johannes Kopf, and Jia-Bin Huang.
\newblock 3d photography using context-aware layered depth inpainting.
\newblock In {\em Proceedings of the IEEE/CVF Conference on Computer Vision and
  Pattern Recognition}, pages 8028--8038, 2020.

\bibitem{song2019neural}
Shuran Song and Thomas Funkhouser.
\newblock Neural illumination: Lighting prediction for indoor environments.
\newblock In {\em Proceedings of the IEEE Conference on Computer Vision and
  Pattern Recognition}, 2019.

\bibitem{srinivasan2020lighthouse}
Pratul~P Srinivasan, Ben Mildenhall, Matthew Tancik, Jonathan~T Barron, Richard
  Tucker, and Noah Snavely.
\newblock Lighthouse: Predicting lighting volumes for spatially-coherent
  illumination.
\newblock In {\em Proceedings of the IEEE Conference on Computer Vision and
  Pattern Recognition}, 2020.

\bibitem{sun2019single}
Tiancheng Sun, Jonathan~T Barron, Yun-Ta Tsai, Zexiang Xu, Xueming Yu, Graham
  Fyffe, Christoph Rhemann, Jay Busch, Paul Debevec, and Ravi Ramamoorthi.
\newblock Single image portrait relighting.
\newblock {\em ACM Transactions on Graphics (Proceedings SIGGRAPH)}, 2019.

\bibitem{sunkavalli2010multi}
Kalyan Sunkavalli, Micah~K Johnson, Wojciech Matusik, and Hanspeter Pfister.
\newblock Multi-scale image harmonization.
\newblock {\em ACM Transactions on Graphics (TOG)}, 29(4):1--10, 2010.

\bibitem{tsai2017Deep}
Yi-Hsuan Tsai, Xiaohui Shen, Zhe Lin, Kalyan Sunkavalli, Xin Lu, and Ming-Hsuan
  Yang.
\newblock Deep image harmonization.
\newblock In {\em Proceedings of the IEEE Conference on Computer Vision and
  Pattern Recognition}, pages 3789--3797, 2017.

\bibitem{ulyanov2018deep}
Dmitry Ulyanov, Andrea Vedaldi, and Victor Lempitsky.
\newblock Deep image prior.
\newblock In {\em Proceedings of the IEEE Conference on Computer Vision and
  Pattern Recognition}, 2018.

\bibitem{wang2019cnngenerated}
Sheng-Yu Wang, Oliver Wang, Richard Zhang, Andrew Owens, and Alexei~A Efros.
\newblock Cnn-generated images are surprisingly easy to spot...for now.
\newblock In {\em CVPR}, 2020.

\bibitem{yu2019inverserendernet}
Ye Yu and William~AP Smith.
\newblock Inverserendernet: Learning single image inverse rendering.
\newblock In {\em Proceedings of the IEEE Conference on Computer Vision and
  Pattern Recognition}, 2019.

\bibitem{zhang2018perceptual}
Richard Zhang, Phillip Isola, Alexei~A Efros, Eli Shechtman, and Oliver Wang.
\newblock The unreasonable effectiveness of deep features as a perceptual
  metric.
\newblock In {\em CVPR}, 2018.

\bibitem{zhou2017scene}
Bolei Zhou, Hang Zhao, Xavier Puig, Sanja Fidler, Adela Barriuso, and Antonio
  Torralba.
\newblock Scene parsing through ade20k dataset.
\newblock In {\em Proceedings of the IEEE conference on computer vision and
  pattern recognition}, 2017.

\bibitem{zhou2019deep}
Hao Zhou, Sunil Hadap, Kalyan Sunkavalli, and David~W Jacobs.
\newblock Deep single-image portrait relighting.
\newblock In {\em Proceedings of the IEEE International Conference on Computer
  Vision}, 2019.

\end{thebibliography}
}
\end{document}